\documentclass[10pt,journal,comsoc]{IEEEtran}
\usepackage[T1]{fontenc}
\usepackage{cite}
\usepackage{amsmath,amssymb,amsfonts}
\usepackage{algorithmic}
\usepackage{graphicx}
\usepackage{textcomp}
\usepackage{bm}
\usepackage{bbm} 
\usepackage{dsfont}
\usepackage{arydshln}
\usepackage[table]{xcolor}
\usepackage{caption}
\usepackage{ifthen}
\usepackage{amsmath}
\usepackage{amsthm}
\usepackage{placeins} 
\usepackage{float}
\usepackage{setspace}
\usepackage{amsfonts}
\usepackage{amssymb}
\usepackage[vvarbb]{newtxmath}
\usepackage{stfloats}
\usepackage{ragged2e}
\usepackage[ruled,vlined,linesnumbered]{algorithm2e}
\usepackage{amsfonts}
\usepackage{mathrsfs}
\usepackage{amsmath,amsthm}
\usepackage{array,booktabs}
\usepackage{subfigure}
\usepackage{multirow}
\usepackage{cuted}
\usepackage{multicol}
\usepackage{subfigure}
\usepackage{threeparttable}
\usepackage{dcolumn}
\usepackage{setspace}
\usepackage{multirow}
\usepackage{newtxtext} 
\usepackage{makecell}
\usepackage{lipsum}
\usepackage{hyperref}
\usepackage{enumerate}
\newtheorem{theorem}{Theorem}

\newtheorem{remark}{Remark}

\usepackage{tabularx}    
\usepackage{multirow}   
\usepackage{amsmath}      
\usepackage{graphicx}      
\usepackage{array}         
\definecolor{verylightgray}{rgb}{0.5, 0.6, 0.85}
\definecolor{veryverylightblue}{rgb}{0.765, 0.824, 0.918}
\usepackage{mathrsfs}
\allowdisplaybreaks
\usepackage[font=footnotesize]{caption}
\begin{document}
\setlength{\abovedisplayskip}{3.5pt}
\setlength{\belowdisplayskip}{3.5pt}
\everymath{\footnotesize} 
\everydisplay{\small} 
	\title{Adaptive UAV-Assisted Hierarchical Federated Learning: Optimizing Energy, Latency, and Resilience for Dynamic Smart IoT}
	
	\author{Xiaohong Yang, Minghui Liwang, \textit{Senior Member},  
   \textit{IEEE}, Liqun Fu, \textit{Senior Member}, \textit{IEEE},\\ Yuhan Su, Seyyedali Hosseinalipour, \textit{Senior Member}, \textit{IEEE},\\ Xianbin Wang, \textit{Fellow}, \textit{IEEE}, Yiguang Hong, \textit{Fellow}, \textit{IEEE}
		\thanks{This work was supported in part by Shanghai Municipal Science and Technology Major Project under Grant no. 2021SHZDZX0100; Shanghai Pujiang Programme under Grant no. 24PJD117, National Natural Science Foundation of China under Grant nos. 62271424, 62401486, 62088101, 72171172, Chinese Academy of Engineering, Strategic Research and Consulting Program under Grant no. 2023-XZ-65, U.S. National Science Foundation (NSF) under Grant No. ECCS-2512911. Xiaohong Yang and Liqun Fu are with the School of Informatics, Xiamen University, Fujian, China. Minghui Liwang and Yiguang Hong are with the Department of Control Science and Engineering, the National Key Laboratory of Autonomous Intelligent Unmanned Systems, and also with Frontiers Science Center for Intelligent Autonomous Systems, Ministry of Education, Tongji University, Shanghai, China. Yuhan Su is with the School of Electronic Science and Engineering, Xiamen University, Xiamen, China. Seyyedali Hosseinalipour is with the department of Electrical Engineering, University at Buffalo-SUNY, Buffalo, NY, USA. Xianbin Wang is with the Department of Electrical and Computer Engineering, Western University, Ontario, Canada (Email: xiaohongyang@stu.xmu.edu.cn, minghuiliwang@tongji.edu.cn, liqun@xmu.edu.cn, ysu@xmu.edu.cn, alipour@buffalo.edu, xianbin.wang@uwo.ca, yghong@tongji.edu.cn). Corresponding author: Minghui Liwang.
	}
    }

	\IEEEtitleabstractindextext{
		\begin{abstract}
			\justifying
            Hierarchical Federated Learning (HFL) extends conventional Federated Learning (FL) by introducing intermediate aggregation layers, enabling distributed learning in geographically dispersed environments, particularly relevant for smart IoT systems, such as remote monitoring and battlefield operations, where cellular connectivity is limited. In these scenarios, UAVs serve as mobile aggregators, dynamically connecting terrestrial IoT devices. This paper investigates an HFL architecture with energy-constrained, dynamically deployed UAVs prone to communication disruptions. We propose a novel approach to minimize global training costs by formulating a joint optimization problem that integrates learning configuration, bandwidth allocation, and device-to-UAV association, ensuring timely global aggregation before UAV disconnections and redeployments. The problem accounts for dynamic IoT devices and intermittent UAV connectivity and is NP-hard. To tackle this, we decompose it into three subproblems: \textit{(i)} optimizing learning configuration and bandwidth allocation via an augmented Lagrangian to reduce training costs; \textit{(ii)} introducing a device fitness score based on data heterogeneity (via Kullback-Leibler divergence), device-to-UAV proximity, and computational resources, using a TD3-based algorithm for adaptive device-to-UAV assignment; \textit{(iii)} developing a low-complexity two-stage greedy strategy for UAV redeployment and global aggregator selection, ensuring efficient aggregation despite UAV disconnections. Experiments on diverse real-world datasets validate the approach, demonstrating cost reduction and robust performance under communication disruptions.

		\end{abstract}

	}
	
\maketitle
\IEEEdisplaynontitleabstractindextext

\IEEEpeerreviewmaketitle
\section{Introduction}
\noindent \IEEEPARstart{W}{ith} the growing demand for machine learning (ML) in the Internet of Things (IoT), traditional centralized approaches face challenges such as privacy risks, security issues, and inefficient use of distributed data \cite{ASurvey, MOBFL, AssistedWNet}.
Federated Learning (FL) addresses these by aggregating locally trained models on the cloud without sharing raw data, making it promising for smart IoT applications \cite{Split, FedMDS, Accelerating, Staleness}. Yet, direct communication between the cloud and numerous devices is often hindered by link outages, latency, and backhaul congestion \cite{VISIT, MobilityAware, Cost-Efficient}. Hierarchical Federated Learning (HFL) alleviates these issues by introducing edge servers for near-device aggregation, where local models are combined before global aggregation at the cloud \cite{JointEdge, Context, HiFlash, HFEL}. This hierarchical design enhances communication efficiency and scalability, making HFL particularly suitable for large-scale, geographically distributed IoT networks \cite{Context, Towarddy}.

\subsection{Motivation and Challenges}
A key application of HFL lies in smart IoT systems deployed in environments with limited or unreliable cellular connectivity, such as remote monitoring, disaster response, and battlefield operations \cite{Adaptive, Stateof}. In these scenarios, direct connections between IoT devices and edge servers via base stations or roadside units are often infeasible. UAVs can instead serve as flexible model aggregators \cite{Privacy, EdgeComputing}, giving rise to UAV-assisted HFL. Recent efforts include \textit{Huang et al.} \cite{FairResource}, who proposed HFL over space-air-ground integrated networks with UAVs as edge servers; \textit{Song et al.} \cite{Multitaska}, who optimized computation offloading in air-ground networks via a DQN-based method; \textit{Li et al.} \cite{HFLOD}, who introduced HFL-object detection (OD) to enhance UAV swarm reliability; \textit{Zhao et al.} \cite{SafS}, who designed a UAV-assisted HFL scheme for IIoT inspection under unstable communications and energy limits; and \textit{Li et al.} \cite{MADDHFL}, who developed an HFL-driven caching strategy for vehicular networks using improved-MADDPG-based optimization. While the aforementioned works have made notable contributions, we identify two critical challenges that remain underexplored: \textit{(i)} IoT Device Dynamics and Heterogeneity and \textit{(ii)} UAV’s Energy Constraints and Reliability.

\noindent \textit{\textbf{Challenge (i). IoT Device Dynamics and Heterogeneity:}} 
Inefficient tuning of iteration frequency, resource allocation, and aggregation topology, coupled with device mobility, data heterogeneity, and limited computation and communication resources, significantly hinders the efficiency and convergence of global models in UAV-assisted HFL. These challenges undermine consistent and optimal performance in dynamic, heterogeneous IoT environments. Prior works have attempted to address these issues. \textit{Liu et al.} \cite{TimeMin} optimized local and edge iterations and device-to-edge associations based on signal-to-noise ratio (SNR). \textit{Qi et al.} \cite{Based} proposed a learning-based synchronization scheme to enhance resource efficiency and accuracy, while \textit{Li et al.} \cite{EnergyWare} jointly optimized model accuracy, resource availability, and energy consumption. Similarly, \textit{Dong et al.} \cite{FuzzyLogic} designed a fuzzy logic-based device selection scheme. Although primarily targeting terrestrial HFL, these approaches remain effective for mitigating device heterogeneity in UAV-assisted settings. Several studies focus directly on UAV-assisted HFL. For instance, \textit{Tong et al.} \cite{Blockchain} optimized UAV-to-edge associations via resource allocation and device/UAV locations. \textit{Zhagypar et al.} \cite{PerfA} developed UAV-assisted HFL for wireless networks to mitigate unreliable channels, and \textit{Khelf et al.} \cite{Optimization} optimized UAV-to-device association based on uplink delay. Despite these advances, most methods rely on static strategies for device–UAV association, resource allocation, and learning configuration, which fail to adapt to the dynamics of smart IoT environments. In practice, device mobility and fluctuating data distributions due to frequent arrivals and departures can lead to inefficient device selection, suboptimal resource utilization, and slow convergence, ultimately degrading UAV-assisted HFL performance. To overcome these limitations, a paradigm shift toward adaptive and intelligent mechanisms is required, enabling real-time adjustment of association, allocation, and learning to network dynamics \cite{Multitaska, Adaptive, Privacy}. Motivated by this, we propose a framework to jointly optimize energy–delay efficiency, enhance learning performance, and accelerate convergence in dynamic UAV-assisted HFL environments.

\noindent \textit{\textbf{Challenge (ii). UAV's Energy Constraints and Reliability:}}
Limited energy supply of UAVs introduces major challenges for UAV-assisted HFL, as shutdowns and downtimes severely disrupt training and slow convergence. Prior works have aimed to minimize UAV energy consumption and mission delays \cite{situatness, Multi-UAV}, proposed recharging strategies \cite{MachineL}, and considered periodic recharging for energy-efficient HFL \cite{WhenH}. While valuable, these studies primarily address energy efficiency, overlooking the broader impact of UAV disconnections. Lost model updates during UAV disconnections cause incomplete global aggregation and degraded performance, particularly in remote or infrastructure-limited areas where UAVs conduct critical tasks without charging access \cite{InformU, UAVAided}. Moreover, fixed-position UAV deployments \cite{Optimization, AssistedWNet} fail to adapt to device mobility and dynamic networks, leading to resource underutilization and slower convergence when UAVs disconnect. Hence, beyond energy efficiency, adaptive mechanisms to handle UAV disconnections and service interruptions are essential, motivating our framework that integrates UAV energy management, adaptive redeployment, and aggregation topology optimization.

\vfill

\subsection{Overview and Summary of Contributions}
Building on the above, this paper investigates a UAV-assisted HFL architecture for dynamic and heterogeneous IoT environments. In this framework, energy-limited UAVs act as both edge/intermediate and global/terminal aggregators, adjusting their positions based on device configurations. During training, each UAV autonomously adapts its association strategy to heterogeneous devices within its coverage, a factor often neglected in prior studies yet vital for robust learning. This motivates our joint optimization of UAV energy management, adaptive redeployment, and aggregation topology. Table~I compares related works, with details in \textbf{Appendix~A.A}. The main contributions are summarized as follows:

\noindent $\bullet$ We investigate a relatively underexplored UAV-assisted HFL architecture for IoT networks, where energy-constrained UAVs face potential downtimes and intermittent communication disruptions. The framework further accounts for device heterogeneity in computation and communication, as well as mobility, enabling IoT devices to transition across different UAV coverage areas during the HFL process.

\noindent $\bullet$ To ensure efficient model training in this dynamic and heterogeneous IoT environment, we formulate an optimization problem that balances HFL training delay and energy consumption while preserving model accuracy and mitigating the impact of UAV disconnections and downtimes. We unveil the NP-hard nature of this problem, and then decompose it into three interdependent subproblems and develop complementary solutions for  them.

\noindent $\bullet$ In the first subproblem, we optimize the learning configuration of IoT devices, focusing on local iteration numbers and bandwidth allocation among devices connected to UAVs, with the goal of minimizing local training time and energy costs. An augmented Lagrangian optimization algorithm with a penalty term is designed to ensure convergence to the optimal solution.

\noindent $\bullet$ In the second subproblem, we determine effective device-to-UAV associations using a Twin Delayed Deep Deterministic Policy Gradient (TD3) approach. We introduce a model difference score based on Kullback-Leibler Divergence (KLD) to capture data heterogeneity, and combine it with device-to-UAV distance and computing resources to compute a fitness score for optimal pairing. The problem is formulated as a Markov Decision Process (MDP) and solved with a tailored TD3-based reinforcement learning strategy.

\noindent $\bullet$ In the third subproblem, we develop a UAV energy inspection and location optimization mechanism to decide whether UAVs act as intermediate or global aggregators, establishing optimal aggregation schedules and mitigating data loss from device-to-UAV (D2U) and UAV-to-UAV
(U2U) link interruptions. Each UAV dynamically adjusts its position based on its role to maximize device coverage and reduce communication costs. A low-complexity dual-stage heuristic algorithm is proposed to efficiently solve this subproblem.

\noindent $\bullet$ Extensive experiments on diverse real-world datasets show that our method reduces training costs in terms of energy and delay while maintaining strong model convergence. Compared to existing approaches, it better prevents loss of device model updates and delays in global aggregation caused by D2U interruptions and UAV downtimes.

\begin{figure*}[t!]
	\centering
	\includegraphics[trim=0cm 0cm 0cm 0cm, clip, width=1.95\columnwidth]{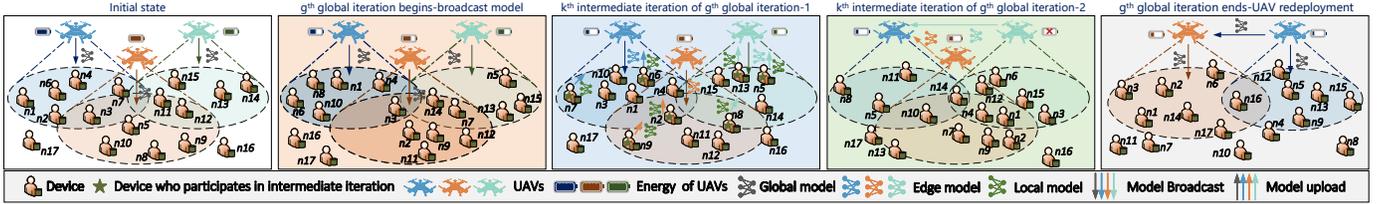}
	\caption{A schematic of HFL architecture over smart IoT of our interest with energy-constrained UAVs and dynamic devices that move between various UAV's coverage areas.}
	\label{fig_1}
\end{figure*}

\section{System Overview} 
\subsection{IoT Device and UAV Operations}
As shown as Fig. 1, we consider a dynamic and heterogeneous IoT network with multiple energy-constrained UAVs collected by the set $\mathcal{M}=\{1,...,M\}$, and multiple terrestrial IoT devices collected by the set $\mathcal{N}=\{1,...,N\}$, which can move across the coverage regions of the UAVs. In a nutshell, the procedure of HFL in our scenario of interest can be summarized as follows:

\noindent {\textbf{$\bullet$ {Part 1. Initial Model Broadcast and Device Selection:}}} The UAV serving as global aggregator (see Parts 2 and 3) distributes the global model to all UAVs. In the first round, the aggregator and model are randomly initialized. Each UAV broadcasts the model to nearby devices, which are adaptively selected for training based on data quality, resource capacity, and proximity. Selected devices train locally via Stochastic Gradient Descent (SGD) and return their updates to the assigned UAVs, initiating the second phase of the methodology described below.

\noindent {$\bullet$ \textbf{Part 2. Intermediate Aggregation and UAV Energy Evaluation:}}  
Upon receiving updated models from devices, each UAV performs intermediate aggregation to generate an updated intermediate local model. UAVs then evaluate their energy status, leading to two possible scenarios: \textit{Case 1:} If all UAVs have sufficient energy, they broadcast their aggregated intermediate models to covered devices, initiating the next local training round. To prevent bias, a UAV is periodically selected for global aggregation after a predefined number of edge aggregation rounds, even when energy is sufficient. \textit{Case 2:} If any UAV lacks sufficient energy, a UAV is chosen as the global aggregator, and all UAVs upload their intermediate models for global aggregation. UAVs with depleted energy exit the network to return to base nodes for recharging. In both cases, completion of global aggregation triggers Part 3 of the methodology, as detailed below.





\noindent {$\bullet$ \textbf{Part 3. Global Aggregation and UAV Relocation:}} Once a global aggregation is triggered, the designated global aggregator UAV disseminates the updated global model to all active UAVs in the network. These UAVs may then relocate to maximize coverage and enhance model performance, particularly in cases where some UAVs have exited the network for recharging. This repositioning allows UAVs and IoT devices to continue participating in subsequent intermediate and global aggregation rounds, ensuring the uninterrupted progression of the HFL process.

This cycle, comprising local training and intermediate iterations, UAV energy evaluations, and global aggregations, repeats until the global model converges to the desired performance level. Henceforth, we use $k$ and $g$ to denote a specific intermediate and global aggregation round, respectively.

\begin{remark}[Departure and Arrival of UAVs]
This paper focuses on the operations triggered by UAV disconnections and downtimes, including D2U re-association, bandwidth reallocation, and UAV repositioning, which are essential for maintaining system performance and training continuity. Although UAVs may later recharge and rejoin the FL process, we do not explicitly describe the rejoining procedures, as they mirror the operations performed during disconnections. Omitting a separate discussion avoids redundancy and keeps the presentation concise.
\end{remark}

\subsection{Modeling of IoT Devices and UAVs}

\noindent\textit{\textbf{Modeling of Devices and UAVs:}} We presume that each terrestrial IoT device $n\in\mathcal{N}$ has a local dataset denoted by $\mathcal{D}_n=\{(x_j,y_j)|1\leq j\leq|\mathcal{D}_n|\}$, where $x_{j}$ and $y_{j}$ refer to the feature vector and label of the $j^\text{th}$ local data point, respectively. We denote the set of IoT devices covered by active UAV $m\in\mathcal{M}$ during the $g^\text{th}$ global aggregation as $\mathcal{N}_{m;[g]}^\mathsf{Cov}$.  Moreover, at each global iteration, a device has a probability $\xi$ of moving into the coverage area of other UAVs. We also denote the subset of IoT devices in the coverage of UAV $m$ that are selected/chosen and participate in each round of model training as $\mathcal{N}_{m;[g]}^\mathsf{Sel}$. We further denote the coordinates of each device $n$ positions during global aggregation $g$ as $\vmathbb{p}_{n;[g]}=(\vmathbb{p}_{n;[g]}(x),\vmathbb{p}_{n;[g]}(y))$, where $\vmathbb{p}_{n;[g]}(x)$ and $\vmathbb{p}_{n;[g]}(y)$ denote the $x$ and $y$ locations of the device, respectively. We assume that the locations of devices remain stationary within each global aggregation round, although they may change between successive rounds. Furthermore, between different global iterations, each device has a certain probability of remaining within the coverage area of its current UAV or transitioning to the coverage area of another UAV. 

Additionally, the location of each UAV $m$ during global aggregation round $g$ is denoted as 
$\mathtt{p}_{m;[g]}=\left(\mathtt{p}_{m;[g]}(x), \mathtt{p}_{m;[g]}(y), \mathscr{H}\right)$, 
where $\mathscr{H}$ represents the altitude at which the UAVs are deployed. This spatial representation plays a crucial role in network  optimization within our UAV-assisted HFL framework. We also consider a realistic scenario where each UAV $m$ has a limited battery capacity, denoted by $E^{\mathsf{Batt}}_m$. The battery depletes at a rate of $\overline{p_m}$ (in Watts) while the UAV is hovering in the air. Additionally, each UAV can relocate at a specific moving speed in the air, represented by $V_m$. These factors influence the UAV's operational time, mobility constraints, and overall network sustainability in our UAV-assisted HFL framework.

\noindent\textit{\textbf{Communication Models in UAV-Assisted HFL:}} In our HFL scenarios of interest, three types of model transfers occur: device-to-UAV (D2U), UAV-to-device (U2D), and UAV-to-UAV (U2U) communications, while detailed definitions in mathematical form are provided by \textbf{Appendix A.B}, due to space limitation. Due to the sequential nature of the local training, intermediate aggregation, and global aggregation processes, these communications occur in separate phases and do not interfere with each other. For example, D2U transmissions never take place when U2U communications are happening, and similarly, U2D transmissions are scheduled separately from both D2U and U2U transmissions. Furthermore, in each mode of communication, bandwidth allocation will later be designed to ensure non-overlapping frequency resources among devices and UAVs, removing the impact of interference in our analysis.


\section{Energy-Constrained UAV-Assisted HFL Over Dynamic IoT}
We let \(K_{[g]} \) denote the number of intermediate aggregations performed during global aggregation \( g \). In our scenario, \( K_{[g]} \) is dynamically tuned based on the interplay between network configurations (e.g., remaining UAV battery levels and delays in model transfers from IoT devices to UAVs) and ML performance. In Section III. B, we address these factors jointly and determine the optimized value of \( K_{[g]} \) for each global aggregation \( g \). To capture the roles and status of UAVs in our HFL scenario, we define \( \mathcal{M}_{[g]} 
\) as the set of UAVs that are  active in model transfer and local aggregation during global aggregation \( g \). Also, we introduce the binary variables \( \phi_{[g]}\in \{0,1\}
\) and \( X_{m;[g]} \in \{0,1\} \) to indicate UAV disconnections and global aggregator selection, respectively. Specifically,  
\(\phi_{[g]} = 1 \) indicates that at least one UAV does not have enough battery to continue staying the network during global round $g$, and    
\(  X_{m;[g]} = 1 \) signifies that UAV \( m \) has been selected as the global aggregator for global round $g$. 
 
\subsection{Energy-Constrained UAV-assisted HFL}
We now formalize the UAV-assisted HFL framework. The process begins at IoT devices, which conduct local training, generate model updates, and communicate them to UAVs. UAVs act as intermediate aggregators, collecting local models and performing partial aggregations. A designated UAV then serves as the global aggregator, gathering intermediate models from other UAVs and executing the final global aggregation.

\noindent\textbf{\textit{(A) Device Operations:}} During the HFL model training period, each device aims to minimize its local loss function. In particular, the local loss function for device \( n \) under an arbitrary model parameter $w$ is given by
$
	L_n (w) = \sum_{j \in \mathcal{D}_n} l_j (w)
$,
where \( l_j \) represents the loss for the $j^\text{th}$ local data point, commonly defined using cross-entropy loss or other relevant error metrics. 

To minimize its local loss function during the local learning iterations, each IoT device maintains a local model, denoted as \( w^{\mathsf{Dev}}_{n;[g,k,h]} \), where the indices represent the specific global aggregation round \( g \), intermediate aggregation round \( k \), and local iteration \( h \). This local model is synchronized with the received global model at the beginning of each global aggregation round, ensuring alignment with the latest updates from the network, as $w^{\mathsf{Dev}}_{n;[g,0,0]} = w_{[g-1]}$, where \( w_{[g-1]} \) is the global model from the previous round. Also, at the beginning of each intermediate aggregation round $k$, the local model at device \( n \) is reinitialized with the latest model received from its assigned UAV $m$ as $w^{\mathsf{Dev}}_{n;[g,k,0]} = w^{\mathsf{UAV}}_{m;[g,k-1]},$
where \( w^{\mathsf{UAV}}_{m;[g,k-1]} \) is the intermediate model aggregated by UAV \( m \), which covers device \( n \).

Subsequently, during local training, each device updates its local model using SGD iterations as follows:
\begin{align}
	w^{\mathsf{Dev}}_{n;[g,k,h]} = w^{\mathsf{Dev}}_{n;[g,k,h-1]} - \eta \widetilde{\nabla} L_n (w^{\mathsf{Dev}}_{n;[g,k,h-1]}), \label{1}
\end{align}
where \( \eta \) is the learning rate, and $\widetilde{\nabla}$ represents the stochastic gradient. In particular, the stochastic gradient is computed as the average loss over the sampled mini-batch \( B_{n;[g,k,h]} \) from the local dataset $\mathcal{D}_n$ as $
\widetilde{\nabla} L_n (w^{\mathsf{Dev}}_{n;[g,k,h-1]}) = \frac{1}{|B_{n;[g,k,h]}|} \sum_{j \in B_{n;[g,k,h]}} \nabla l_j(w^{\mathsf{Dev}}_{n;[g,k,h-1]}).$ Assuming that each device performs \( H \) SGD iterations per local training phase, the final trained model at device \( n \), denoted as \( w^{\mathsf{Dev}}_{n;[g,k,H]} \), is transmitted to its assigned UAV. This updated model is then used to compute the next intermediate model at the UAV \( w^{\mathsf{UAV}}_{m;[g,k]} \), as detailed below.

\noindent\textit{\textbf{(B) Intermediate Aggregations:}} After the reception of the local models of the IoT devices contained in the set  $\mathcal{N}_{m;[g]}^\mathsf{Sel}$ at each active UAV $m$, this UAV obtains its intermediate model as follows\footnote{Both intermediate and global aggregation processes adopt FedAvg \cite{FedAvg}. Since the time and energy overhead of aggregation is negligible, it is excluded from the subsequent analysis, according to supportive works \cite{Blockchain} and \cite{Multi-UAV}.}:
\begin{align}
	w^{\mathsf{UAV}}_{m;[g,k]}=\sum_{n\in\mathcal{N}_{m;[g]}^\mathsf{Sel}}|\mathcal{D}_{n}|w^{\mathsf{Dev}}_{n;[g,k,H]}/|\mathcal{D}_{m;[g]}^\mathsf{Sel}|,\label{2}
\end{align}
where $w^{\mathsf{Dev}}_{n;[g,k,H]}$ represents local model parameters received form device $n\in\mathcal{N}_{m;[g]}^\mathsf{Sel}$. Besides,  $ |\mathcal{D}^{\mathsf{Sel}}_{m;[g]}|=\sum_{n\in \mathcal{N}^{\mathsf{Sel}}_{m;[g]}}|\mathcal{D}_n|$ represents the total size of datasets of all devices covered by UAV $m$. This model is then sent back to the devices covered by the UAV and is used to synchronize their local models, initiating the next round of local training. After several repetitions of the above procedure, when $k = K_{[g]}$, where the value of $K_{[g]}$ will be later optimized according to UAVs' battery levels and model performance, each active UAV $m$ sends its intermediate model $w^{\mathsf{UAV}}_{m;[g,K_{[g]}]}$ to the UAV designated as the global model aggregator, which computes the next global model $w_{[g]}$ as described next.

\noindent\textbf{\textit{(C) Global Aggregations: }}After the reception of models from UAVs at the UAV $m$ that was designated to be the global model aggregator (i.e., $X_{m;[g]} = 1$), this UAV combines the received models to form the next global model as follows:
\begin{align}
	w_{[g]}=\sum_{m\in\mathcal{M}_{[g]}}|\mathcal{D}^{\mathsf{Sel}}_{m;[g]}|w^{\mathsf{UAV}}_{m;[g,K_{[g]}]}/|\mathcal{D}_{[g]}|, \label{3}
\end{align}
where  $|\mathcal{D}_{[g]}|=\sum_{m\in\mathcal{M}_{[g]}}|\mathcal{D}^{\mathsf{Sel}}_{m;[g]}|$ indicates the combined size of datasets of all devices engaged in global aggregation $g$. Further, we consider that the training ends at global aggregation $g$ once two consecutive global models satisfy the following condition:
\begin{align}
    \Vert w_{[g]}-w_{[g-1]}\Vert \leq \delta, \label{4}
\end{align}
where $\|\cdot\|$ represents Euclidean 2-norm, and $\delta$ is a small positive value used to control the convergence criterion. If the convergence criterion is not met, $w_{[g]}$ is broadcasted across the UAVs and the next round of device operations starts. The processes that take place during our designed HFL are summarized in \textbf{Appendix A.C}.  

\subsection{Multi-Metric Device Evaluation Scores and Device-to-UAV Associations}
In dynamic and heterogeneous scenarios, D2U association is influenced by factors such as device–UAV distance and device computational capacity. More critically, prior studies \cite{MobilityAware,Convergence} show that diversity in device data distribution strongly affects global model convergence. In our setting, intermediate aggregations involving devices with more diverse data characteristics at each UAV generally promote faster and more stable global convergence. Given these considerations, quantifying devices' values/benefits across multiple dimensions is essential for optimizing D2U association. To achieve this, we propose the following fitness score between each device $n$ and UAV $m$ for each global round $g$ that quantifies the value of a device for UAV association based on three key aspects, namely, data distribution diversity, distance to the UAV, and available computing resources:
\begin{align}
	\alpha_{m,n:[g]}=\lambda_{1}S_{m,n;[g]}^\mathsf{Sim}+\lambda_{2}S_{m,n;[g]}^\mathsf{Dis}+\lambda_{3}S_{m,n;[g]}^\mathsf{Fre}, \label{5} 
\end{align}
where $\lambda_{1}, \lambda_{2}, \lambda_{3}$ are positive weighting coefficients, and $\lambda_{1}+\lambda_{2}+\lambda_{3}=1$. Besides, \( S^{\mathsf{Sim}}_{m,n;[g]} \) represents the \textit{data distribution similarity score}, evaluating the relevance and diversity of device $n$ dataset to the rest of the devices covered by UAV $m$,  \( S^{\mathsf{Dis}}_{m,n;[g]} \) denotes the \textit{distance score}, capturing the proximity of device $n$ to  UAV $m$, and \( S^{\mathsf{Fre}}_{m,n;[g]} \) measures \textit{computing resources}, reflecting the processing capability of  device $n$ relative to the rest of the devices covered by UAV $m$ in terms of its CPU frequency availability denoted by $f_n$.

Measuring data distribution differences is challenging in HFL since devices do not share private data with UAVs. To address this, we define a Kullback-Leibler Divergence (KLD)-based model difference score \(R_{m,n;[g]}\), quantifying the divergence between device \(n\)’s data and other devices covered by UAV \(m\). Each UAV \(m\) is assigned a personalized model \(\nu^{\mathsf{Per}}_m\) trained on a limited local dataset, following the edge-data assumption in \cite{NonIIDe}. Before each global aggregation, UAVs broadcast \(\nu^{\mathsf{Per}}_m\) to devices. Each device computes a KLD-based score by comparing the UAV model outputs with its local model using a small batch \(\mathcal{D}_n^\mathsf{Small}\); higher scores indicate greater divergence.  
The model difference score for device \(n\) under UAV \(m\) is thus defined as:
\begin{align}
	R_{m,n;[g]}=\sum_{j\in \mathcal{D}_n^\mathsf{Small}}\Phi(\nu_m^\mathsf{Per},x_j)\mathrm{log}\frac{\Phi(\nu_m^\mathsf{Per},x_j)}{\Phi(w^{\mathsf{Dev}}_{n; [g-1]},x_j)}, \label{6}
\end{align}
where $x_j$ represents the local sample of the $j^\text{th}$ input, $\Phi(\nu_m^\mathsf{Per},x_j)$ corresponds to the pre-softmax output of the UAV’s personalized model, and $\Phi(w^{\mathsf{Dev}}_{n; [g-1]},x_j)$ is the pre-softmax output of the local model trained on the device. Consequently, UAVs prioritize selecting devices with higher model difference scores for participation in model training as those often have the highest diversity of data.

Subsequently, to compute \eqref{5}, we normalize  {\footnotesize$R_{m,n;[g]}$, $d_{m,n;[g]}$} and {\footnotesize$f_{n}$} to obtain {\footnotesize $S_{m,n;[g]}^\mathsf{Sim}={R_{m,n;[g]}}/{R_{m;[g]}^\mathsf{Max}}$, $S_{m,n;[g]}^\mathsf{Dis}={d_{m;[g]}^\mathsf{Min}}/{d_{m,n;[g]}}$, $S_{m,n;[g]}^\mathsf{Fre}={f_{n}}/{f_{m;[g]}^\mathsf{Max}}$}, where {\footnotesize$R_{m;[g]}^\mathsf{Max}$, $d_{m;[g]}^\mathsf{Min}$, $f_{m;[g]}^\mathsf{Max}$} represent the maximum model difference, the minimum distance between UAV and device, and the maximum clock frequency of the computing processor, respectively, among all devices within the coverage area of UAV $m$ in the $g^\text{th}$ round of global iteration.

After obtaining the above parameters, to ensure that only the most relevant devices contribute to the learning process, each computed fitness score in \eqref{5} is compared against an adaptive selection threshold, denoted by $\beta_{m;[g]}$. In particular, a device $n$ under the coverage of active UAV $m$ (i.e.,   $n\in\mathcal{N}^{\mathsf{Cov}}_{m;[g]}$) is selected for participation in model training following the rule below
\begin{align}
	\begin{cases}\alpha_{m,n;[g]}\geq\beta_{m;[g]},&n\in\mathcal{N}_{m;[g]}^\mathsf{Sel},\\
		\text{otherwise},&n\notin\mathcal{N}_{m;[g]}^\mathsf{Sel},\end{cases}
\end{align}
where the adaptive threshold \( \beta_{m;[g]} \) is dynamically adjusted and later optimized in our problem formulation based on the network environment, including factors such as device mobility patterns, computational load, and communication constraints. Devices that satisfy this condition form the set of successfully selected participants under the coverage of each UAV $m$.

\subsection{Modeling of Delay and Energy Costs}
In this section, we first model the delay and energy consumption associated with model training and transfers on IoT devices, followed by the energy consumption of UAVs during hovering and information exchanges. Finally, we obtain the total energy and delay costs of our UAV-assisted HFL scenario of interest.

\subsubsection{Delay and Energy Consumption of IoT Devices}
At each global aggregation $g$, the model training latency of device $n$ during each intermediate aggregation is given by $t_n^\mathsf{Cmp}=Ht_n^\mathsf{Unit}$, where $t_n^\mathsf{Unit}$ represents the unit time taken by device $n$ to complete one round of local training, that is,
\begin{align}
	t_n^\mathsf{Unit}=t^\mathsf{Fix}+{\varphi_n c_{n}|\mathcal{D}_n|}/{f_{n}}, \label{8}
\end{align}
where $\varphi_n\in(0,1]$ captures the fraction of datapoints of the local dataset that are contained in each minibatch of SGD. Here, $c_{n}$ captures the number of required cycles of computing processors (e.g., CPU cycles) for device $n$ to process one sample data, and $t^\mathsf{Fix}$ represents a fixed time duration, capturing various factors, e.g., model training transfer between GPU and CPU. We can subsequently obtain the energy consumption of each device $n$ during each intermediate aggregation round as 
\begin{align}
	e_{n}^\mathsf{Cmp}=Hf_{n}^{2}\varphi_nc_{n}|\mathcal{D}_{n}|\vartheta_{n}/2, \label{9}
\end{align}
where $\vartheta_{n}/2$ represents the effective capacitance coefficient of the computing processor of device $n$~\cite{HFEL}.

Focusing on the model transfers between the devices and UAVs, each device $n$ experiences two types of delays: \textit{(i)} The delay when the device transmits its model to its associated UAV, which is 
$t_{n\to m;[g]}^\mathsf{D2U}=I_{n}^\mathsf{D2U}/r_{n\to m;[g]}^\mathsf{D2U}$, where $I_{n}^\mathsf{D2U}$ describes the size of local model parameters of device $n$. \textit{(ii)} The delay when the device waits to receive the intermediate model from its associated UAV, which is 
$
t_{m\to n;[g]}^\mathsf{U2D}=I_{m}^\mathsf{U2D} / r_{m\to n;[g]}^\mathsf{U2D} 
$, where $I_{m}^\mathsf{U2D}$ represents the size of edge model parameters of UAV $m$. Accordingly, the overall communication delay of each device $n$ during any of the intermediate aggregations of global round $g$ is given by
\begin{align}
    t_{n;[g]}^\mathsf{Com}=t_{m\to n;[g]}^\mathsf{U2D}+t_{n\to m;[g]}^\mathsf{D2U}. \label{10}
\end{align}

Consequently, during each intermediate model aggregation of global round $g$, the overall delay, energy consumption of model transfers and energy cost at device 
at device $n$ are
\begin{align}
    t_{n;[g]}^\mathsf{Dev}&=t_{n;[g]}^\mathsf{Cmp}+t_{n;[g]}^\mathsf{Com},\label{11}\\
	e_{n;[g]}^\mathsf{Com}&=t_{n\to m;[g]}^\mathsf{D2U}p_{n}^\mathsf{D2U}, \label{12}\\
    e_{n;[g]}^\mathsf{Dev}&=e_n^\mathsf{Cmp}+e_{n;[g]}^\mathsf{Com}.\label{13}
\end{align} 

We note that to guarantee uninterrupted communication between each device $n$ and its associated UAV during a global iteration round, it is necessary to constrain its $t_{n;[g]}^\mathsf{Dev}$ to be less than the residence time within the UAV's coverage area (i.e., $t_{n;[g]}^\mathsf{Stay}$), which will be enforced in our later optimization.

\subsubsection{Energy Consumption of UAVs}
When UAV \( m \) serves as an intermediate aggregator during the \( k^\text{th} \) intermediate aggregation of global aggregation \( g \), its energy consumption consists of two key components: \textit{(i) Hovering energy}, which depends on the time required for all IoT devices within the coverage of UAV \( m \) to successfully upload their local models. This duration is determined by the maximum upload time among the devices associated with UAV \( m \) during the \( k^\text{th} \) iteration, given by: $ t^{\mathsf{Hover}}_{m;[g,k]} = \max_{n \in \mathcal{M}^{\mathsf{Sel}}_{m;[g]}} \{t^{\mathsf{Dev}}_{n;[g]}\}
$. \textit{(ii) Broadcast energy}, which is influenced by the data transmission delay associated with U2D communication, denoted as \( t^{\mathsf{U2D}}_{m \to n;[g,k]} \). 
Subsequently, the total energy consumed by UAV \( m \) during the \( k^\text{th} \) intermediate aggregation of global aggregation \( g \) is given by
\begin{align}
	e^{\mathsf{UAV}}_{m;[g,k]} = \underbrace{t^{\mathsf{Hover}}_{m;[g,k]} \overline{p_m}}_{\text{hovering energy}} + \underbrace{t^{\mathsf{U2D}}_{m \to n;[g]} p^{\mathsf{U2D}}_m}_{\text{broadcast energy}}. \label{14}
\end{align}

In our HFL architecture, UAVs have limited energy that decreases as intermediate and global aggregations proceed, potentially causing disconnections and interrupting training. To address this, we introduce \textit{energy check rules}, where each UAV monitors its energy after every intermediate aggregation. If the remaining energy exceeds a predefined threshold, training continues; otherwise, a global aggregation is triggered to preserve the model parameters of terrestrial IoT devices before their UAVs disconnect. During each global aggregation $g$, each active UAV can assume the following roles:

\noindent
$\bullet$ \textit{\textbf{Role 1.}} The UAV functions solely as an intermediate aggregator, responsible for intermediate aggregations.

\noindent
$\bullet$ \textit{\textbf{Role 2.}} The UAV serves both as an intermediate aggregator and the global aggregator, meaning it also performs the final aggregation at the global level.

UAV energy consumption depends on its role. For instance, UAVs in Role 1 avoid the broadcast energy cost of global aggregation. If a UAV disconnects due to low energy, devices in its coverage may lose association, potentially affecting convergence. To mitigate this and maximize resource utilization, UAV positions are dynamically adjusted, though movement incurs energy costs. Accordingly, we first model UAV energy consumption during training and then define rules for global aggregation based on UAV energy levels.
During the first $\mathbb{k}^\text{th}$ (i.e., $\mathbb{k} \leq K_{[g]}$) rounds of intermediate iterations within the $g^\text{th}$ round of global iterations, the energy loss of an active UAV $m$ can be defined as:
$
	E_{m;[g]}^\mathsf{ UAV,\mathbb{k}}=\sum_{k=1}^{\mathbb{k}}e_{m;[g,k]}^\mathsf{UAV}
$.
Before the start of a new global iteration round \( g \), let the available energy of UAV \( m \) be denoted as \( E^{\mathsf{Batt}}_{m;[g]} \). To determine whether the UAV has sufficient energy to support the next round of intermediate iterations, we estimate the energy required for the upcoming \( (\mathbb{k} + 1)^\text{th} \) intermediate iteration based on the highest energy consumption observed in previous \( \mathbb{k} \) iterations. If the energy required for the next intermediate iteration exceeds the UAV’s available energy, it indicates that the UAV will disconnect during the next round of local training. This undesired scenario can be captured by the following condition:
\begin{align}
	E_{m;[g]}^\mathsf{UAV,\mathbb{k}}\leq E_{m;[g]}^\mathsf{Batt}\leq E_{m;[g]}^\mathsf{ UAV,\mathbb{k}}+\max_{1\leq k \leq \mathbb{k}}\left\{e_{m;[g,k]}^\mathsf{ UAV}\right\}. \label{15}
\end{align}

Under this condition, the total number of intermediate iterations in the \( g^\text{th} \) global iteration is \( \mathbb{k} \). However, if a UAV has sufficient energy to train continuously, infrequent global aggregations may cause model divergence due to prolonged local updates. To ensure stability, we set a maximum number of edge iterations per global aggregation, denoted as \( K^{\mathsf{Max}} \). Thus, the number of edge iterations per global iteration is given by
\begin{align}
	K_{[g]}=\begin{cases}\mathbb{k},&\phi_{[g]}=1,\\
		K^\mathsf{Max},&\phi_{[g]}=0.\end{cases}\label{16}
\end{align}
Here, \( \phi[g] = 1 \) indicates that the condition in \eqref{16} holds for at least a UAV $m$,  and thus a UAV disconnection will occur due to insufficient energy during global round $g$, limiting the edge iterations to \( \mathbb{k} \). If \( \phi[g] = 0 \), meaning no UAV disconnections, the system enforces the maximum allowable edge iterations \( K^{\mathsf{Max}} \).

\subsubsection{Total Energy and Delay Costs}
Next, we discuss the time and energy costs of the intermediate and global aggregation processes. During the $g^{\text{th}}$ global iteration, the time and energy costs at the local edge network rooted at UAV $m$ (i.e., each UAV $m$ and all its subsequent assigned devices) can be computed as follows:
\begin{align}
	T_{m;[g]}^\mathsf{Edge}&=\sum_{k=1}^{K_{[g]}}t_{m;[g,k]}^\mathsf{Hover}, \\
	E_{m;[g]}^\mathsf{Edge}&=\sum_{k=1}^{K_{[g]}}\left\{e_{m;[g,k]}^\mathsf{ UAV}+\sum_{n\in\mathcal{N}_{m;[g]}^\mathsf{Sel}}e_{n;[g]}^\mathsf{Dev}\right\}.
\end{align}


Also, given the (potential)  UAV relocations across two consecutive global aggregations, the total hovering energy consumption is formally defined as a function of delay and the traveling distance of each UAV $m$ as follows:
\begin{align}
	&\hspace{-3mm}E_{m;[g]}^\mathsf{Delay}=T_{m;[g]}^\mathsf{E2G}\overline{p_m}+\overline{p_m}^\mathsf{Move}d_{m;[g]}/{V_m}, \\
    &\hspace{-3mm}T_{m;[g]}^\mathsf{Delay}=T_{m;[g]}^\mathsf{E2G}+d_{m;[g]}/{V_m},\\
    &\hspace{-3mm}d_{m;[g]}{=}\sqrt{\hspace{-0.45mm}(\mathtt{p}_{m;{[g]}}\hspace{-0.55mm}(x){-}\mathtt{p}_{m;[g-1]}\hspace{-0.45mm}(x)\hspace{-0.45mm})^2+(\mathtt{p}_{m;{[g]}}\hspace{-0.45mm}(y){-}\mathtt{p}_{m;{[g-1]}}\hspace{-0.45mm}(y)\hspace{-0.45mm})^2},\hspace{-3mm}
\end{align}
where $\overline{p_m}^\mathsf{Move}$ represents the average power consumption of the UAV during movement, $T_{m;[g]}^\mathsf{Delay}$ represents the delay in this process, $T_{m;[g]}^\mathsf{E2G}=I_{m}^\mathsf{U2U}/r_{m\to m'}^\mathsf{U2U}$ captures the delay of model offloading from UAV $m$ to the global aggregator UAV $m'$ (i.e., $X_{m';[g]}=1$), and $d_{m;[g]}$ represents the traveling distance between the designated locations of the UAV across two consecutive global aggregations.

Once the global aggregation is completed, the global aggregator UAV (we denote it by $m'$ for analytical simplicity) broadcasts the updated global model to all UAVs that subsequently relay it to their assigned devices, initiating the next global round. The broadcast time and energy required for this process are given by
\begin{align}
    & T^{\mathsf{Broad}}_{[g]}= \sum_{m'\in\mathcal{M}}X_{m';[g]}\bigg( \max_{m\in\mathcal{M}_{[g]}\setminus\{m'\}}\Big\{ I^{\mathsf{G}}/r^{\mathsf{U2U}}_{m'\to m} \notag\\[-.5em]
    & ~~~~~~~~~~~~~~~~~~~~~~~~~~~~~~~~~~~\quad +\max_{n\in\mathcal{N}^{\mathsf{Sel}}_{m;[g]}}\{I^{\mathsf{G}}/r^{\mathsf{U2D}}_{m\to n}\} \Big\} \bigg), \\
    & E^{\mathsf{Broad}}_{[g]}= \sum_{m'\in\mathcal{M}} X_{m';[g]}\bigg( \max_{m\in\mathcal{M}_{[g]}\setminus\{m'\}}\Big\{ I^{\mathsf{G}}/r^{\mathsf{U2U}}_{m'\to m} \Big\}p_{m'}^{\mathsf{U2U}} \notag\\[-.5em]
    & ~~~~~~~~~~~~~~~~~~~~~~~\quad +\sum_{m\in\mathcal{M}_{[g]}}\max_{n\in\mathcal{N}^{\mathsf{Sel}}_{m;[g]}}\Big\{I^{\mathsf{G}}/r^{\mathsf{U2D}}_{m\to n}\Big\}p_{m}^{\mathsf{U2D}}  \bigg),
\end{align}
where $I^\mathsf{G}$ represents the size of global model. Besides, the hovering energy required by UAVs participating in the next global round is defined as
\begin{align}
	E_{[g]}^\mathsf{Bwait}=&\sum_{m\in\mathcal{M}_{[g]}}T_{[g]}^{\mathsf{Broad}}\overline{p_m}.
\end{align} 

Accordingly, the overall time and energy costs during the $g^{\text{th}}$ global round can be calculated as follows:
\begin{align}
	T_{[g]}&=T_{[g]}^\mathsf{Broad}+\max_{m\in \mathcal{M}_{[g]}}\left\{T_{m;[g]}^\mathsf{Edge}+T_{m;[g]}^\mathsf{Delay}\right\}, \label{28}\\
    E_{[g]}&=E_{[g]}^\mathsf{Broad}+E_{[g]}^\mathsf{Bwait}+\sum_{m\in\mathcal{M}_{[g]}}\left\{E_{m;[g]}^\mathsf{Edge}+E_{m;[g]}^\mathsf{Delay}\right\}. \label{29}
\end{align}

\section{Problem Formulation}
We formalize the network orchestration in our HFL scenario of interest as an optimization problem $\mathcal{P}_0$, where its optimization variables include the number of local SGD iterations $H$, the upload bandwidth of D2U links $B_{m,n:[g]}^\mathsf{D2U}$, the download bandwidth of U2D links $B_{m,n:[g]}^\mathsf{U2D}$, D2U association adaptive threshold $\beta_{m;[g]}$, the location $\mathtt{p}_{m;[g]}$ of active UAVs, and the selection of the global aggregator $X_{m;[g]}$ in each round of global iteration. This optimization problem is given by
\begin{align}
	\hspace{-4mm}\bm{\mathcal{P}}_0:\underset{{H}, {B_{m,n;[g]}^\mathsf{D2U}}, {B_{m,n;[g]}^\mathsf{U2D}}, {\beta_{m;[g]}}, {\mathtt{p}_{m;[g]}}, {X_{m;[g]}}}{\hspace{-4mm}\operatorname{argmin}}\hspace{-4mm}\lambda_4 E_{[g]}+\lambda_5 T_{[g]}
\end{align}
\vspace{-0.5mm} 
\begin{subequations}\label{p0}{
		\begin{align*}
			\text{s.t.}~~~~&
			0\leq\sum_{n \in \mathcal{N}_{m;[g]}^\mathsf{Sel}} B_{m,n;[g]}^\mathsf{D2U}\leq B_m^\mathsf{D2U},~~~\forall m\in\mathcal{M}_{[g]}\tag{27a} \label{27a}\\
            &0\leq\sum_{n \in \mathcal{N}_{m;[g]}^\mathsf{Sel}} B_{m,n;[g]}^\mathsf{U2D}\leq B_m^\mathsf{U2D},~~~\forall m\in\mathcal{M}_{[g]}\tag{27b} \label{27b} \\
			&\mathcal{N}_{m;[g]}^\mathsf{Sel}\cap\mathcal{N}_{m';[g]}^\mathsf{Sel}=\emptyset,~~~\forall m,m'\in\mathcal{M}_{[g]},m\neq m'\tag{27c} \label{27c}\\
                &E_{m;[g]}^\mathsf{UAV,\mathbb{k}}\leq E^{\mathsf{Batt}}_{m;[g]} \leq E^{\mathsf{Batt}}_m,~~~\forall m\in\mathcal{M}_{[g]}\tag{27d} \label{27d}\\
                &\mathcal{N}^\mathsf{Sel}_{m;[g]} \subset \mathcal{N}^\mathsf{Cov}_{m;[g]},~~~\forall m\in\mathcal{M}_{[g]}\tag{27e} \label{27e}\\
                &t_{n;[g]}^\mathsf{Dev}\leq t_{n;[g]}^\mathsf{Stay},~~~\forall n\in\mathcal{N}_{m;[g]}^\mathsf{Sel}\tag{27f} \label{27f}\\
			&\alpha_{m,n;[g]}\geq\beta_{m;[g]},~~~\forall n\in\mathcal{N}_{m;[g]}^\mathsf{Sel}\tag{27g} \label{27g}\\
		&H\in\mathbb{N}^+,\alpha_{m,n;[g]},\beta_{m;[g]},X_{m;[g]}\in[0,1]\tag{27h} 
        \label{27h}\\
        &\sum_{m\in\mathcal{M}_{[g]}} X_{m;[g]}=1, \tag{27i}\label{27i}
	\end{align*}}	
\end{subequations}
\hspace{-0.45em}where $\lambda_4,~\lambda_5 \in [0,1]$ weight time and energy. Constraints \eqref{27a}–\eqref{27b} limit per-UAV bandwidth ($B_m^\mathsf{D2U}$ and $B_m^\mathsf{U2D}$). Constraint \eqref{27c} avoids redundant transfers, while \eqref{27d}–\eqref{27e} bound UAV energy and device coverage. Constraint \eqref{27f} ensures model training and upload within device dwell time. Constraints \eqref{27g}–\eqref{27h} define feasible regions, and \eqref{27i} enforces one UAV as global aggregator per round.
\begin{figure}[htbp]
	\centering
	\includegraphics[trim=0cm 0cm 0cm 0cm, clip, width=0.9\columnwidth]{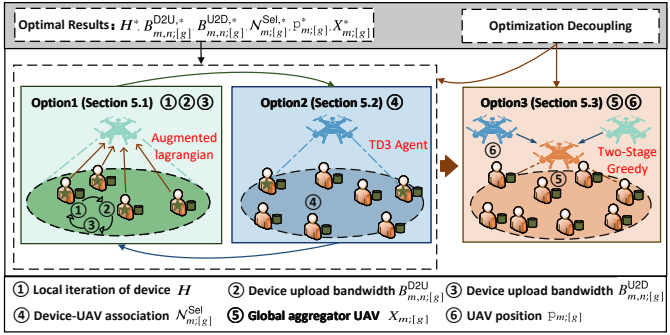}
	\caption{An overview of our decoupling framework for solving  problem $\mathcal{P}_0$.}
\end{figure}

\section{Problem Decomposition and Solution Design}
\( \mathcal{P}_0 \) presents a mixed-integer nonlinear programming problem with NP-hardness~\cite{Cost-Efficient, JointEdge}, requiring optimization over continuous variables \( B^{\mathsf{D2U}}_{m,n;[g]} \), \( B^{\mathsf{U2D}}_{m;[g]} \), adaptive threshold \( \beta_{m;[g]} \), UAV positions \( \mathtt{p}_{m;[g]} \), and discrete variables such as global aggregator selection \( X_{m;[g]} \) and the number of local iterations \( H \). Its objective function is non-convex, making direct solutions intractable and prone to convergence issues (e.g., RL-based methods).  
Following the decomposition paradigm in \cite{Optimization, federatedp, HiFlash}, we divide \( \mathcal{P}_0 \) into three subproblems:

\noindent
$\bullet$ \( \mathcal{P}_1 \): optimizes \( H \) (relaxed as continuous), \( B^{\mathsf{D2U}}_{m,n;[g]} \), and \( B^{\mathsf{U2D}}_{m;[g]} \) to balance local computation and communication.  

\noindent
$\bullet$ \( \mathcal{P}_2 \): optimizes adaptive D2U association via \( \beta_{m;[g]} \) to select devices that most effectively contribute to convergence.

\noindent
$\bullet$ \( \mathcal{P}_3 \): optimizes UAV positions \( \mathtt{p}_{m;[g]} \) and global aggregator selection \( X_{m;[g]} \).  

The solution proceeds in two stages. First, \( \mathcal{P}_1 \) and \( \mathcal{P}_2 \) are solved iteratively: starting from an initial D2U association, solve \( \mathcal{P}_1 \), update the association via \( \mathcal{P}_2 \), and repeat until convergence, jointly minimizing iteration and communication costs. Second, with \( \mathcal{P}_1 \) and \( \mathcal{P}_2 \) fixed, \( \mathcal{P}_3 \) is solved to refine UAV positions, maximize covered devices \( \mathcal{N}^{\mathsf{Cov}}_{m;[g]} \), and reduce inter-UAV communication. This decomposition yields an efficient and scalable solution, as illustrated in Fig.~2.
\subsection{Problem Formulation and Solution Design of $\mathcal{P}_1$}

Since a single global round consists of multiple intermediate iterations, each coordinated by multiple UAVs, we define subproblem \( \mathcal{P}_1 \) as the optimization of the number of local iterations \( H \), which is relaxed and considered to be a continuous variable, uplink bandwidth allocation \( B^{\mathsf{D2U}}_{m,n;[g]} \) and downlink bandwidth allocation \( B^{\mathsf{U2D}}_{m;[g]} \) for each UAV $m$ and its assigned devices. The objective of \( \mathcal{P}_1 \) is to minimize the time and energy costs incurred by active each UAV (e.g., UAV \( m \)) during each intermediate aggregation. Accordingly, subproblem \( \mathcal{P}_1 \) is formulated as follows:
\setcounter{equation}{27}
\begin{multline}
	\bm{\mathcal{P}}_1:\underset{{H},{B_{m,n;[g]}^\mathsf{D2U}}, {B_{m;[g]}^\mathsf{U2D}}}{\operatorname{argmin}}~~ \lambda_{4}\left\{\sum_{n\in\mathcal{N}_{m;[g]}^\mathsf{Sel}}e_{n;[g]}^\mathsf{Dev}+e_{m;[g,k]}^\mathsf{UAV}\right\}
	+\lambda_{5}t^\mathsf{Hover}_{m;[g,k]}
\end{multline}
\vspace{-5.5mm} 
\begin{subequations}\label{p1}{
		
		\begin{align}
			\text{s.t.}~~~~
			&\eqref{27a}, \eqref{27b}, \eqref{27h} \notag
	\end{align}}
\end{subequations}

Recalling our previous discussions and \eqref{8}-\eqref{14}, $\mathcal{P}_1$ can be further rewritten as the following optimization problem:
\setcounter{equation}{28}
\begin{multline}
	\bm{\mathcal{P}}_{1a}:\underset{{H},{B_{m,n;[g]}^\mathsf{D2U}}, {B_{m;[g]}^\mathsf{U2D}}}{\operatorname{argmin}}~~\lambda_{4}\left\{\sum_{n\in\mathcal{N}_{m;[g]}^\mathsf{Sel}}Hf_{n}^2\varphi_nc_n|\mathcal{D}_n|\theta_n/2+\right. \\t^\mathsf{D2U}_{n\rightarrow m;[g]}p^\mathsf{D2U}_n 
	\left. +t^\mathsf{U2D}_{m\rightarrow n;[g]}p^\mathsf{U2D}_m+\max_{n\in\mathcal{N}_{m;[g]}^\mathsf{Sel}}(t_{n;[g]}^\mathsf{Dev})\overline{p_m}\right\}\\
    +\lambda_5\max_{n\in\mathcal{N}_{m;[g]}^\mathsf{Sel}}(t_{n;[g]}^\mathsf{Dev})
\end{multline}
\vspace{-9mm} 
\begin{subequations}\label{p2}{
		
		\begin{align}
			\text{s.t.}~~~~
			&\text{\eqref{27a}}, \text{\eqref{27b}}, \text{\eqref{27h}} \notag
	\end{align}}
\end{subequations}

To facilitate finding the solution of $\mathcal{P}_{1a}$, we further introduce the following (positive) variables for notational simplicity: 
\begin{align}
	~~A_n^\mathsf{D2U}&=\lambda_4I_{n}^\mathsf{D2U}p_n^\mathsf{D2U},~\mathcal{A}_{m,n}^\mathsf{D2U}=p^{\mathsf{D2U}}_n d_{m,n;[g]}^{-\alpha_{\mathsf{D2U}}}/N_0,\notag \\
	A_m^\mathsf{U2D}&=\lambda_4I_{m}^\mathsf{U2D}p_m^\mathsf{U2D},~\mathcal{A}_{m,n}^\mathsf{U2D}=p^{\mathsf{U2D}}_m d_{m,n;[g]}^{-\alpha_{\mathsf{U2D}}}/N_0,\notag \\
	U_{m,n}^\mathsf{D2U}&=(\lambda_4\overline{p_m}+\lambda_5)I_{n}^\mathsf{D2U}p_n^\mathsf{D2U},~
	U_{m,n}^\mathsf{U2D}=(\lambda_4\overline{p_m}+\lambda_5)I_{m}^\mathsf{U2D}p_m^\mathsf{U2D},\notag \\
        Z_{m,n}&=(\lambda_4\overline{p_m}+\lambda_5)\left\{\frac{|\mathcal{D}_n|\varphi_nc_n}{f_n}+t^\mathsf{Fix}\right\},~\notag 
	C_n=\frac{\lambda_4f_n^2\varphi_nc_n|\mathcal{D}_n|\vartheta_n}2. \notag 
\end{align}

\setcounter{equation}{29}
\begin{figure*}[t]
{\footnotesize
\begin{align}\bm{\mathcal{P}}_{1b}:\underset{{H},{B_{m,n;[g]}^\mathsf{D2U}},{B_{m;[g]}^\mathsf{U2D}}}{\operatorname{argmin}}~~\sum_{n\in\mathcal{N}_{m;[g]}^\mathsf{Sel}}\left\{\frac{A_n^\mathsf{D2U}}{B_{m,n;[g]}^\mathsf{D2U}\text{log}_2(1+\frac{\mathcal{A}_{m,n}^\mathsf{D2U}}{B_{m,n;[g]}^\mathsf{D2U}})}+
    \frac{A_m^\mathsf{U2D}}{B_{m,n;[g]}^\mathsf{U2D}\text{log}_2(1+\frac{\mathcal{A}_{m,n}^\mathsf{U2D}}{B_{m,n;[g]}^\mathsf{U2D}})}+HC_n\right\}+ \notag
	\\ \max_{n\in\mathcal{N}_{m;[g]}^\mathsf{Sel}}\left\{\frac{U_n^\mathsf{D2U}}{B_{m,n;[g]}^\mathsf{D2U}\text{log}_2(1+\frac{\mathcal{A}_{m,n}^\mathsf{D2U}}{B_{m,n;[g]}^\mathsf{D2U}})}+
    \frac{U_m^\mathsf{U2D}}{B_{m,n;[g]}^\mathsf{U2D}\text{log}_2(1+\frac{\mathcal{A}_{m,n}^\mathsf{U2D}}{B_{m,n;[g]}^\mathsf{U2D}})}+HZ_{m,n}\right\}. \label{30}
\end{align}}

\vspace{-2mm} 
\begin{subequations}\label{p3}{
		
		\begin{align}
			\text{s.t.}~~~~
			&\text{\eqref{27a}}, \text{\eqref{27b}}, \text{\eqref{27h}} \notag
	\end{align}}
\end{subequations}
\vspace{-7.5mm} 

\noindent\rule{\textwidth}{0.4pt}
\vspace{-10mm} 
\end{figure*}

Accordingly, $\mathcal{P}_{1a}$ can be further transformed into $\mathcal{P}_{1b}$ given by \eqref{30}, for which we obtain the following result.

\begin{theorem} $\mathcal{P}_{1b}$ represents a convex optimization problem.
\end{theorem} 
\vspace{-1mm}
\noindent \textit{Proof:} Please see \textbf{Appendix B.A}. 

Given the convexity of subproblem \( \mathcal{P}_{1b} \), we adopt a penalty-based approach and transform it into an augmented Lagrangian function to facilitate optimization. This transformation allows us to effectively handle constraints while improving the stability of the solution. In particular, to determine an optimal value for the number of local iterations \( H \), while keeping other variables (e.g., bandwidth allocation \( B^{\mathsf{D2U}}_{m,n;[g]} \), \( B^{\mathsf{U2D}}_{m,n;[g]} \) and adaptive threshold \( \beta_{m;[g]} \)) fixed, we introduce the following two constraints:
\setcounter{equation}{30}
\begin{align}
	\mathcal{G}(H) = \max_{n\in\mathcal{N}_{m;[g]}^\mathsf{Sel}}\left\{\frac{U_n^\mathsf{D2U}}{B_{m,n;[g]}^\mathsf{D2U}\text{log}_2(1+\frac{\mathcal{A}_{m,n}^\mathsf{D2U}}{B_{m,n;[g]}^\mathsf{D2U}})}+\notag \right.\\ \left. \frac{U_m^\mathsf{U2D}}{B_{m,n;[g]}^\mathsf{U2D}\text{log}_2(1+\frac{\mathcal{A}_{m,n}^\mathsf{U2D}}{B_{m,n;[g]}^{U2D}})}+HZ_{m,n}\right\},  \label{31} \\
   \hspace{-3cm}\mathcal{Y} \geq \mathcal{G}(H), \label{32}
\end{align}
where $\mathcal{Y}$ indicates a slack variable. Combining \eqref{31} and \eqref{32}, we rewrite the objective function of problem $\mathcal{P}_{1b}$ as $f(H)$
\begin{align}
	f(H)= \sum_{n\in\mathcal{N}_{m;[g]}^\mathsf{Sel}}\left\{\frac{A_n^\mathsf{D2U}}{B_{m,n;[g]}^\mathsf{D2U}\text{log}_2(1+\frac{\mathcal{A}_{m,n}^\mathsf{D2U}}{B_{m,n;[g]}^\mathsf{D2U}})}\right.\notag\\ \left. +
    \frac{A_m^\mathsf{U2D}}{B_{m,n;[g]}^\mathsf{U2D}\text{log}_2(1+\frac{\mathcal{A}_{m,n}^\mathsf{U2D}}{B_{m,n;[g]}^\mathsf{U2D}})}+HC_n\right\}+\mathcal{Y}. \label{33}
\end{align}

To effectively handle the constraints in subproblem \( \mathcal{P}_{1b} \), we introduce Lagrangian relaxation and construct an augmented Lagrangian function as follows: 
\begin{align}
	\mathcal{L}(H,\mathcal{Y},\upsilon)&=f(H)+\upsilon(\mathcal{G}(\text{H})-\mathcal{Y}), \label{34}
\end{align}
where \( \upsilon \) is the Lagrange multiplier. To further improve convergence, we augment the Lagrangian function by adding a quadratic penalty term as follows:
\begin{align}
    \mathcal{L}_\sigma(H,\mathcal{Y},\upsilon)&=f(H)+\upsilon(\mathcal{G}(H)-\mathcal{Y})+\frac{\sigma}{2}P(H,\mathcal{Y}), 
\end{align}
where \( P(H, \mathcal{Y}) = (\mathcal{G}(H) - \mathcal{Y})^2 \) is the penalty function, and \( \sigma \) is the penalty factor controlling the weight of constraint violations. The optimization problem is then solved though a series of iterations, where in the \( j^\text{th} \) iteration (denoted by super-script $\langle j\rangle$ in the notations) the problem is formulated as:
\begin{align}
	\bm{\mathcal{P}}_{1c}:\underset{{H^{\langle j\rangle},\mathcal{Y}^{\langle j\rangle}}}{\operatorname{argmin}}~~\mathcal{L}_{\sigma^{\langle j\rangle}}(H^{\langle j\rangle},\mathcal{Y}^{\langle j\rangle},\upsilon^{\langle j\rangle})
\end{align}
\vspace{-8mm} 
\begin{subequations}\label{p4}{
		
		\begin{align}
			\text{s.t.}~~~~
			&\text{\eqref{27h}, \eqref{32}} \notag
	\end{align}}
\end{subequations}

\noindent Here $\upsilon^{\langle j\rangle}$ is regarded as a constant (its update will be discussed later). Since solving for both \( H^{\langle j \rangle} \) and \( \mathcal{Y}^{\langle j \rangle} \) simultaneously is challenging, we employ an iterative approach, first solving for \( \mathcal{Y}^{\langle j \rangle} \) and then updating \( H^{\langle j+1 \rangle} \). In particular, given a fixed \( H^{\langle j \rangle} \), the subproblem related to \( \mathcal{Y}^{\langle j \rangle} \) can be written as
\setcounter{equation}{36}
\begin{align}
	\bm{\mathcal{P}}_{1d}:\underset{{\mathcal{Y}^{\langle j\rangle}}}{\operatorname{argmin}}~~\upsilon^{\langle j\rangle}(\mathcal{G}(H^{\langle j\rangle})-\mathcal{Y}^{\langle j\rangle})+\frac{\sigma^{\langle j\rangle}}{2}(\mathcal{G}(H^{\langle j\rangle})-\mathcal{Y}^{\langle j\rangle})^2 
\end{align}
\vspace{-9mm} 
\begin{subequations}\label{p5}{
		
		\begin{align*}
			\text{s.t.}~~~~
			&\text{\eqref{32}}
	\end{align*}}
\end{subequations}
For which, we obtain the following result.

\begin{theorem} The value of $\mathcal{Y}^{\langle j\rangle}=\max\left\{-\frac{\upsilon^{\langle j\rangle}}{\sigma^{\langle j\rangle}}-\mathcal{G}(H^{\langle j\rangle}),0\right\}$ is the optimal solution for $\mathcal{P}_{1d}$. 
\end{theorem}
\vspace{-1mm}
\noindent \textit{Proof:} Please refer to \textbf{Appendix B.B}. 

Substituting the above result back into \( \mathcal{L}_{\sigma^{\langle j \rangle}} \), we obtain
\setcounter{equation}{37}
\begin{multline}
	\mathcal{L}_{\sigma^{\langle j\rangle}}(H^{\langle j+1\rangle})=f(H^{\langle j+1\rangle})+\frac{\sigma^{\langle j\rangle}}2\mathcal{G}(H^{\langle j+1\rangle})^2-\\
	\frac{\sigma^{\langle j\rangle}}2\left\{\max\left\{\frac{\upsilon^{\langle j\rangle}}{\sigma^{\langle j\rangle}}+\mathcal{G}(H^{\langle j+1\rangle}),0\right\}^2-(\frac{{\upsilon^{\langle j\rangle}}}{\sigma^{\langle j\rangle}})^2\right\}.
\end{multline}
Thus, $\mathcal{P}_{1c}$ can further be transformed into the following problem:
\begin{align}
	\bm{\mathcal{P}}_{1e}:\underset{{H^{\langle j+1\rangle}}}{\operatorname{argmin}}~~\mathcal{L}_{\sigma^{\langle j\rangle}}(H^{\langle j+1\rangle})
\end{align}
\vspace{-8mm} 
\begin{subequations}\label{p6}{
		
		\begin{align*}
			\text{s.t.}~~~~
			&\text{\eqref{27h}} \notag
	\end{align*}}
\end{subequations}

\setcounter{equation}{39}
\begin{figure*} 
{\footnotesize
\begin{align}
	\nabla_{H^{\langle j+1\rangle}} \mathcal{L}_{\sigma^{\langle j\rangle}} &= \nabla f(H^{\langle j+1\rangle}) + \nabla \left( \frac{\sigma^{\langle j\rangle}}{2} \mathcal{G}(H^{{\langle j+1\rangle}})^2 \right) - \nabla \left( \frac{\sigma^{\langle j\rangle}}{2} \left( \max \left\{\frac{\upsilon^{\langle j\rangle}}{\sigma^{\langle j\rangle}} + \mathcal{G}(H^{\langle j+1\rangle}), 0 \right\}\right) ^2 \right) \notag\\
	 &= \nabla f(H^{\langle j+1\rangle}) + \sigma^{\langle j\rangle} \mathcal{G}(H^{\langle j+1\rangle}) \nabla \mathcal{G}(H^{\langle j+1\rangle}) + \sigma^{\langle j\rangle} \cdot \max \left\{ \hat{u}(H^{\langle j+1\rangle}), 0 \right\} \cdot \nabla \hat{u}(H^{\langle j+1\rangle}) \cdot \mathbb{I}(\hat{u}(H^{\langle j+1\rangle}))\notag\\
	 &= \nabla f(H^{\langle j+1\rangle}) + \sigma^{\langle j\rangle} \mathcal{G}(H^{\langle j+1\rangle}) Z_{m,n^*} - \sigma^{\langle j\rangle} \cdot \max \left\{ \hat{u}(H^{\langle j+1\rangle}), 0 \right\} Z_{m,n^*} \cdot \mathbb{I}(\hat{u}(H^{\langle j+1\rangle})) \notag\\
	 & = \nabla f(H^{\langle j+1\rangle}) + \sigma^{\langle j\rangle} \left( \mathcal{G}(H^{\langle j+1\rangle}) - \max \left\{ \hat{u}(H^{\langle j+1\rangle}), 0 \right\} \right) Z_{m,n^*}, (\text{if}~\mathbb{I}(\hat{u}(H^{\langle j+1\rangle}))=1) \label{40}\\
	 H^{\langle j+1\rangle} &= H^{\langle j\rangle} - \hat{\eta} \nabla_{H^{\langle j+1\rangle}} \mathcal{L}_{\sigma^{\langle j\rangle}}; 
     Z_{m,n^*} ~\text{represents the largest}~Z_{m,n}~\text{in}~n\in \mathcal{N}^\mathsf{Sel}_{m;[g]}. \label{41}
\end{align}
}
\vspace{-7mm}

\noindent\rule{\textwidth}{0.4pt}
\vspace{-8mm}
\end{figure*}
Since $\mathcal{Y}^{\langle j\rangle}$ is eliminated, the above problem is  an unconstrained optimization problem in a lower-dimensional space, enabling efficient solutions via gradient descent. Specifically, we first obtain the gradient of $H^{\langle j+1\rangle}$ with respect to $\mathcal{P}_{1e}$ according to \eqref{40} and \eqref{41}\footnote{To simplify the expressions, we define  $\hat{u}(H^{\langle j+1\rangle})=\frac{\upsilon^{\langle j\rangle}}{\sigma^{\langle j\rangle}} + \mathcal{G}(H^{\langle j+1\rangle})$. In addition, $\mathbb{I}(\hat{u}(H^{\langle j+1\rangle}))$ is defined as the indicator function, where $\mathbb{I}(\hat{u}(H^{\langle j+1\rangle}))=1$ if $\hat{u}(H^{\langle j+1\rangle}) >0$; otherwise, $\mathbb{I}(\hat{u}(H^{\langle j+1\rangle}))=0$.}, and then determine whether its gradient norm $\|\nabla_{H^{\langle j+1\rangle}}L_{\sigma^{\langle j\rangle}}(H^{\langle j+1\rangle},\mathcal{Y}^{\langle j\rangle})\|_2$ meets the  precision constants $\kappa^{\langle j\rangle}$ to prove the optimality of the solution (i.e., $\|\nabla_{H^{\langle j+1\rangle}}L_{\sigma^{\langle j\rangle}}(H^{\langle j+1\rangle},\mathcal{Y}^{\langle j\rangle})\|_2 \leq \kappa^{\langle j\rangle}$). To further ensure the adherence to constraints, we follow three steps to determine whether the iterations can stop or has to continue, which can be found in \textbf{Appendix B.C}. Besides, we obtain the following theorem to show the optimality of our proposed method.

\begin{theorem} The local minimum solution obtained by augmented Lagrangian function matches the optimal solution of the objective function $f(H)$. 
\end{theorem}
\noindent \textit{Proof:} Please refer to \textbf{Appendix B.D}. 

To obtain the solution for bandwidth allocation variables, i.e., optimizing \( B^{\mathsf{D2U}}_{m,n;[g]} \) and \( B^{\mathsf{U2D}}_{m,n;[g]} \), we need to follow the exact same procedure detailed above with $H$ being replaced by these variables in the above formulations. Furthermore, the above process is formally outlined in Alg. 2 (see \textbf{Appendix B.E}).
\subsection{Problem Formulation and Solution Design of $\mathcal{P}_2$}
Subproblem $\mathcal{P}_2$ aims to determine the D2U association during each global round, considering the mobility and heterogeneity of devices. The goal is to optimize the selection of participating devices to maximize training efficiency while ensuring reliable convergence. To achieve this, we define two key metrics for each UAV $m$ during the $g^{\text{th}}$ global iteration. The first metric, $\varpi_{m;[g]}^{1}$ captures the difference in training loss between the previous global iteration $(g-1)$ and the current iteration $g$, measuring the improvement in model optimization. The second metric, $\varpi_{m;[g]}^{2}$ evaluates the change in training accuracy, assessed using a small batch of  data (by using the parameters of models such as $w_{m;[g-1,K_{[g-1]}]}^\mathsf{UAV}$ and $w_{m;[g,K_{[g]}]}^\mathsf{UAV}$). These metrics are formally defined as follows:
\begin{align}
	\varpi_{m;[g]}^{1}&=L(w_{m;[g-1,K_{[g-1]}]}^\mathsf{UAV})-L(w_{m;[g,K_{[g]}]}^\mathsf{UAV}),\\
	\varpi_{m;[g]}^{2}&=Acc_{m;[g,K_{[g]}]}-Acc_{m;[g-1,K_{[g-1]}]},
\end{align}
where $L(w_{m;[g,K_{[g]}]}^\mathsf{UAV})$ and $Acc_{m;[g,K_{[g]}]}$ represent the training loss and accuracy of the model of UAV $m$. 

To balance time and energy efficiency while ensuring training accuracy, devices that are distant or computationally limited are avoided, even if their model similarity is high. A time constraint ensures each device's training satisfies $t_{n;[g]}^\mathsf{Dev} \leq t^{\mathsf{Max}}_{n}$, where $t^{\mathsf{Max}}_n$ is the deadline. The intermediate model’s contribution to the global model is a weighted combination of $\varpi_{m;[g]}^{1}$ and $\varpi_{m;[g]}^{2}$, leading to the following optimization problem:
\begin{align}
	\bm{\mathcal{P}}_2:\underset{\beta_{m;[g]}}{\operatorname{argmax}}~~ \lambda_{6}\varpi_{m;[g]}^{1}+\lambda_{7}\varpi_{m;[g]}^{2}
\end{align}
\vspace{-6mm} 
\begin{subequations}\label{p7}{
		
		\begin{align*}
			\text{s.t.}~~~~
			&t_{n;[g]}^\mathsf{Dev}\leq  t^{\mathsf{Max}}_{n} \tag{44a} \label{44a} 
	\end{align*}}
\end{subequations}
\hspace{-0.5em}where $\lambda_{6}$ and $\lambda_{7}$ represent the  weighting coefficients, $\lambda_{6},\lambda_{7}\in [0,1]$. The values of these coefficients can be adjusted to reflect different preference settings, thereby adapting the device selection strategy based on the dataset characteristics.

Solving \( \mathcal{P}_2 \) directly is challenging (see analysis in \textbf{Appendix C.A}). To effectively capture these dynamic dependencies, we reformulate the problem as a constrained Markov Decision Process (MDP). In this framework, state transitions are driven by the evolving performance of the UAV models and the D2U associations across multiple global rounds. By leveraging MDP-based modeling, we enable a more adaptive and learning-driven solution, where device selection strategies can be dynamically adjusted based on both historical performance trends and real-time system conditions.

\noindent \textbf{\textit{Problem Transformation using MDP:}} To determine the adaptive threshold \( \beta_{m;[g]} \) in each round of global iteration, we model the problem as an MDP, defined by $(\vmathbb{s},\vmathbb{a},\vmathbb{r},\gamma)$, with its elements defined as follows:

\noindent\textit{(I) State ($\vmathbb{s}$):} A direct indicator of UAV-aggregated edge model performance is its prediction accuracy and training loss evaluated on mini-batch datasets of global test data. Therefore, the state for UAV 
$m$, which is later fed to the TD3 agent, is defined as: $\vmathbb{s}_{m;[g]}=[L({w^\mathsf{UAV}_{m;[g,K_{[g]}]}}), Acc_{m;[g,K_{[g]}]}]$.

\noindent\textit{(II) Action ($\vmathbb{a}$):} Once the state is observed, the TD3 agent aims to change the state of MDP via determining the adaptive threshold, which dictates the selection of devices for model training. The action is thus defined as: $\vmathbb{a}_{m;[g]}=[\beta_{m;[g]}]$.

\noindent\textit{(III) Reward ($\vmathbb{r}$):} The reward function evaluates the contribution of selected devices to global model convergence based on the UAV-aggregated model’s performance. It quantifies the improvement in loss and accuracy, guiding the TD3 agent toward better exploration and is defined as
\setcounter{equation}{44}
\begin{align}
    \vmathbb{r}_{m;[g]}(\vmathbb{s},\vmathbb{a})=\lambda_{6}\varpi^1_{m;[g]}+\lambda_{7}\varpi^{2}_{m;[g]}.
\end{align}

\noindent \textbf{MDP-Based Optimization of Device Selection:} To maximize the cumulative reward across global iterations, we reformulate $\mathcal{P}_2$ as the following constrained MDP problem:
\begin{align}
	\bm{\mathcal{P}}_{2a}:\underset{\{\beta_{m;[g]}\}_{g=1}^G}{\operatorname{argmax}}~~\sum_{g=1}^{G}\gamma^{g-1} \vmathbb{r}_{m;[g]}(\vmathbb{s},\vmathbb{a})
\end{align}
\vspace{-6mm} 
\begin{subequations}\label{p8}{
		
		\begin{align*}
			\text{s.t.}~~~~
			&\eqref{44a}, \notag
	\end{align*}}
\end{subequations}
\hspace{-0.5em}where $\gamma^{g-1}$ is the discount factor ensuring long-term reward maximization and $G$ represents the number of global iterations required to reach \eqref{4}. Note that the actions and states of $\mathcal{P}_{2a}$ are in high-dimensional space and the environment has certain dynamics, which is difficult to solve directly. To address this challenge, we introduce a penalty term in the reward function, transforming the problem  into the following form:
\setcounter{equation}{46}
\begin{align} 
    \bm{\mathcal{P}}_{2b}:\underset{\{\beta_{m;[g]}\}_{g=1}^G}{\operatorname{argmax}}~~ \sum_{g=1}^G\gamma^{g-1}\Bigl\{  \vmathbb{r}_{m;[g]}(\vmathbb{s}, \vmathbb{a}) - \widetilde{\alpha}(g)\widetilde{\mathcal{Y}}_{m;[g]} \Bigr\}
\end{align} 
\vspace{-6mm} 
\begin{subequations}\label{p8}{
		
		\begin{align*}
			\text{s.t.}~~~~
			&\widetilde{G}_{m;[g]}(\vmathbb{s}) \leq 0,~\forall g \tag{47a}
	\end{align*}}
\end{subequations}
\hspace{-0.5em}where $\widetilde{\mathcal{Y}}_{m;[g]}=\max(\widetilde{G}_{m;[g]}(\vmathbb{s}),0)^2$ represents the penalty term accounts for constraint violations. Specifically, we have $\widetilde{G}_{m;[g]}(\vmathbb{s})=t_{n;[g]}^\mathsf{Dev}-t^{\mathsf{Max}}_{n}$, and $\widetilde{\alpha}(g)$ denotes the penalty coefficient during the $g^{\text{th}}$ global iteration. Regarding the optimality of this transformation, we obtain the following theorem:

\begin{theorem} When $\widetilde{\alpha}(g)$ is sufficiently large, the solution of $\mathcal{P}_{2b}$ approximates that of $\mathcal{P}_{2a}$, thereby effectively managing the trade-off between constraint satisfaction and reward maximization.
\end{theorem}
\vspace{-1mm}
\noindent \textit{Proof:} Please refer to \textbf{Appendix C.B}.

\noindent\textbf{\textit{TD3-Based Adaptive Threshold Determination:}}  
TD3 is an off-policy RL algorithm extending DDPG to improve stability in continuous action spaces. It mitigates overestimation bias common in Q-learning-based methods\cite{Ratio-Based, Collaborative-TD3}, which can lead to suboptimal policies. This is critical in tasks like optimizing \(\mathcal{P}_{2b}\), where precise selection of \(\beta_{m;[g]}\) is required. TD3’s three key improvements over DDPG (detailed in \textbf{Appendix C.C}) make it well-suited for dynamically adjusting \(\beta_{m;[g]}\) across iterations.

To tackle $\mathcal{P}_{2b}$, we deploy a TD3-based agent at each UAV to enable efficient and adaptive decision-making in dynamic environments. However, training such agents directly in real-time incurs high sample complexity and computational overhead, especially under limited communication and energy resources. Inspired by \cite{offline1} and \cite{offline2}, we adopt a hybrid offline pre-training with online fine-tuning scheme, which significantly reduces the overall training cost while maintaining strong convergence performance. Accordingly, the online training overhead can be greatly lowered (further discussions are detailed by the following sections). In particular, each TD3 agent uses a deep neural network (DNN) to learn the approximate action value function $Q_{\boldsymbol{\vartheta}}(\vmathbb{s}_{m;[g]},\vmathbb{a}_{m;[g]})$ and the deterministic policy $\mu_{\Omega}(\vmathbb{s}_{m;[g]})$, where $\boldsymbol{\vartheta}$ and $\Omega$ represent the parameters of the value network and the policy network. Specifically, for each TD3 agent, there are four value networks (2 predictive value networks $Q_{\boldsymbol{\vartheta}_1}$, $Q_{\boldsymbol{\vartheta}_2}$ and 2 target value networks $Q_{{\boldsymbol{\vartheta}_1^{\prime}}}, Q_{\boldsymbol{\vartheta}_2^{\prime}}$) and two policy networks (predictive policy network $\mu_\Omega$, target policy network $\mu_{\Omega^{\prime}}$). Each TD3 agent follows a DNN setup tailored to the state and action space of $\mathcal{P}_{2b}$: The value network consists of an input layer, two hidden layers, and an output layer. The input layer takes two state variables $\vmathbb{s}_{m;[g]}$ (training loss and accuracy) and one action variable $\vmathbb{a}_{m;[g]}$  (adaptive threshold). The output layer produces the Q-value. Also, the actor network obeys a similar structure to the value network but outputs a continuous action instead of a Q-value.

\begin{figure}[htbp]
	\centering
	\includegraphics[trim=0cm 0cm 0cm 0cm, clip, width=0.9\columnwidth]{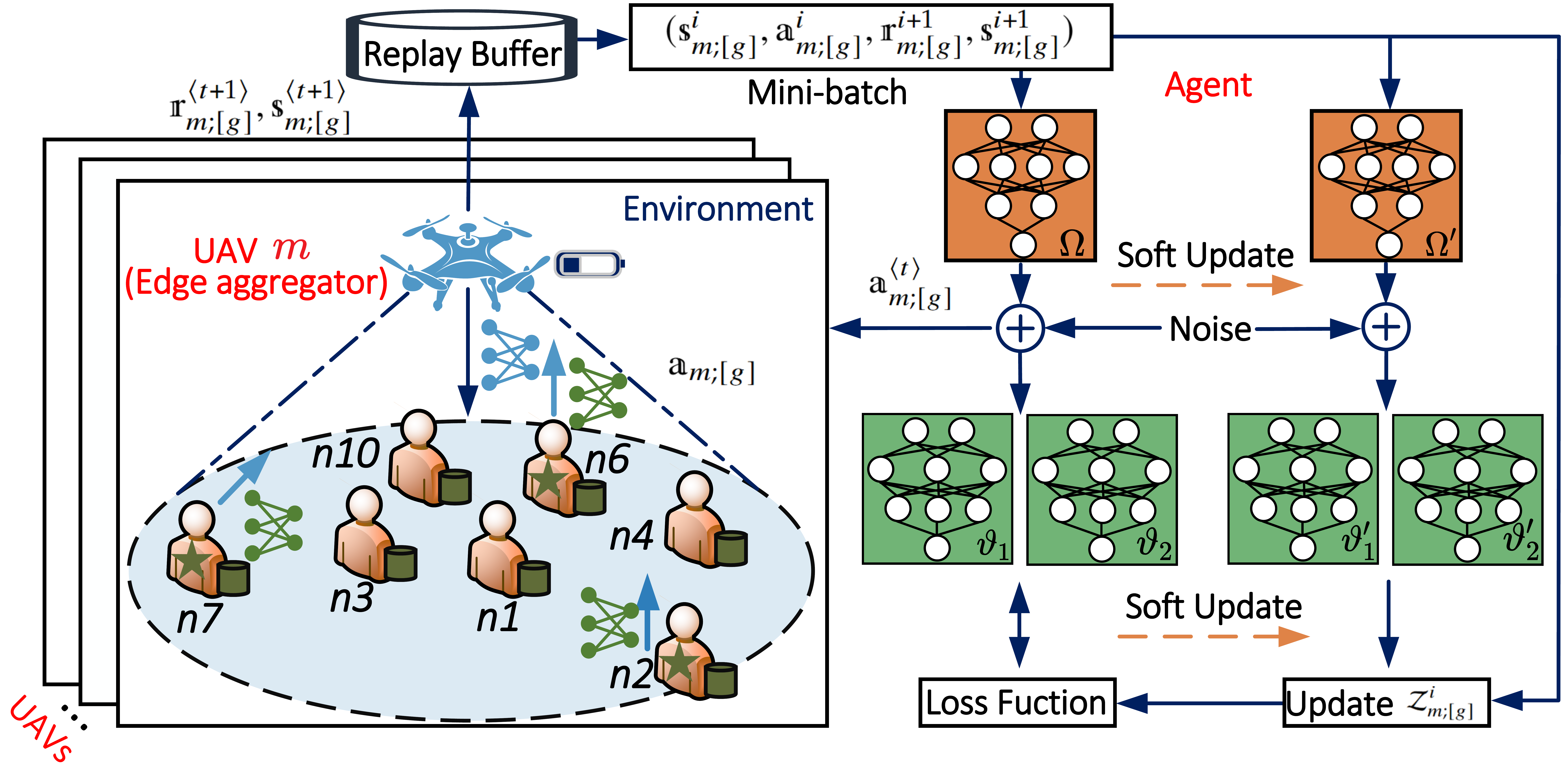}
	
	\caption{A schematic of our TD3 framework.}
	\label{fig_3}
\end{figure}

In Fig. 3, we present our TD3 framework, and at each global iteration \( g \), the TD3 agent at each UAV $m$ follows seven steps: 

\noindent \textbf{(I) Action Selection:} For each time step \( t \in \{1, 2, ..., t^{\mathsf{Step}}\} \), the agent generates an action based on the deterministic policy and exploration noise:
\setcounter{equation}{47}
\begin{align}
    \vmathbb{a}^{\langle t \rangle}_{m;[g]} = \mu_{\Omega}(\vmathbb{s}^{\langle t \rangle}_{m;[g]}) + \tilde{\epsilon}, \label{46}
\end{align}
where \( \tilde{\epsilon} \sim \text{clip}(\tilde{N}(0, \tilde{\sigma}), -\tilde{c}, \tilde{c}) \) is Gaussian exploration noise,  \( \tilde{c} \) represents the amplitude limit of the added noise, and their values are considered according to \cite{Approximation}. The optimal threshold \( \beta^*_{m;[g]} \) is selected based on the action that maximizes the cumulative reward.

\noindent \textbf{(II) Reward Computation:}
After executing the selected action \( \vmathbb{a}_{m;[g]} \), the TD3 agent computes the reward:
\begin{align}
    \vmathbb{r}^{\langle t+1 \rangle}_{m;[g]} = \vmathbb{r}^{\langle t+1 \rangle}_{m;[g]} (\vmathbb{s}^{\langle t \rangle}, \vmathbb{a}^{\langle t \rangle}) - \widetilde{\alpha}(g) \widetilde{\mathcal{Y}}^{\langle t+1 \rangle}_{m;[g]}, \label{47}
\end{align}
where \( \widetilde{\mathcal{Y}}^{\langle t+1 \rangle}_{m;[g]} \) is a penalty term for constraint violations.

\noindent \textbf{(III) Experience Replay:}
The agent stores each transition experience \( (\vmathbb{s}^{\langle t \rangle}_{m;[g]}, \vmathbb{a}^{\langle t \rangle}_{m;[g]}, \vmathbb{r}^{\langle t+1 \rangle}_{m;[g]}, \vmathbb{s}^{\langle t+1 \rangle}_{m;[g]}) \) in an experience replay buffer \( \mathcal{B} \) with a maximum capacity \( |\mathcal{B}|_{\mathsf{max}} \).

\noindent \textbf{(IV) Value Network Update:}
The TD3 agent samples a mini-batch  $B$ from $\mathcal{B}$ (i.e., $B\subset \mathcal{B}$) and computes the updated action using the target policy network:
\begin{align}
    \vmathbb{a}_{m;[g]}^{i+1,\prime}=\mu_{\Omega^{\prime}}(\vmathbb{s}_{m;[g]}^{i+1})+\widetilde{\varepsilon}^i,
    \widetilde{\varepsilon}^i\sim clip(\widetilde{N}(0,\widetilde{\sigma}),-\widetilde{c},\widetilde{c}), \label{48}
\end{align}
where $i = \{1,2,...,|B|\}$ is defined as the $i^\text{th}$ sample. The target Q-value is then computed as:
\begin{align}
    \mathcal{Z}_{m;[g]}^i=\vmathbb{r}_{m;[g]}^{i+1}+\gamma^{g-1}\min_{j=1,2}Q_{\boldsymbol{\vartheta}_j^{\prime}}(\vmathbb{s}_{m;[g]}^{i+1},\vmathbb{a}_{m;[g]}^{i+1,\prime}), \label{49}
\end{align}
where $j$ represents the $j^\text{th}$ Q-value networks. Finally, the value network parameters \( \vartheta \) are updated using Mini-Batch Gradient Descent (MBGD) with the gradient
\begin{multline}
	\nabla_{\boldsymbol{\vartheta}}L(\boldsymbol{\vartheta})\approx
\frac1{|B|}\sum_{i=1}^{|B|}(\mathcal{Z}_{m;[g]}^i-\\Q(\vmathbb{s}_{m;[g]}^i,\vmathbb{a}_{m;[g]}^i))\nabla_{\boldsymbol{\vartheta}}Q(\vmathbb{s}_{m;[g]}^i,\vmathbb{a}_{m;[g]}^i). \label{50}
\end{multline}

\noindent \textbf{(V) Policy Network Update:}
After every \( \vmathbb{d} \) updates of the value network, the policy network parameters \( \Omega \) are updated using Mini-Batch Gradient Ascent (MBGA) with the gradient
\begin{align}
    \nabla_{\Omega} J(\Omega) \approx \frac{1}{|B|} \sum_{i=1}^{|B|} \nabla_{\Omega} \mu_{\Omega}(\vmathbb{s}^i_{m;[g]}) \nabla_{\vmathbb{a}^i} Q_{\Omega}(\vmathbb{s}^i_{m;[g]}, \vmathbb{a}^i_{m;[g]}). \label{51}
\end{align}

\noindent \textbf{(VI) Penalty Coefficient Update:} The penalty coefficient is updated incrementally as follows:
\begin{align}
    \widetilde{\alpha}(g+1)=\left\{\begin{array}{ll}\widetilde{\alpha}(g)+\Delta \widetilde{\alpha}, & \text { if } t \bmod \vmathbb{d}=0, \\\widetilde{\alpha}(g), & \text { otherwise,}\end{array}\right. \label{52}
\end{align}
where $\Delta \widetilde{\alpha}$ represents a constant used to gradually increase the penalty coefficient. 

\noindent \textbf{(VII) Target Network Soft Update:} The value and policy networks are soft-updated as follows:
\begin{align}
	\begin{cases}\boldsymbol{\vartheta}^{\prime}&\leftarrow\tau \boldsymbol{\vartheta}+(1-\tau)\boldsymbol{\vartheta}^{\prime},\\
 \Omega^{\prime}&\leftarrow\tau\Omega+(1-\tau)\Omega^{\prime},\end{cases}  \label{53}
\end{align}
where $\tau \in (0,1)$ is the update coefficient. Based on the TD3 agent, we select appropriate thresholds $\beta_{m;[g]}$ for selecting devices in each round of global iterations for each UAV. We summarize the detailed procedure of our method in Alg. 3 (see \textbf{Appendix C.D)}.

\subsection{Problem Formulation and Solution Design of $\mathcal{P}_3$}
After UAV disconnections, device coverage drops, causing two challenges: delayed aggregation from disconnected UAVs slows convergence, and fewer active UAVs reduce learning efficiency. We propose a proactive UAV selection and redeployment strategy. A global aggregator UAV is designated before disconnections to collect models from active UAVs, while remaining UAVs reposition to maximize device coverage and maintain participation in aggregation rounds. Jointly optimizing repositioning and aggregator selection is critical, as UAV movement consumes mobility energy and aggregation energy depends on the aggregator’s location. Optimizing UAV positions and aggregator selection is key to minimizing system cost while ensuring robust learning. We formulate an optimization problem to determine both UAV repositioning and global aggregator selection:
\begin{multline}
	\bm{\mathcal{P}}_3\hspace{-0.3em}:\underset{{\mathtt{p}_{m;[g]}}, {X_{m;[g]}}}{\operatorname{argmin}}~~\lambda_4\left\{E_{[g]}^\mathsf{Broad}+E_{[g]}^\mathsf{Bwait}+\sum_{m\in\mathcal{M}_{[g]}}E_{m;[g]}^\mathsf{Delay}\right\}\\
    +\lambda_5 \left\{T_{[g]}^\mathsf{Broad}+\max_{m\in \mathcal{M}_{[g]}}T_{m;[g]}^\mathsf{Delay} \right\}
\end{multline}
\vspace{-6mm} 
\begin{subequations}\label{p9}{
		
		\begin{align*}
			\text{s.t.}~~~~
			&\text{\eqref{27d}, \eqref{27e}, \eqref{27h}, \eqref{27i}}  \notag
	\end{align*}}
\end{subequations}

Directly solving this problem is intractable due to the coupling between UAV repositioning and aggregator selection. In dynamic scenarios, decisions must be fast and lightweight. Motivated by greedy methods in UAV deployment \cite{Optimization, Multi-UAV}, we adopt a two-stage framework: first, optimize UAV positions in the \(g^\text{th}\) global iteration (\(\mathtt{p}_{m;[g]}\)); second, select the central aggregator UAV (\(X_{m;[g]}=1\)) (details in \textbf{Appendix D}).

\section{Evaluations}
In the following, we evaluate our proposed approach, ``CEHFed'' (Cost Effective Hierarchical Federated Learning). First, CEHFed is benchmarked against state-of-the-art HFL methods using test accuracy, training time, and energy efficiency (see Section VI.C.1)). Second, scenario-based evaluations assess CEHFed under UAV dropouts and dynamic device mobility (see Section VI.C.2)).

\subsection{Network and Machine Learning Settings}
Our experiments are conducted in a 20 km × 20 km area, where 5 UAVs provide coverage for 150 terrestrial IoT devices. Each UAV has a coverage radius of 5 km, enabling dynamic interactions between devices and UAVs as they move within the network. We consider that in each global iteration, each device will leave the coverage area of its associated UAV with probability $\xi$, the default value of which is chosen to be $0.3$.

\noindent $\bullet$ \textit{Dataset and training models:} We use MNIST, Fashion-MNIST (FaMNIST), and Aerial Landscape Images (ALI) datasets for classification tasks. The ALI dataset is constructed from publicly available AID and NWPU-Resisc45 satellite/UAV image datasets \cite{AA, BB} to better reflect real UAV scenarios. MNIST and FaMNIST contain 10 categories with 28×28 grayscale images, while ALI includes 15 real-world aerial scene categories (e.g., rivers, forests, farmland), from which 10 representative ones are selected. To evaluate performance, we employ three neural network models—CNN, LeNet5, and VGG—with parameter sizes of 21,840, 60,074, and 206,922, respectively \cite{Adaptive, Automaton, 3D-TrIM}.

\noindent $\bullet$ \textit{Data heterogeneity:} For MNIST and FaMNIST, we examine two non-iid settings. In non-iid (A), each device holds samples from two labels, creating a highly skewed distribution. In non-iid (B), each device has samples from at least two labels, randomly selected between two and ten, yielding a more balanced distribution with the same total data size.

\noindent $\bullet$ \textit{Resource heterogeneity:} Device resource heterogeneity mainly arises from differences in computing capabilities (e.g., CPU frequency). Detailed configurations and other system settings are given in \textbf{Appendix~E.A}.

\subsection{Benchmark Methods}
We compare against several benchmark methods described below, representing different strategies in FL and HFL.

\noindent $\bullet$ \textit{CFed:} A conventional FL mechanism, where a certain number of devices are efficiently selected for each global iteration \cite{PerformanceO}.

\noindent $\bullet$ \textit{HFed:} An HFL mechanism, aims to optimize device selection during edge iterations only, without optimizing the number of local training and bandwidth allocation (use our solution of $\mathcal{P}_2$) \cite{Managementf}.

\noindent $\bullet$ \textit{RHFed:} An HFL mechanism, randomly selects devices while optimizing the number of local training and bandwidth allocation (use our solution of $\mathcal{P}_1$).

\noindent $\bullet$ \textit{GDHFed:} An HFL mechanism, selects devices based on their proximity to UAVs (i.e., only $S^\mathsf{Dis}_{m,n;[g]}$ is considered while calculating $\alpha_{m,n;[g]}$), while optimizing the number of local training and bandwidth allocation (use our solution of $\mathcal{P}_1$).

\noindent $\bullet$ \textit{GSHFed:} An HFL mechanism, selects devices with model difference score (i.e., only $S^\mathsf{Sim}_{m,n;[g]}$ is considered while calculating $\alpha_{m,n;[g]}$),  while optimizing the number of local training and bandwidth allocation (use our solution of $\mathcal{P}_1$).

\noindent $\bullet$ \textit{AHFed:} Adds adversarial training to conventional HFL, reducing the negative impact of device data heterogeneity \cite{Asymmetric}.

\noindent $\bullet$ \textit{HFedAT:} An HFL mechanism that combines synchronous inner-layer training and asynchronous cross-layer training, without considering data heterogeneity of devices \cite{asynchronoust}.

\subsection{Performance Comparisons}
We begin by comparing performance using standard evaluation metrics, including test accuracy, training time, and energy efficiency (Section VI.C.1)\footnote{Owing to spatial constraints, simulations and corresponding analysis on ALI dataset are moved to \textbf{Appendix E.B}.}). Since our work addresses unique challenges, such as UAV dropouts during the training process, we further conduct scenario-based simulations to demonstrate the adaptability of our proposed framework, highlighting its advantages in dynamic UAV-assisted HFL environments (Section VI.C.2)).
\begin{figure}[!t]
	\centering
    \scriptsize  
	\subfigure[CNN on MNIST]{
		\includegraphics[trim=0.3cm 0cm 1.6cm 0cm, clip, width=0.305\columnwidth]{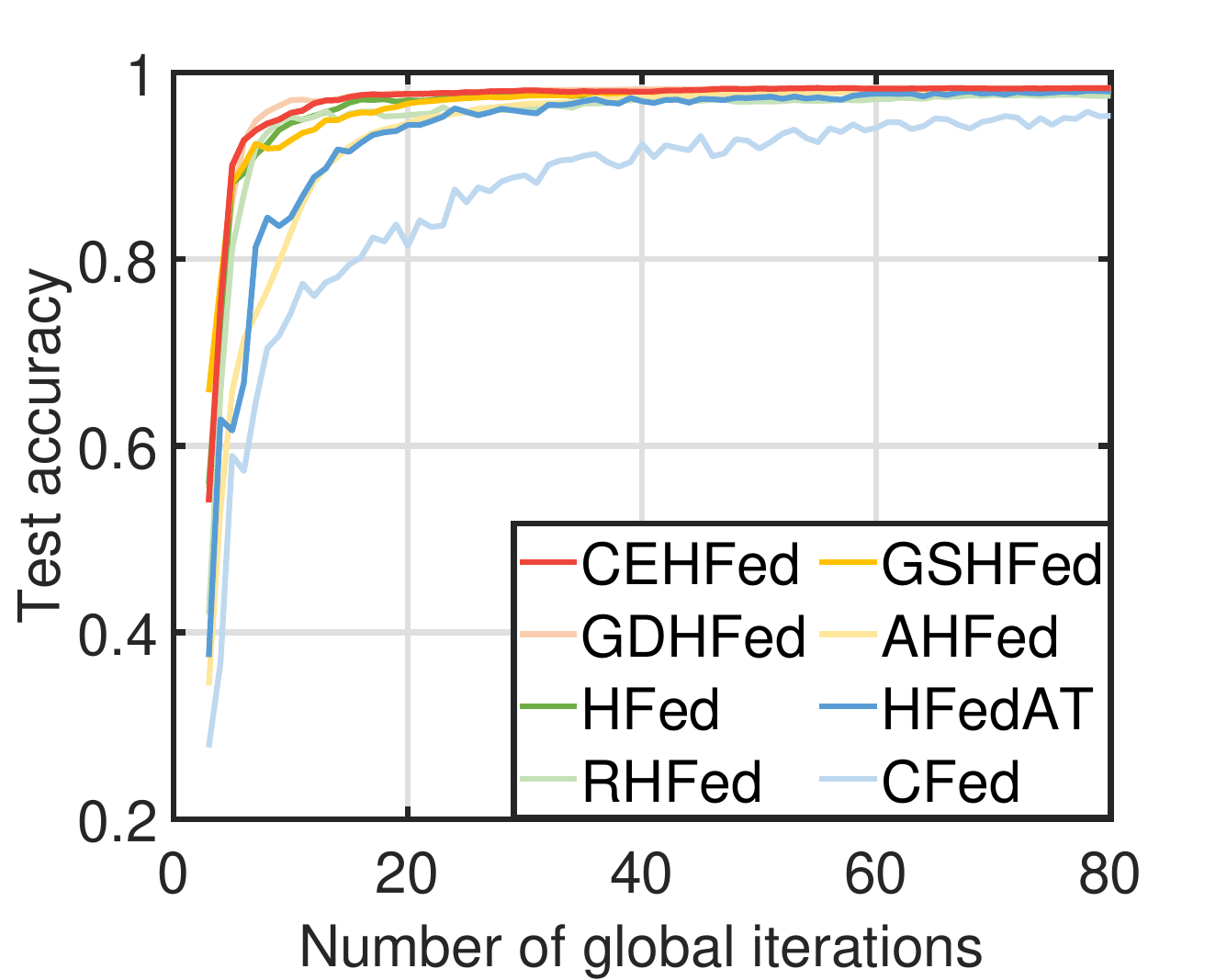}
	}
	\subfigure[LeNet5 on MNIST]{
		\includegraphics[trim=0.4cm 0cm 1.6cm 0cm, clip, width=0.305\columnwidth]{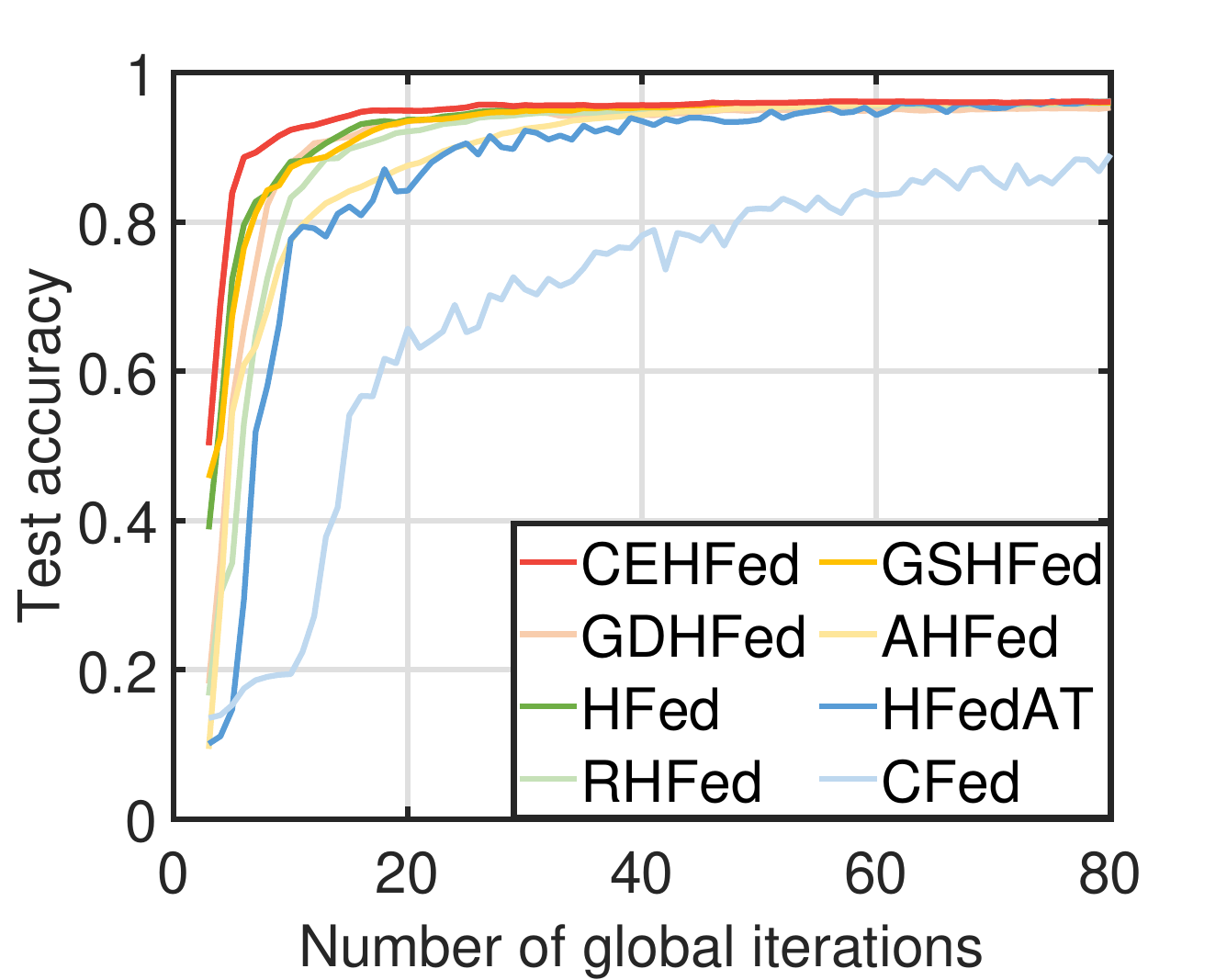}
	}
	\subfigure[VGG on MNIST]{
		\includegraphics[trim=0.4cm 0cm 1.6cm 0cm, clip, width=0.305\columnwidth]{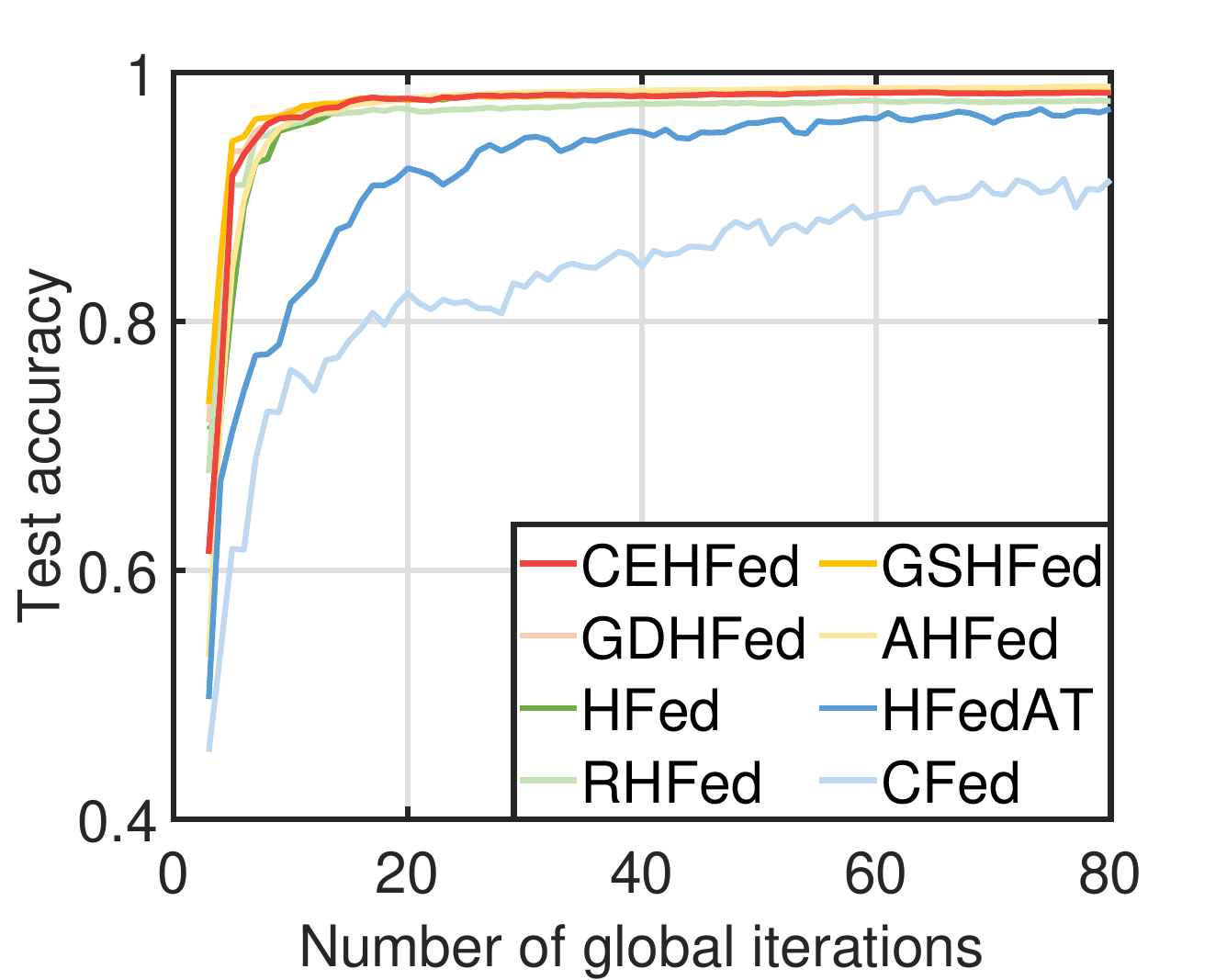}
	}
	
	\subfigure[CNN on FaMNIST]{
		\includegraphics[trim=0.3cm 0cm 1.6cm 0cm, clip, width=0.305\columnwidth]{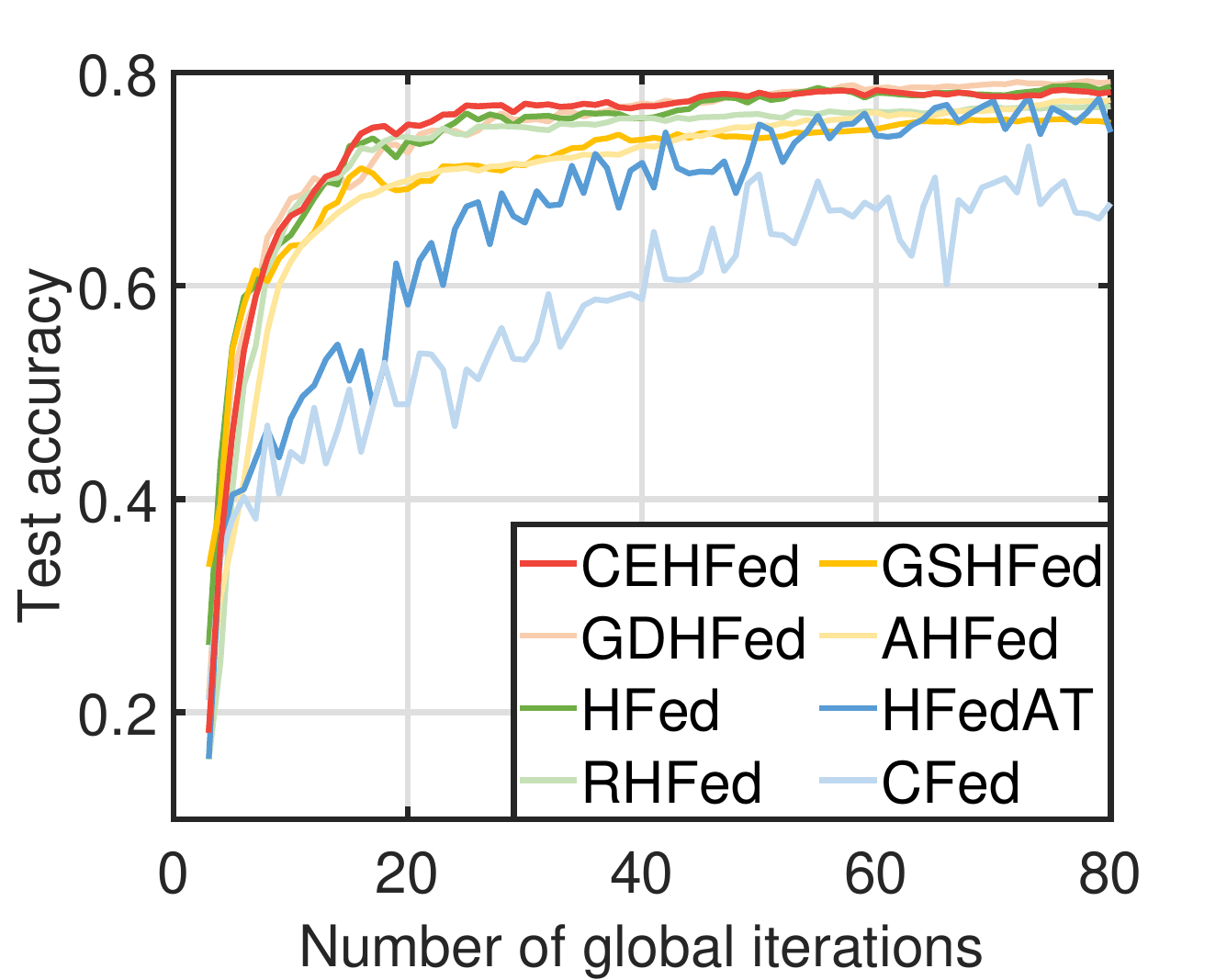}
	}
	\subfigure[LeNet5 on FaMNIST]{
		\includegraphics[trim=0.3cm 0cm 1.6cm 0cm, clip, width=0.305\columnwidth]{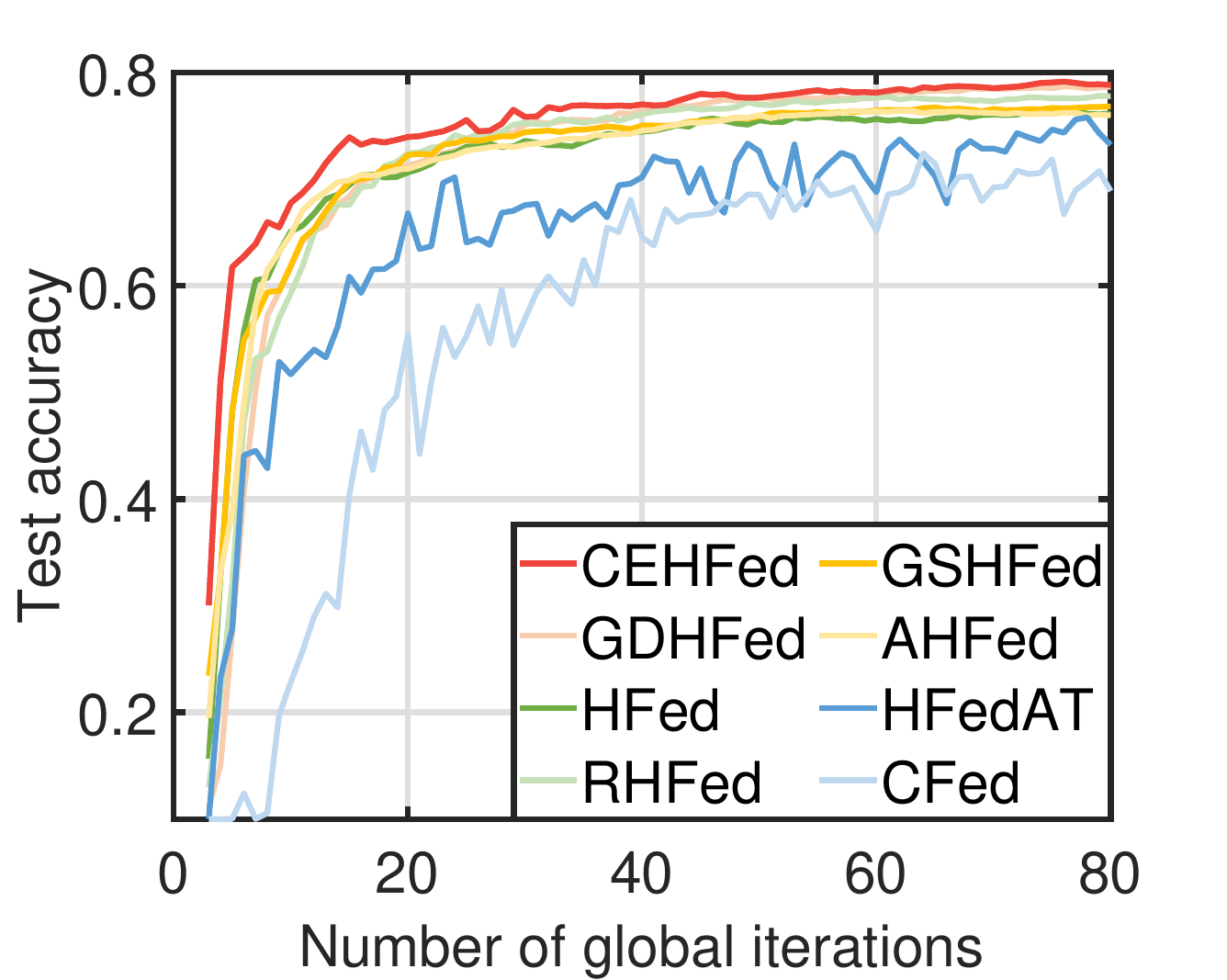}
	}
	\subfigure[VGG on FaMNIST]{
		\includegraphics[trim=0.3cm 0cm 1.6cm 0cm, clip, width=0.305\columnwidth]{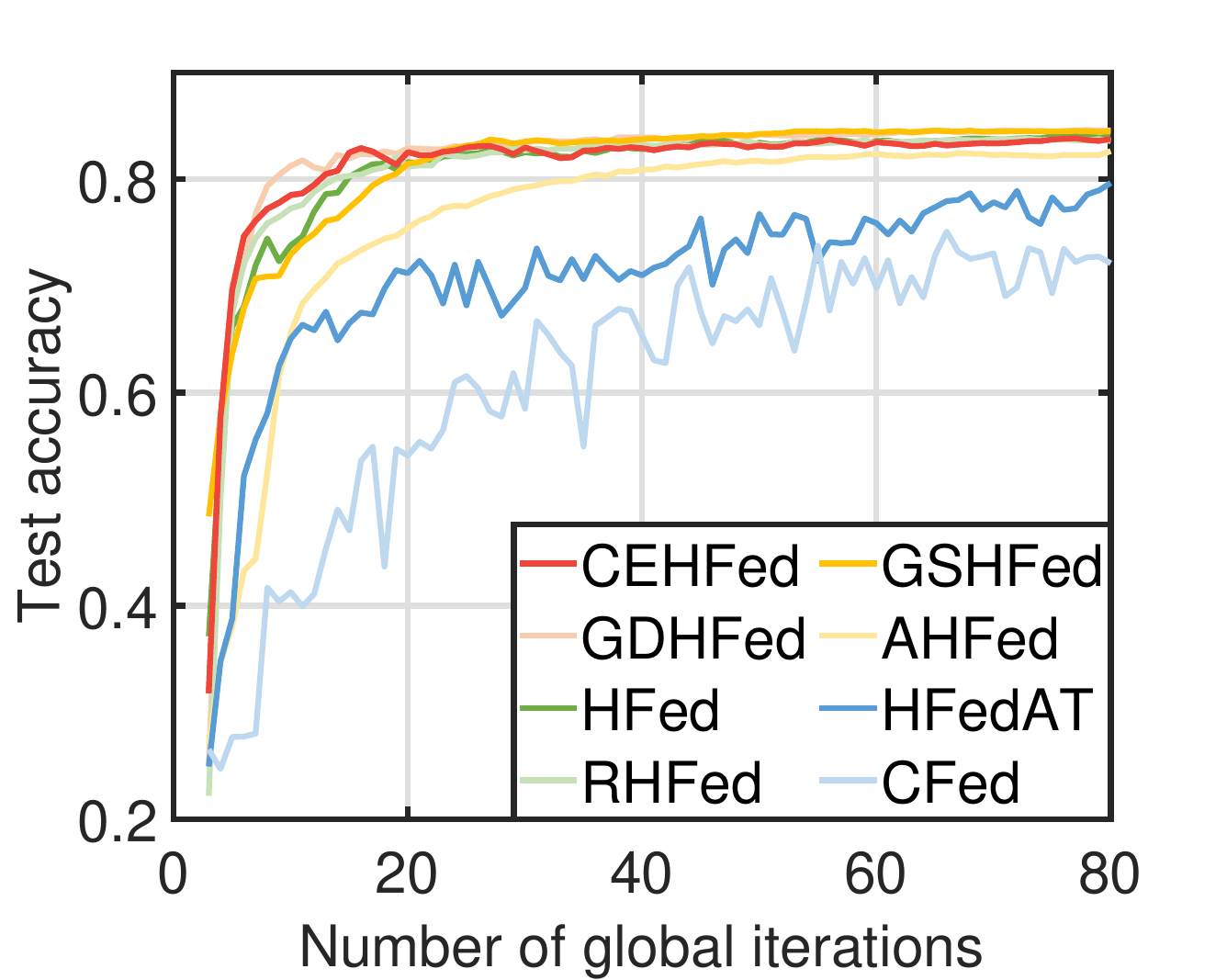}
	}
	\caption{Performance comparisons in terms of test accuracy on MNIST and FaMNIST datasets using different models.}
	\label{fig_4}
\end{figure}

\subsubsection{Experiments with conventional evaluating metrics}
\noindent
\textbf{Convergence Performance:} We first evaluate convergence across different learning models and datasets, as shown in Fig.~4. Overall, CEHFed achieves comparable or superior convergence to benchmark methods, with only minor trade-offs in some cases\footnote{Detailed analysis can be found in \textbf{Appendix E.C}.}. However, convergence alone does not capture key metrics such as energy consumption and latency. Crucially, CEHFed attains this performance with substantially lower network resource usage. To demonstrate this, we first examine training delay, followed by an analysis of energy consumption, highlighting CEHFed's efficiency in UAV-assisted HFL.

 
\noindent
\textbf{Time Cost (Delay) Performance:} 
We next analyze time cost across different methods during HFL, considering varying data volumes\footnote{Data volumes refers to the number of data points used for training.}. As shown in Fig.~5, CEHFed consistently outperforms all benchmarks in time efficiency. This is due to optimized bandwidth allocation, adaptive device selection, and dynamic global aggregator choice, which reduce edge iteration time and communication delays. For example, in Fig.~5(a), training CNN on MNIST with 4k data points, CEHFed reduces time by 17$\%$, 63$\%$, and 55$\%$ compared to GDHFed, GSHFed, and RHFed, respectively. Benchmark methods relying on random or greedy device selection fail to adaptively choose optimal devices, causing inefficiencies. Compared to HFed, CEHFed lowers training time by 31$\%$, and against CFed, AHFed, and HFedAT, reductions reach 79$\%$, 69$\%$, and 73$\%$, respectively. These results highlight CEHFed’s superior efficiency in accelerating HFL while optimizing UAV-assisted network resources.
\begin{figure}[!t]
	\centering
	\subfigure[CNN on MNIST]{
		\includegraphics[trim=0.3cm 0.1cm 1.6cm 0cm, clip, width=0.295\columnwidth]{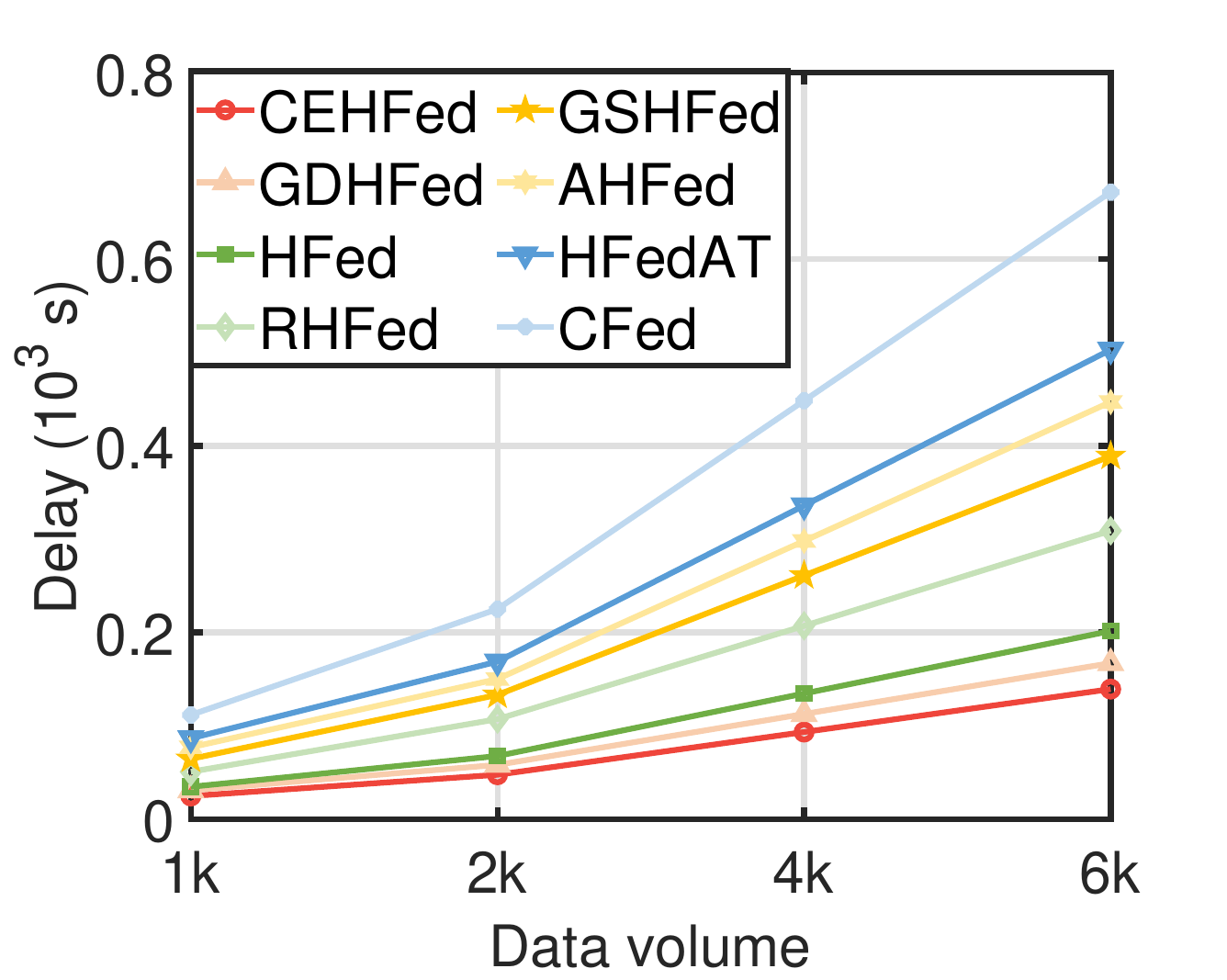}
	}
	\subfigure[LeNet5 on MNIST]{
		\includegraphics[trim=0.3cm 0.1cm 1.6cm 0cm, clip, width=0.295\columnwidth]{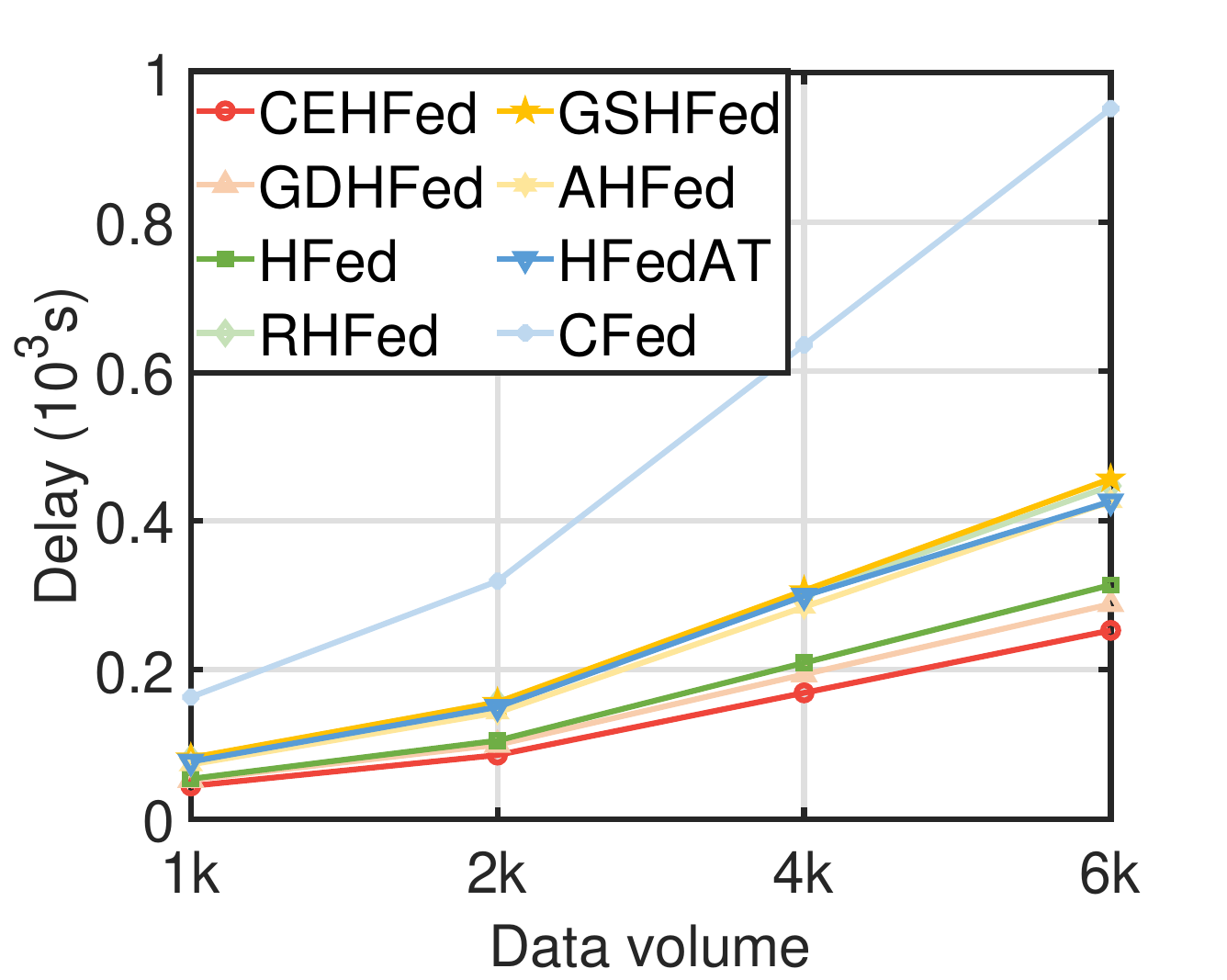}
	}
	\subfigure[VGG on MNIST]{
		\includegraphics[trim=0.3cm 0.1cm 1.6cm 0cm, clip, width=0.295\columnwidth]{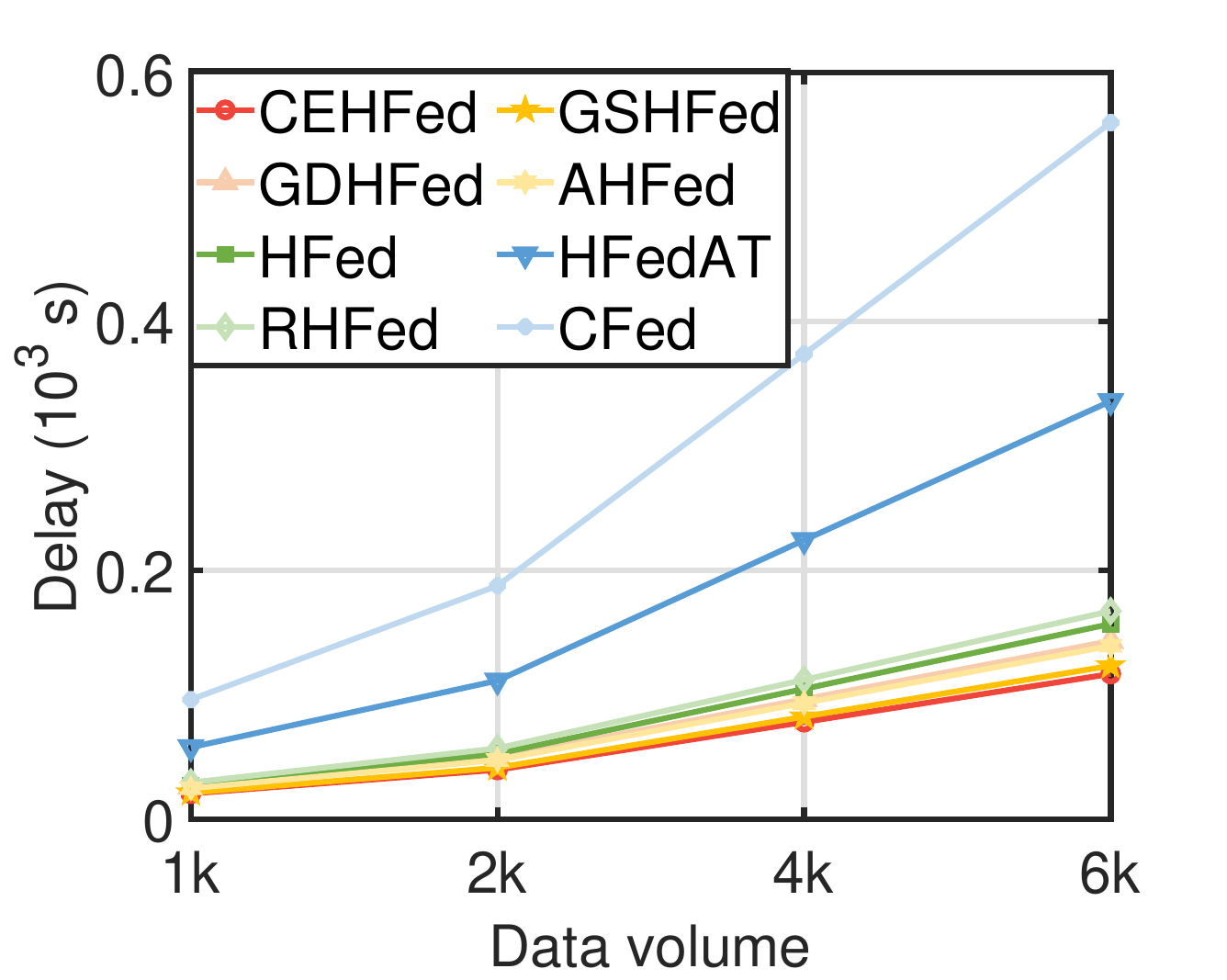}
	}
	
	\subfigure[CNN on FaMNIST]{
		\includegraphics[trim=0.3cm 0.1cm 1.6cm 0cm, clip, width=0.295\columnwidth]{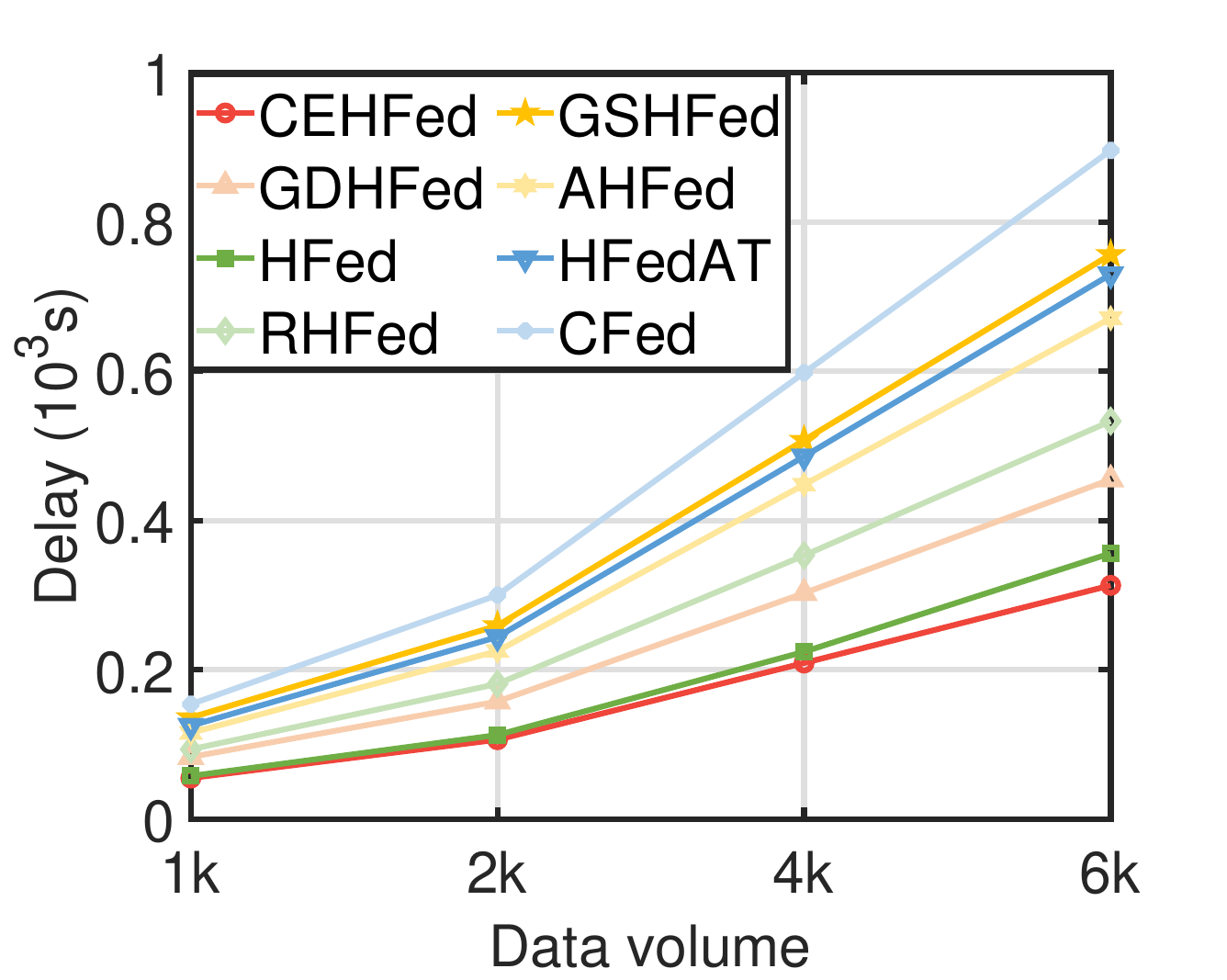}
	}
	\subfigure[LeNet5 on FaMNIST]{
		\includegraphics[trim=0.3cm 0.1cm 1.6cm 0cm, clip, width=0.295\columnwidth]{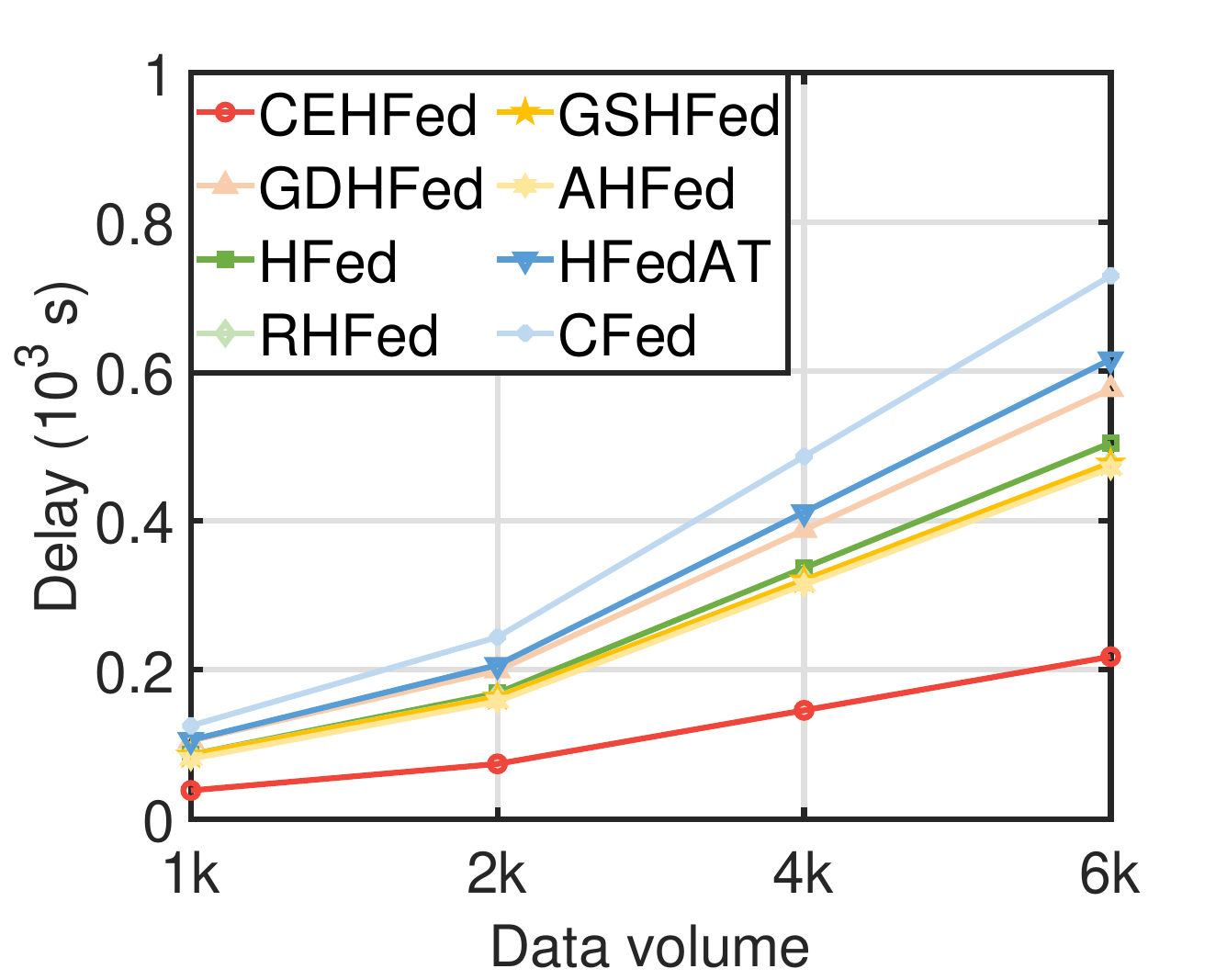}
	}
	\subfigure[VGG on FaMNIST]{
		\includegraphics[trim=0.3cm 0.1cm 1.6cm 0cm, clip, width=0.295\columnwidth]{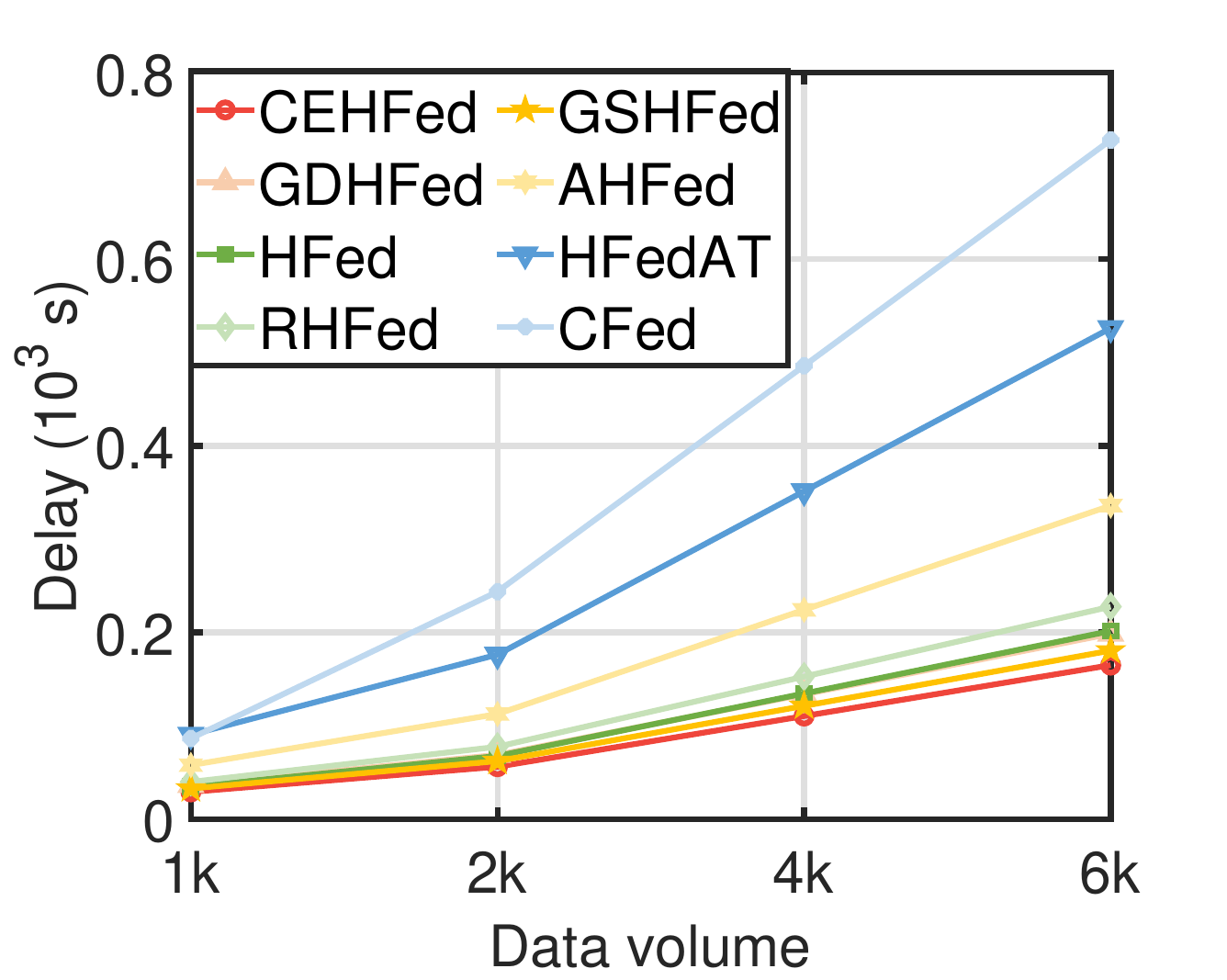}
	}
	\caption{Time cost of model training on MNIST and FaMNIST datasets upon having different models.}
	\label{fig_5}
\end{figure}

\begin{figure}[!t]
	\centering
	\subfigure[CNN on MNIST]{
		\includegraphics[trim=0.3cm 0.1cm 1.9cm 0cm, clip, width=0.295\columnwidth]{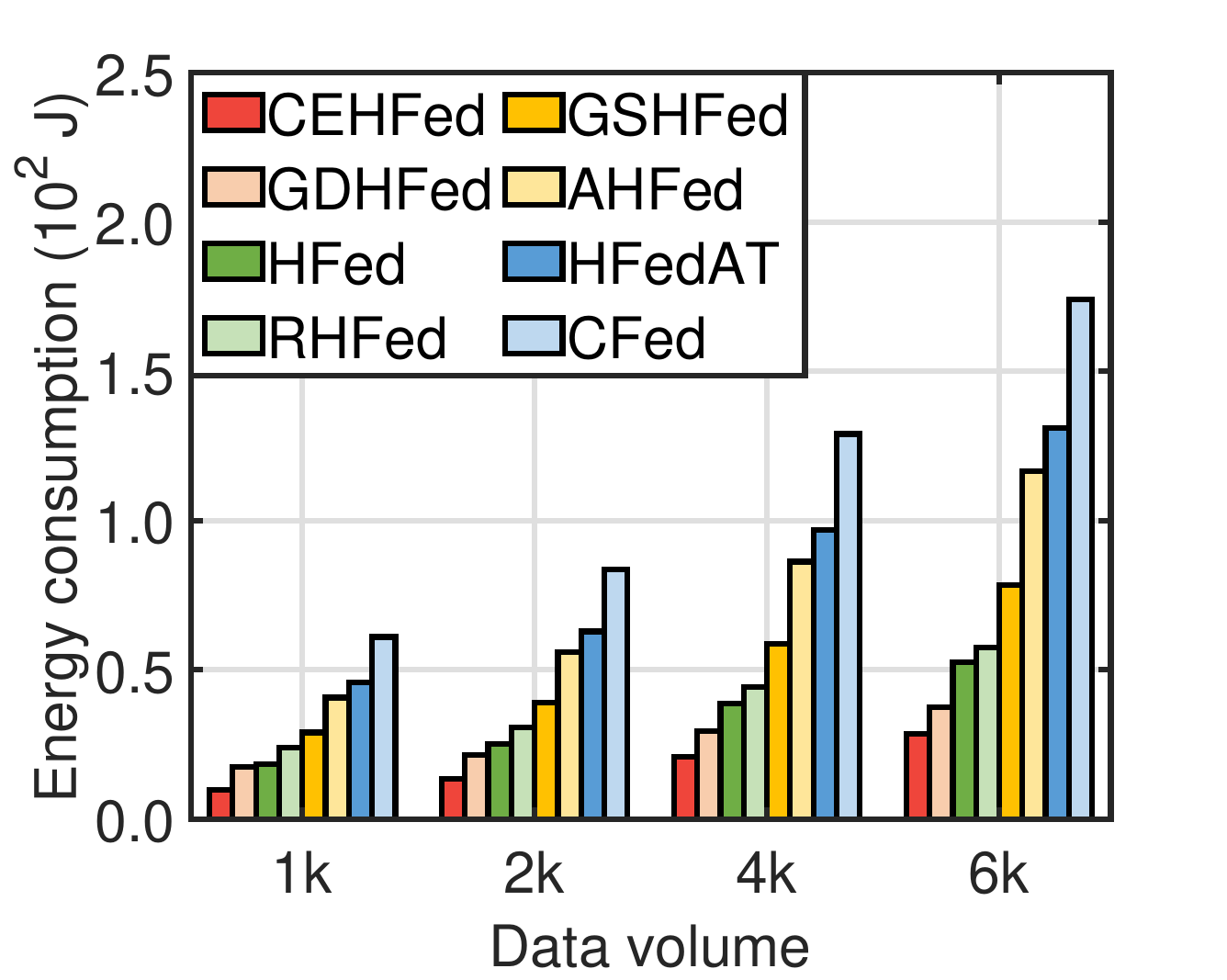}
	}
	\subfigure[LeNet5 on MNIST]{
		\includegraphics[trim=0.3cm 0.1cm 1.9cm 0cm, clip, width=0.295\columnwidth]{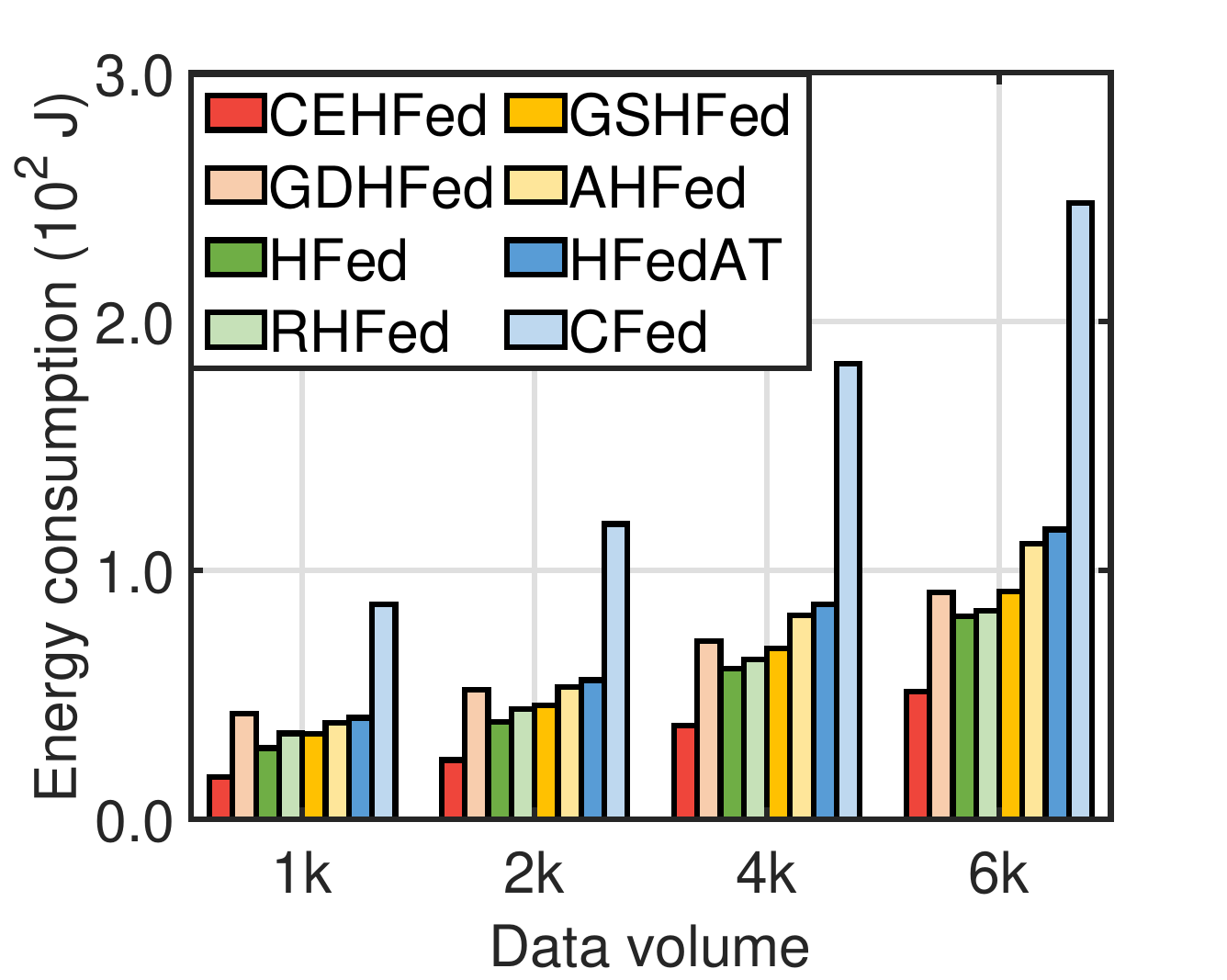}
	}
	\subfigure[VGG on MNIST]{
		\includegraphics[trim=0.3cm 0.1cm 1.9cm 0cm, clip, width=0.295\columnwidth]{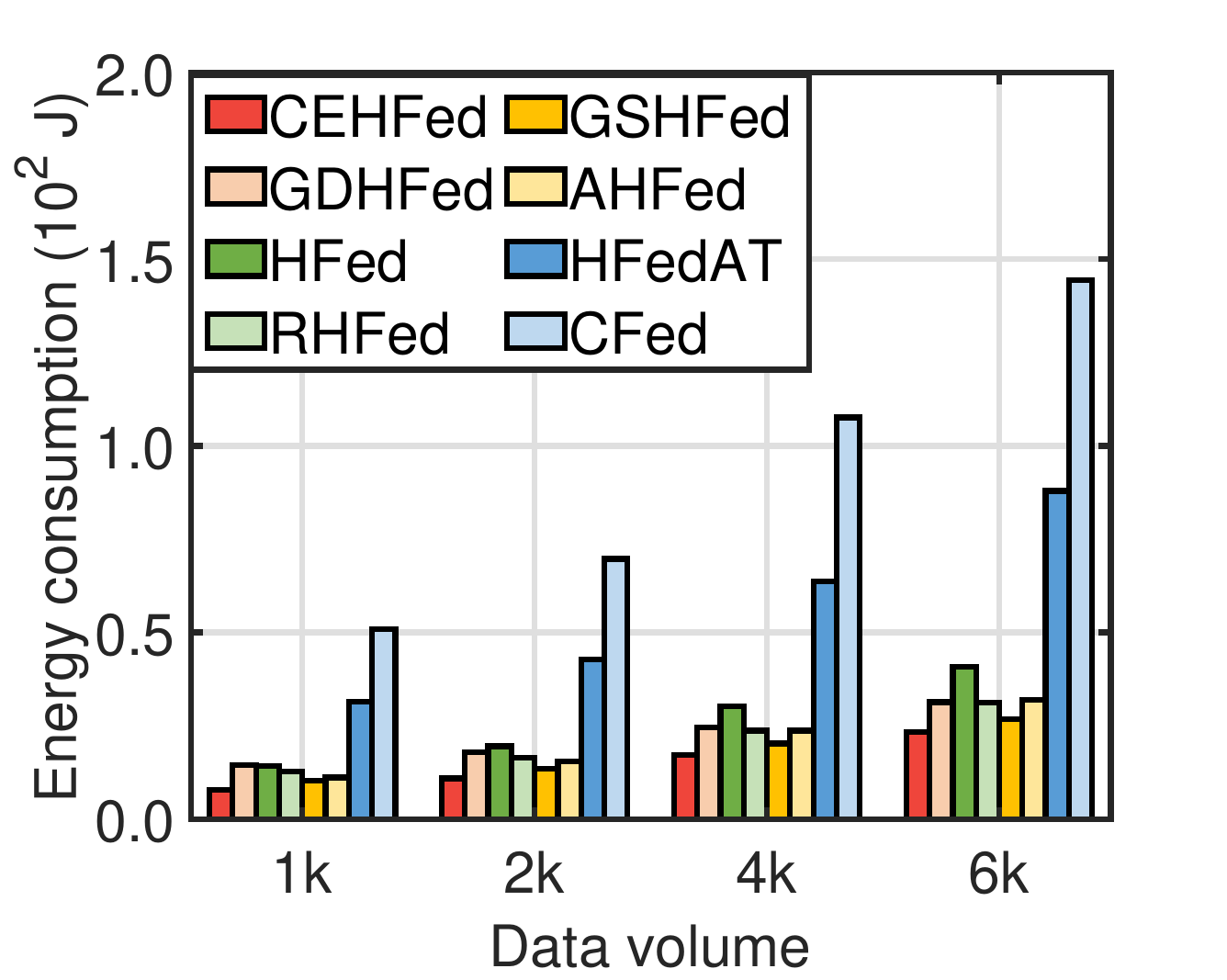}
	}
	
	\subfigure[CNN on FaMNIST]{
		\includegraphics[trim=0.3cm 0.1cm 1.9cm 0cm, clip, width=0.295\columnwidth]{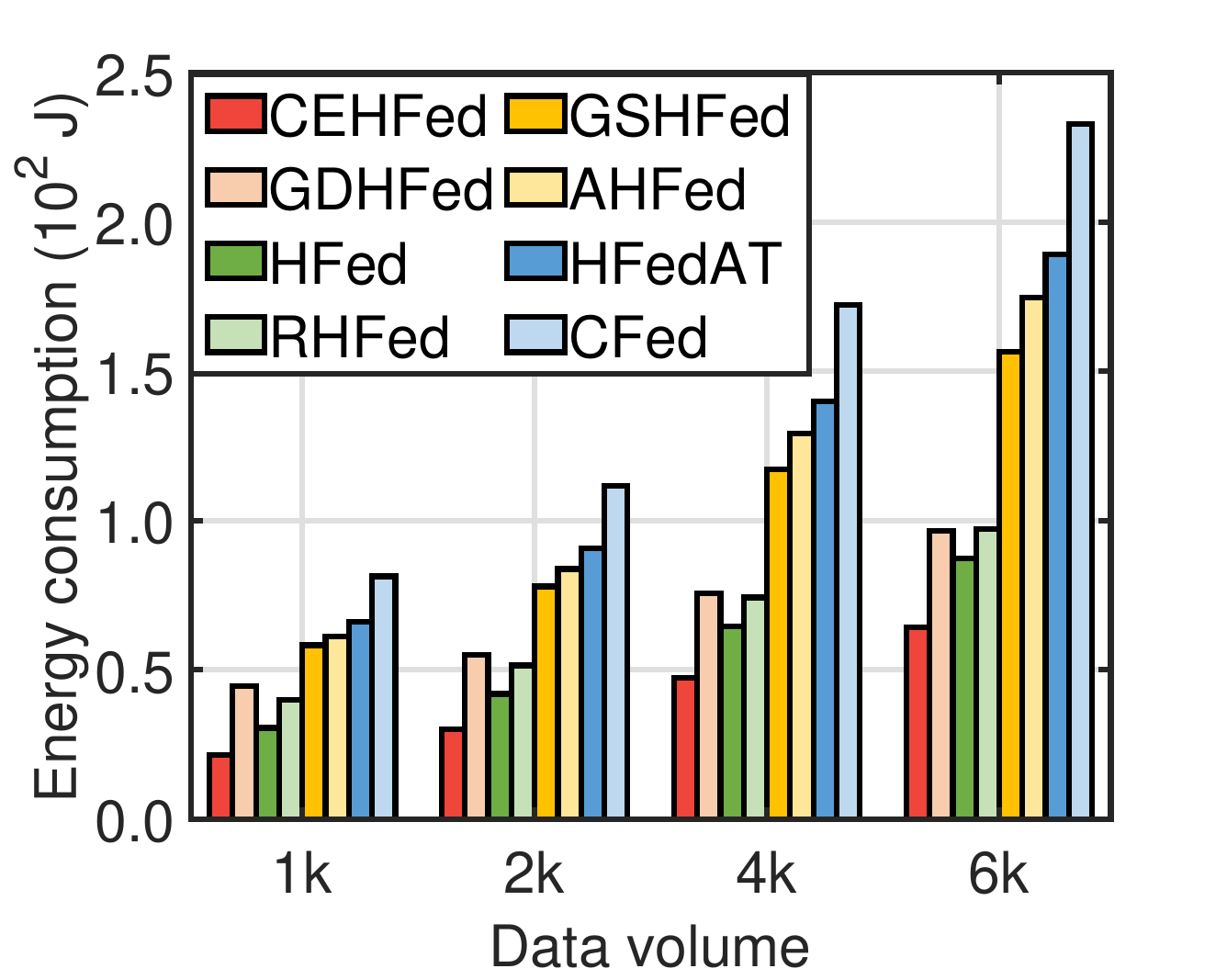}
	}
	\subfigure[LeNet5 on FaMNIST]{
		\includegraphics[trim=0.3cm 0.1cm 1.9cm 0cm, clip, width=0.295\columnwidth]{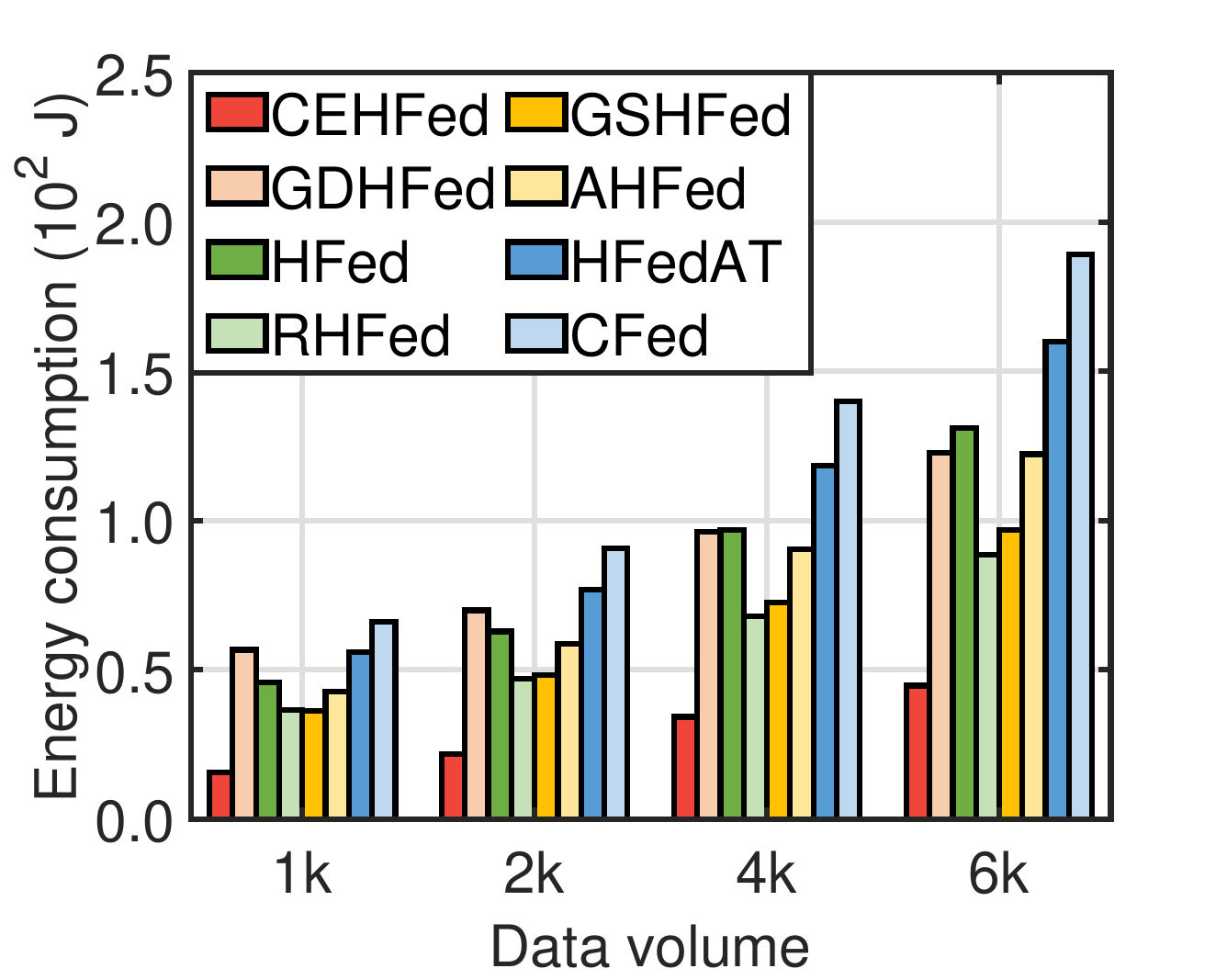}
	}
	\subfigure[VGG on FaMNIST]{
		\includegraphics[trim=0.3cm 0.1cm 1.9cm 0cm, clip, width=0.295\columnwidth]{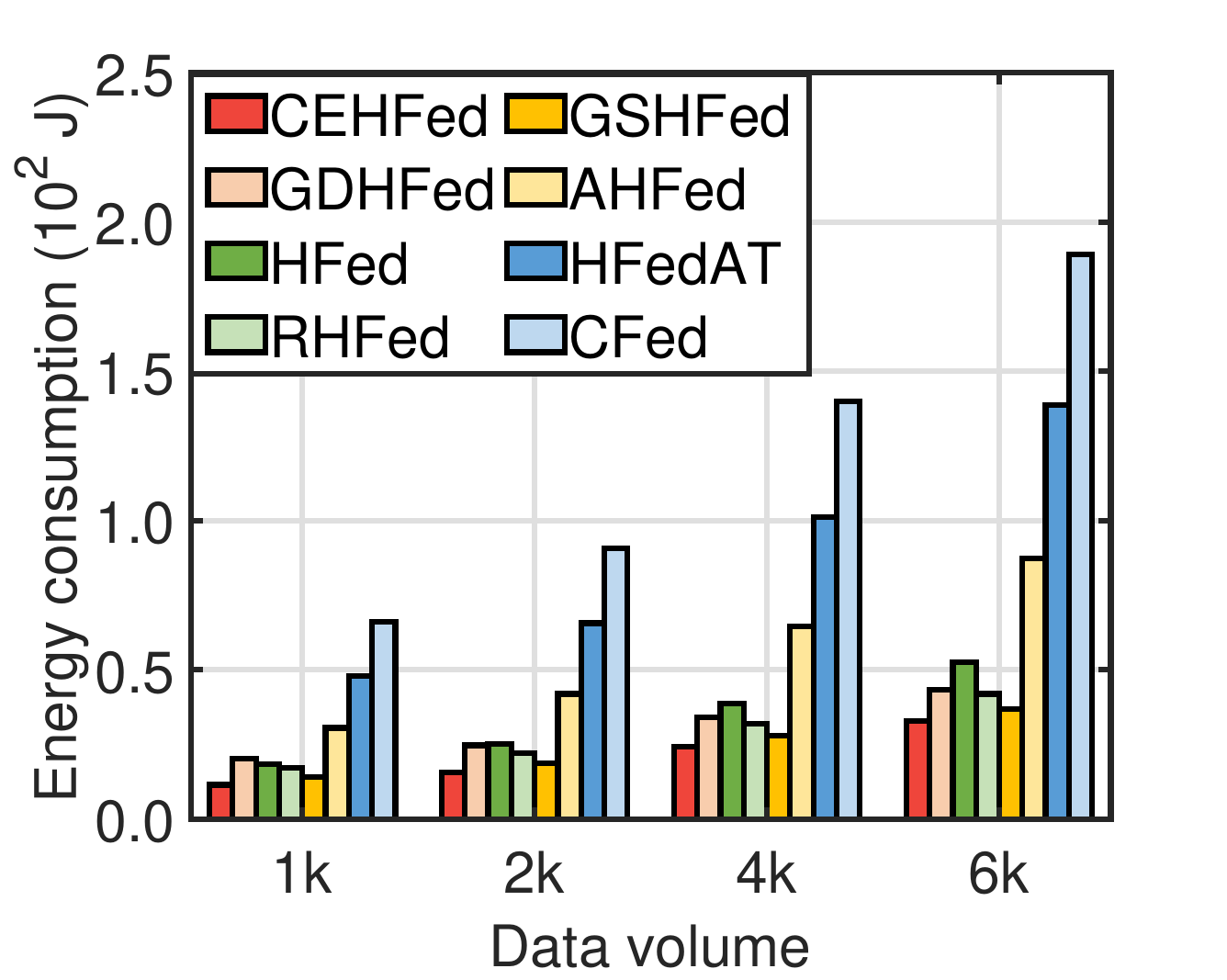}
	}
	\caption{Energy consumption of the model training operations on MNIST and FaMNIST datasets upon having
different models.}
	\label{fig_6}
\end{figure}

\noindent
\textbf{Energy Consumption Performance:}  
We analyze overall energy consumption in HFL (Fig.~6). Energy cost does not always correlate with time; lower training time does not imply lower energy. For instance, RHFed on CNN and FaMNIST (Fig.~5(d), Fig.~6(d)) has higher time cost than GDHFed but comparable or lower energy consumption, due to the non-linear relationship between time and energy (Section~III.C). Fig.~6 shows that CEHFed significantly outperforms all benchmarks. It reduces energy consumption by 62\%, 52\%, and 47\% compared to GDHFed, GSHFed, and RHFed, respectively. GDHFed ignores computation costs, GSHFed partially considers them but overlooks communication energy, and RHFed neglects both. CEHFed also achieves 64\% lower energy than HFed, and reduces energy by 75\%, 61.8\%, and 70.8\% compared to CFed, AHFed, and HFedAT. These gains highlight CEHFed’s joint optimization: optimizing local iterations and bandwidth in \(\mathcal{P}_1\) lowers training cost; adaptive multi-criteria client selection in \(\mathcal{P}_2\) improves performance under device heterogeneity; UAV redeployment and aggregator selection in \(\mathcal{P}_3\) enhance resilience (Section~VI.C.1)). Consequently, CEHFed achieves superior robustness and lower overall energy cost than existing methods.

\begin{figure}[!t]
	\centering
	\subfigure[Convergence]{
		\includegraphics[trim=0.3cm 0.05cm 1.6cm 0cm, clip, width=0.295\columnwidth]{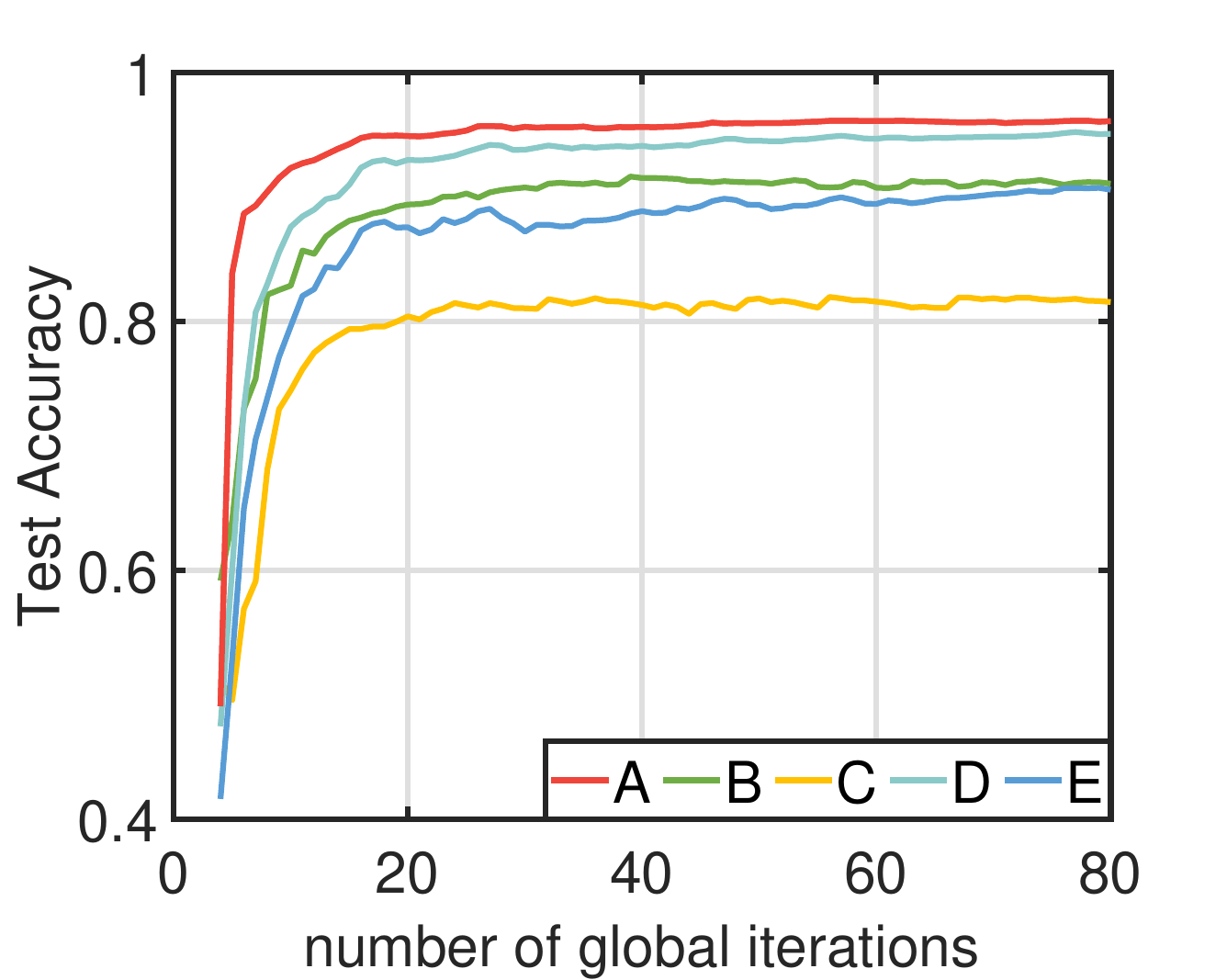}
	}
	\subfigure[Time cost]{
		\includegraphics[trim=0.3cm 0.1cm 1.6cm 0cm, clip, width=0.295\columnwidth]{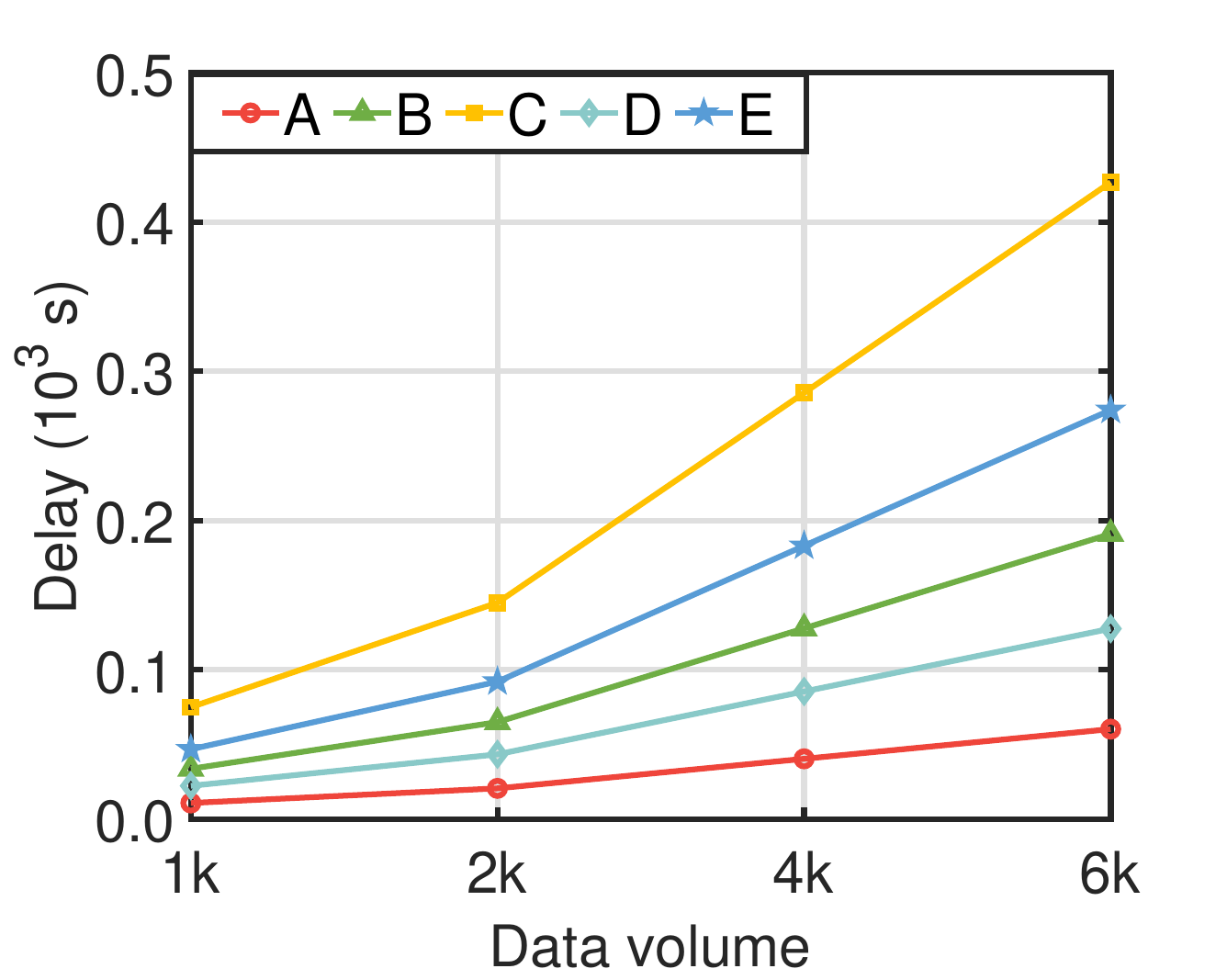}
	}
	\subfigure[Energy cost]{
		\includegraphics[trim=0.3cm 0.1cm 1.9cm 0cm, clip, width=0.295\columnwidth]{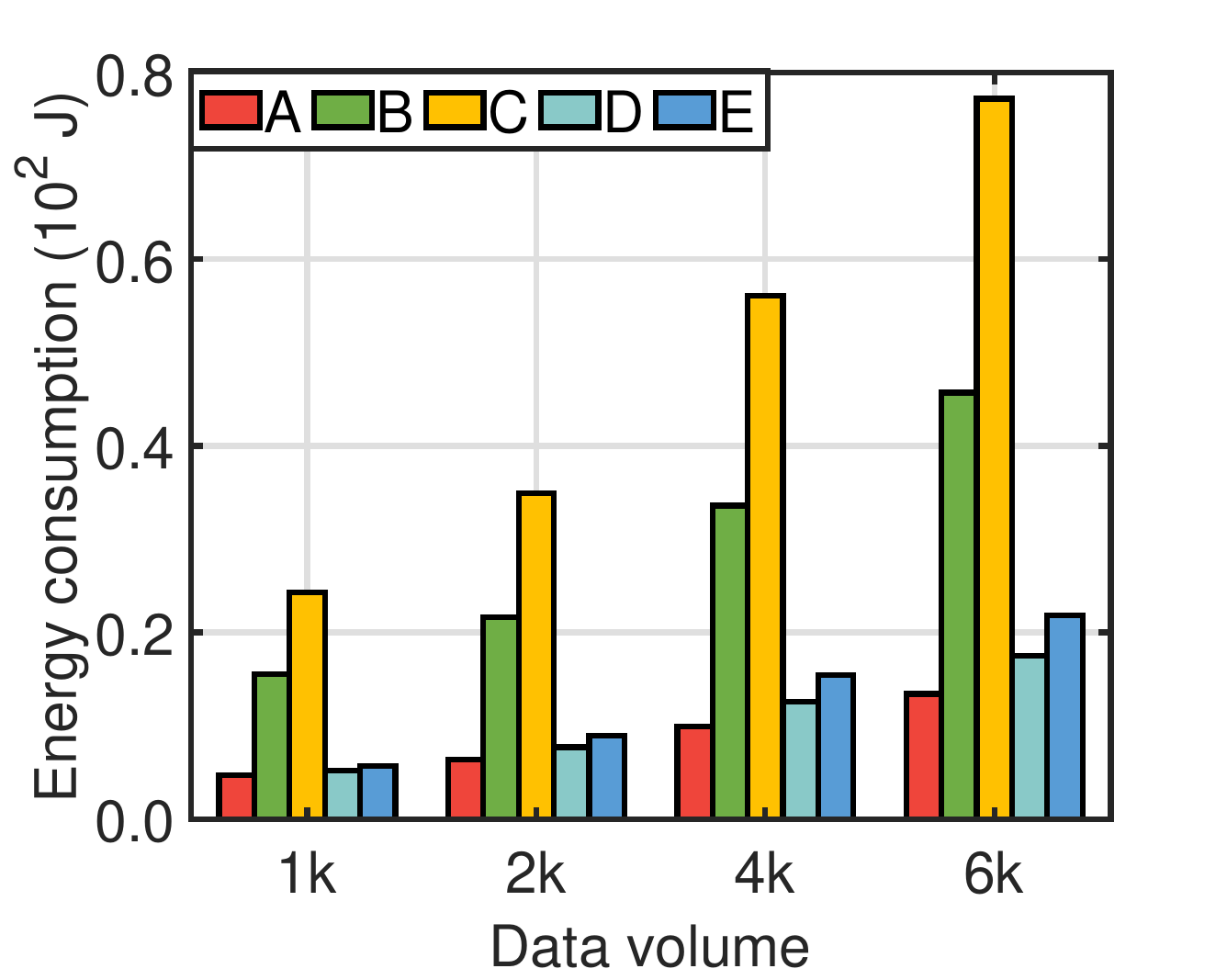}
	}

    	\subfigure[Convergence]{
		\includegraphics[trim=0.3cm 0.05cm 1.6cm 0cm, clip, width=0.295\columnwidth]{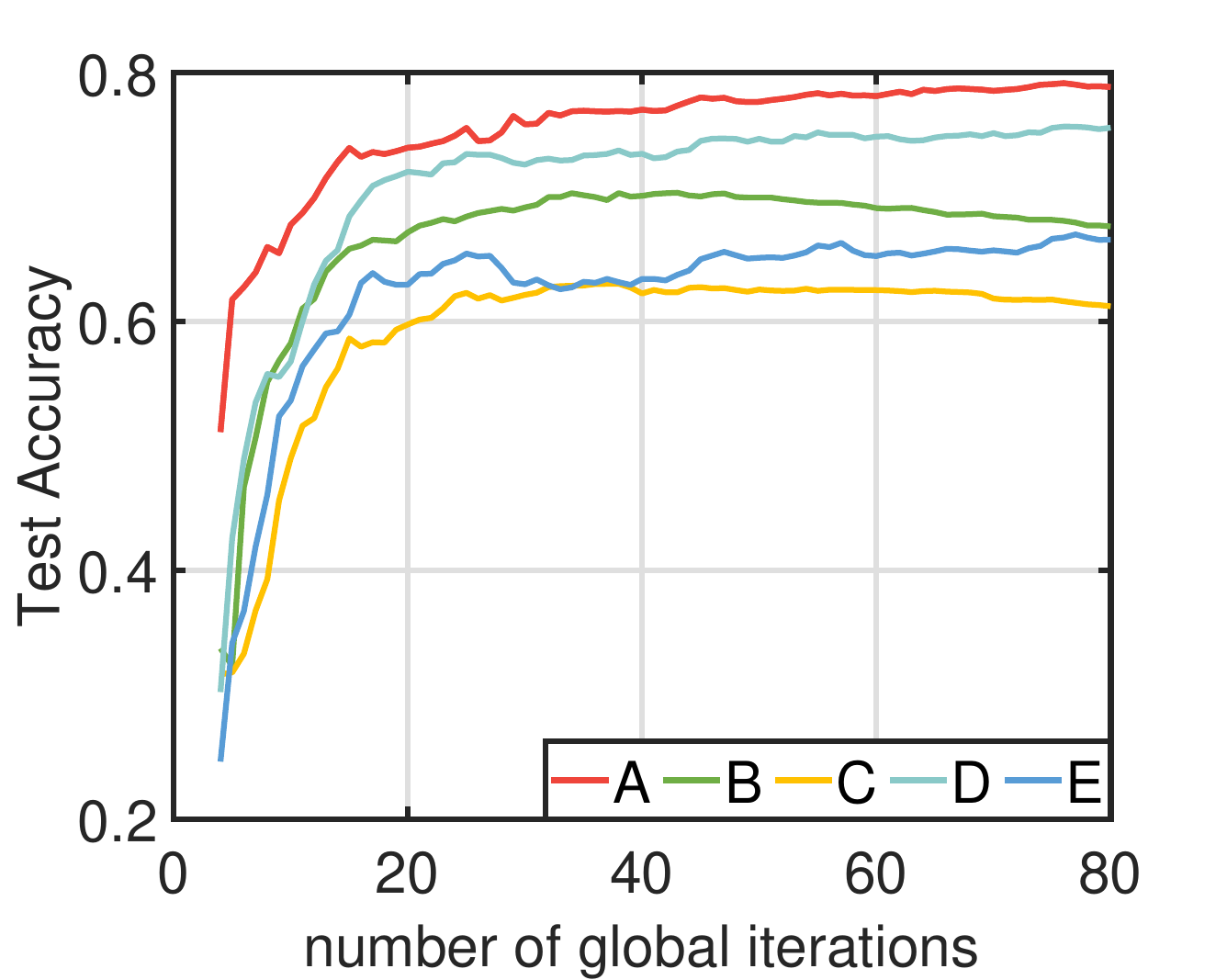}
	}
	\subfigure[Time cost]{
		\includegraphics[trim=0.3cm 0.1cm 1.6cm 0cm, clip, width=0.295\columnwidth]{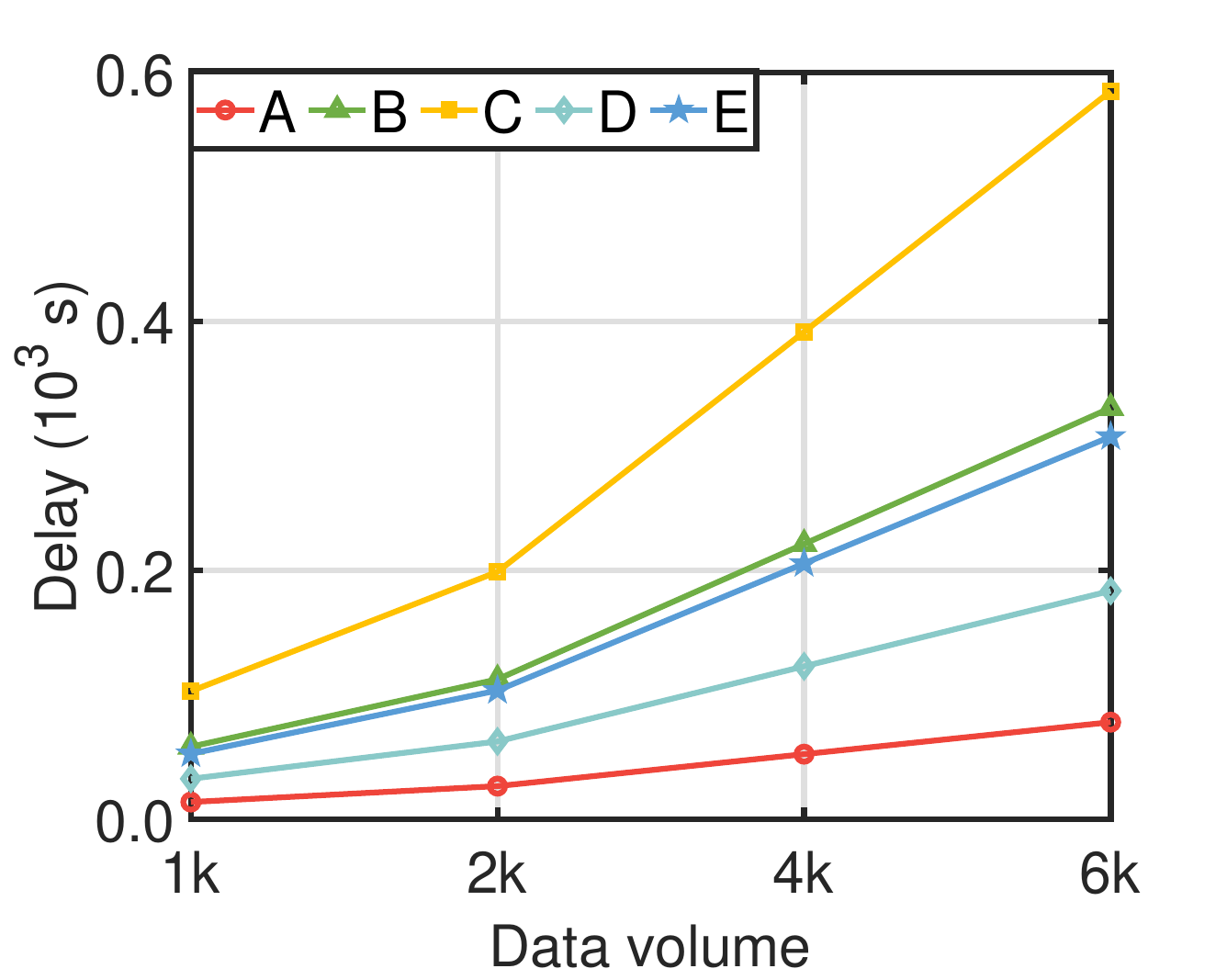}
	}
	\subfigure[Energy cost]{
		\includegraphics[trim=0.3cm 0.1cm 1.9cm 0cm, clip, width=0.295\columnwidth]{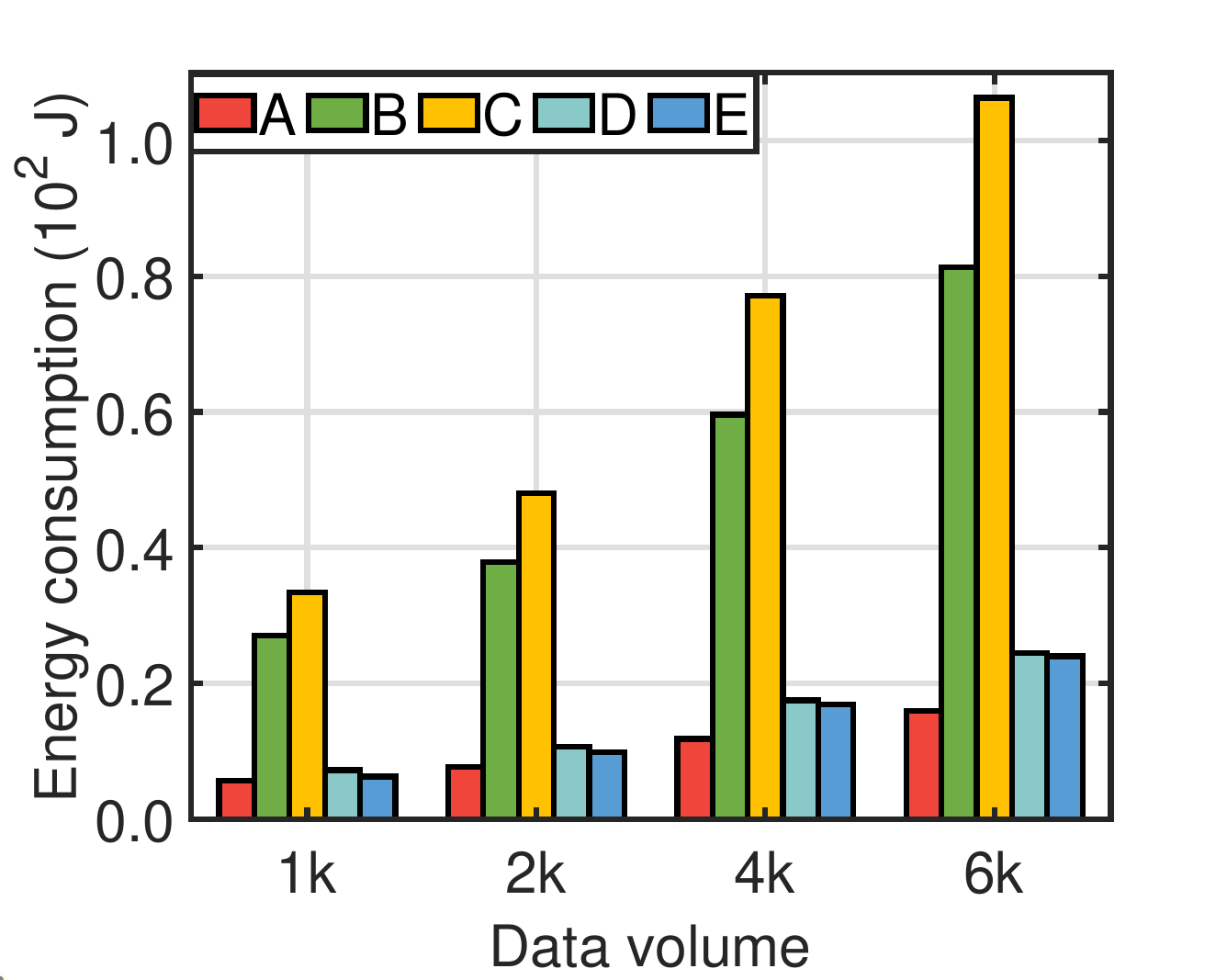}
	}
	\caption{Performance Comparisons upon having either adaptive and fixed thresholds. `A' represents the Adaptive Threshold, while `B', `C', `D', and `E' correspond to fixed thresholds set at 0.40, 0.55, 0.70, and 0.85, respectively. The comparison is conducted using the LeNet5 model on the MNIST (top row plots) and FaMNIST datasets (bottom row plots).}
	\label{fig_7}
\end{figure}

\subsubsection{Scenario-based evaluations} 
Given that our study focuses on UAV-assisted HFL scenarios, where UAVs may drop out due to energy constraints and devices with non-iid data dynamically move across different UAVs, this section presents simulations to assess CEHFed’s performance under various configurations of such factors.

\noindent \textbf{Impact of Adaptive vs. Fixed Thresholds on Convergence and Resource Utilization:} In Fig. 7, we evaluate the impact of adaptive thresholding compared to fixed thresholds in UAV-assisted HFL using LeNet5 on MNIST (top row plots) and FaMNIST (bottom row plots). Our results indicate that CEHFed with adaptive thresholding achieves faster convergence than its variations with fixed thresholds. This improvement stems from the fact that excessively high thresholds limit device participation, slowing convergence, while low thresholds allow remote or less-capable devices to participate, reducing training efficiency. For instance, when training LeNet5 on MNIST with a 4k data volume (Figs. 7(b)), the adaptive threshold (`A') reduces time costs by 68.4$\%$, 85.9$\%$, 52.7$\%$, and 78$\%$ compared to `B', `C', `D', and `E', respectively. Likewise, energy costs (Figs. 7(c)) decrease by 70.6$\%$, 82.6$\%$, 23.6$\%$, and 38.5$\%$, respectively. This demonstrates that an appropriate, UAV-specific threshold accelerates global model convergence and improves resource efficiency. The same trend is observed for LeNet5 on FaMNIST (Figs. 7(d)-(f)), revealing the effectiveness of CEHFed’s adaptive thresholding strategy.

\begin{figure}[!t]
	\centering
	\subfigure[non-iid (A)]{
		\includegraphics[trim=0.3cm 0.1cm 1.9cm 0cm, clip, width=0.295\columnwidth]{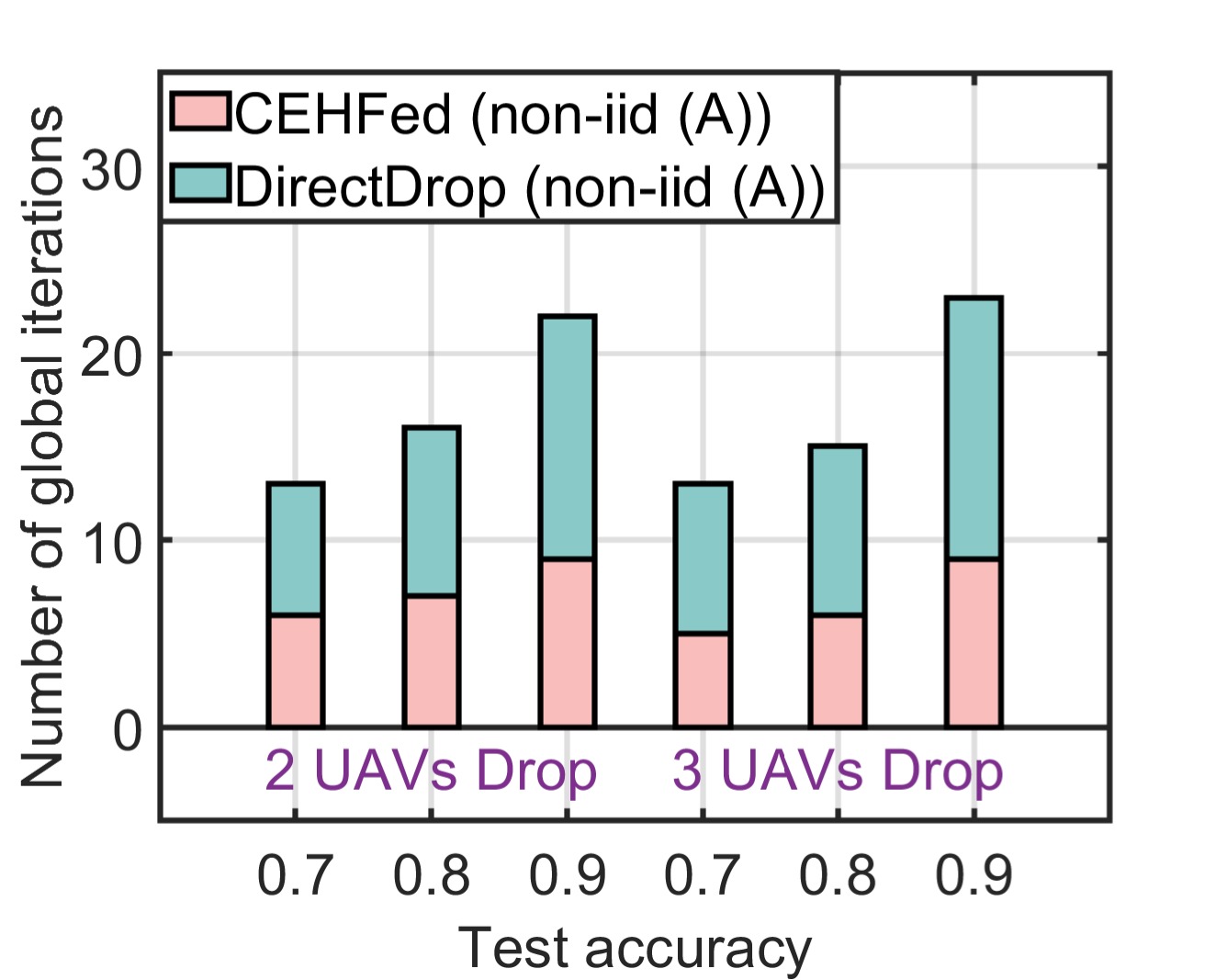}
	}
	\subfigure[non-iid (B)]{
		\includegraphics[trim=0.3cm 0.1cm 1.9cm 0cm, clip, width=0.295\columnwidth]{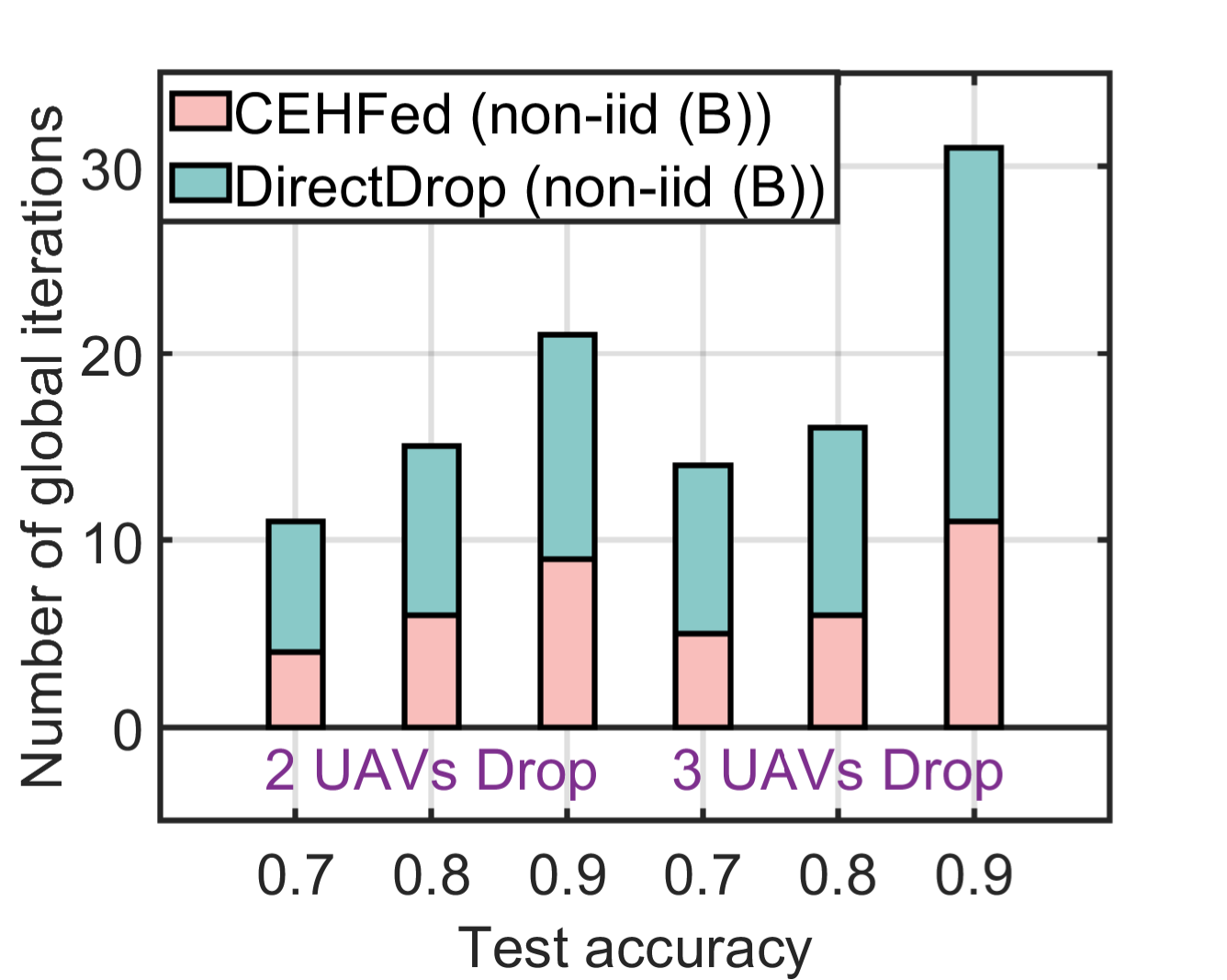}
	}
	\subfigure[non-iid (A)]{
		\includegraphics[trim=0.3cm 0.1cm 1.9cm 0cm, clip, width=0.295\columnwidth]{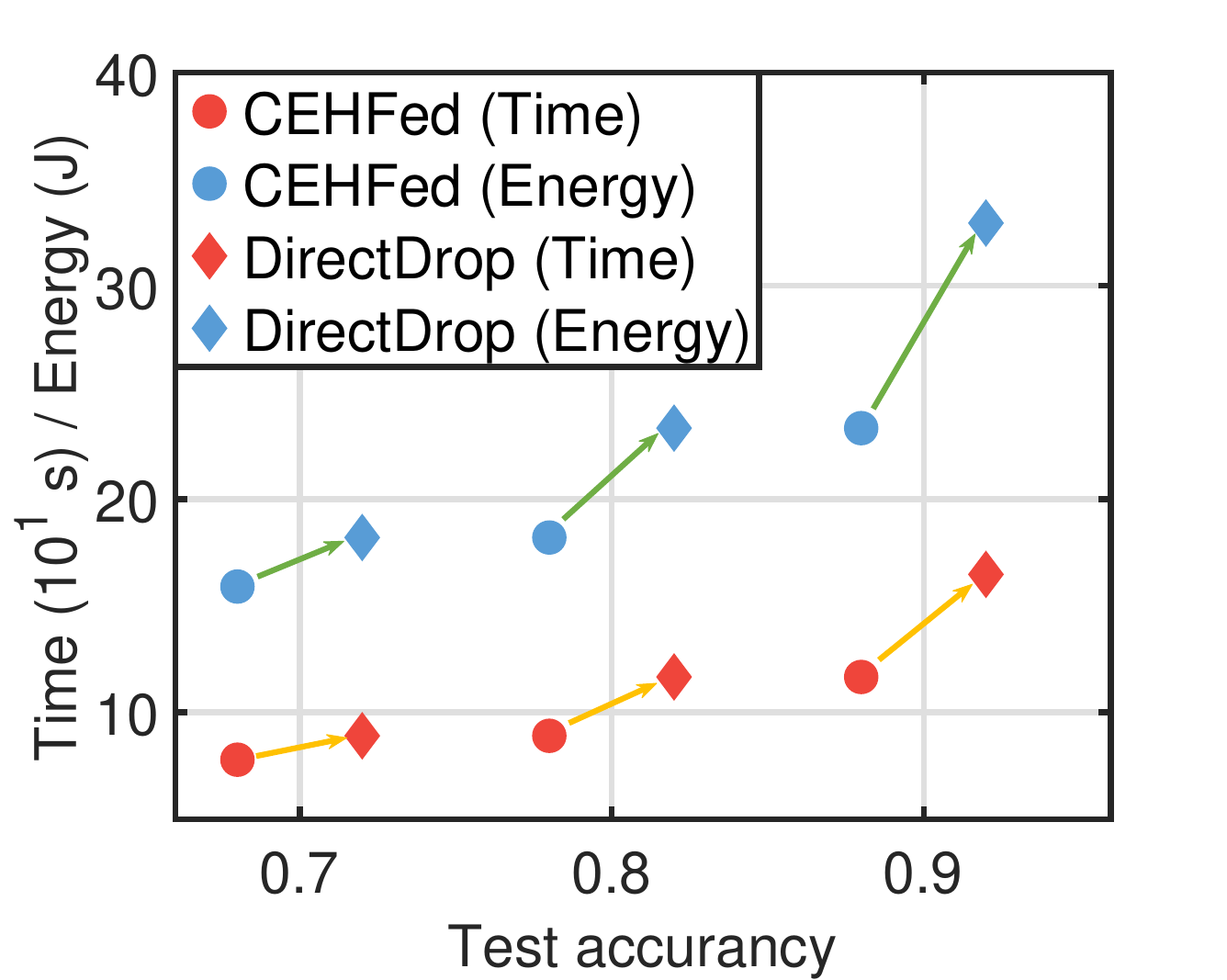}
	}
    
	\subfigure[non-iid (B)]{
		\includegraphics[trim=0.3cm 0.1cm 1.9cm 0cm, clip, width=0.295\columnwidth]{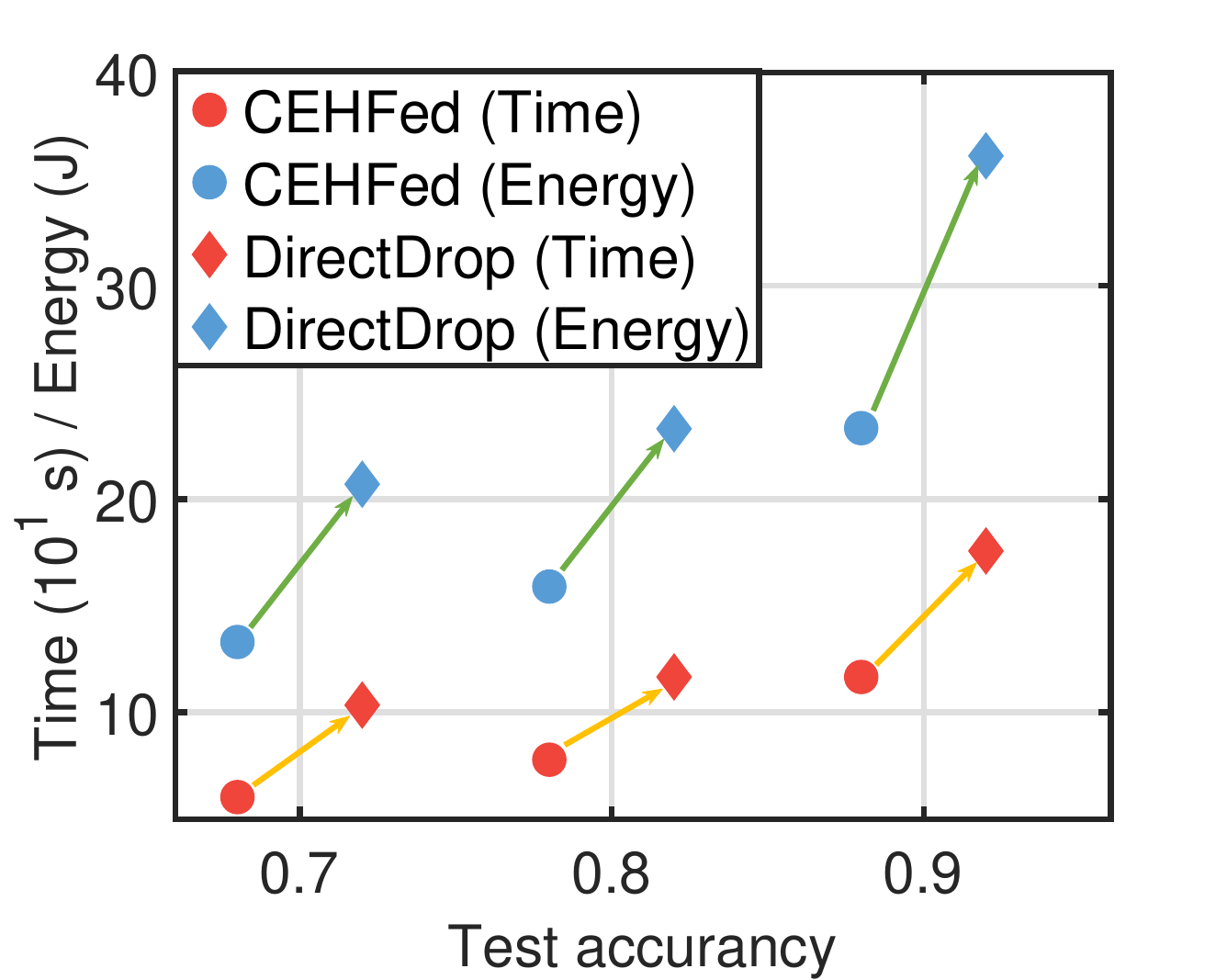}
	}
	\subfigure[Convergence]{
		\includegraphics[trim=0.3cm 0.1cm 1.9cm 0cm, clip, width=0.295\columnwidth]{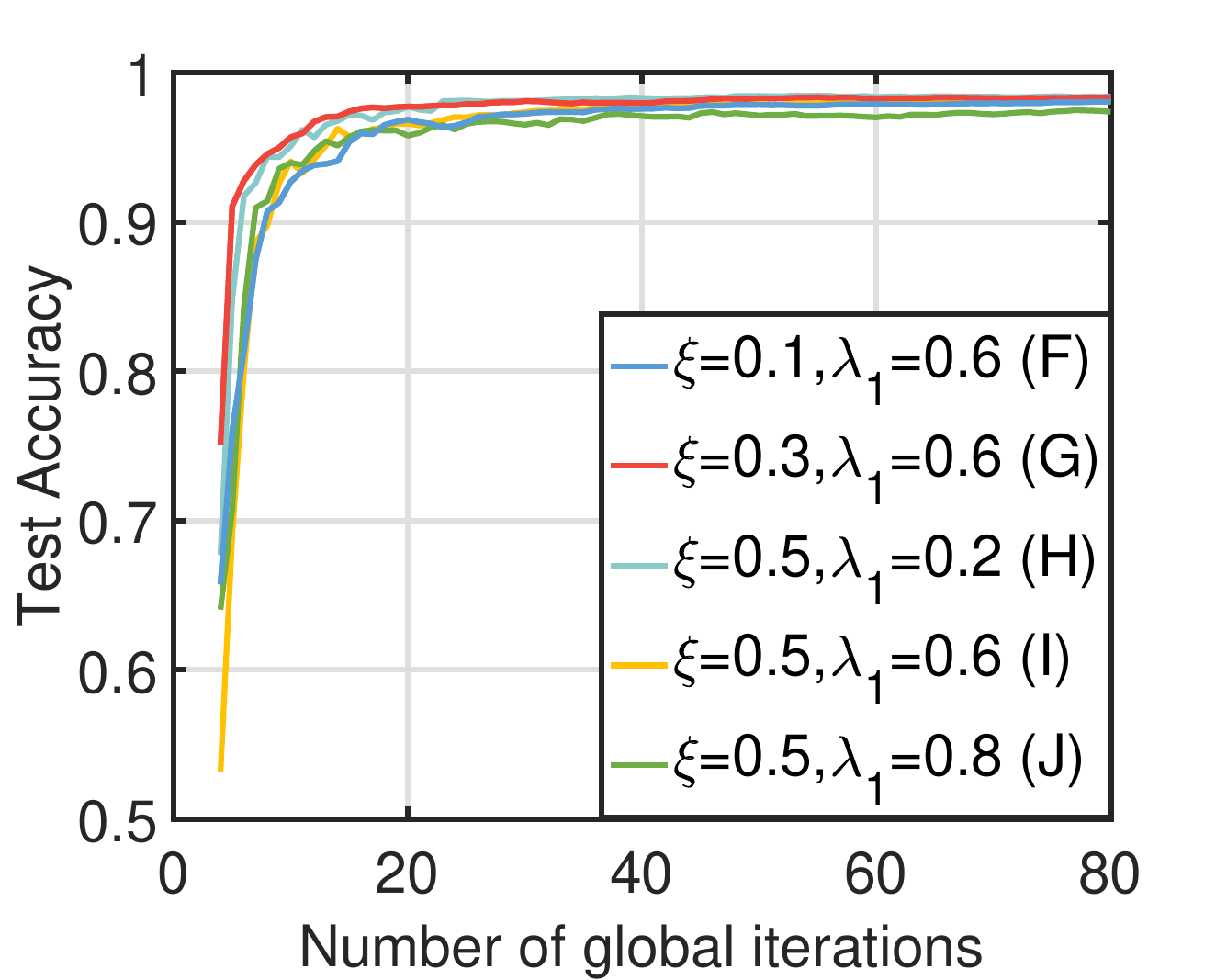}
	}
	\subfigure[Global cost]{
		\includegraphics[trim=0.3cm 0.1cm 1.9cm 0cm, clip, width=0.295\columnwidth]{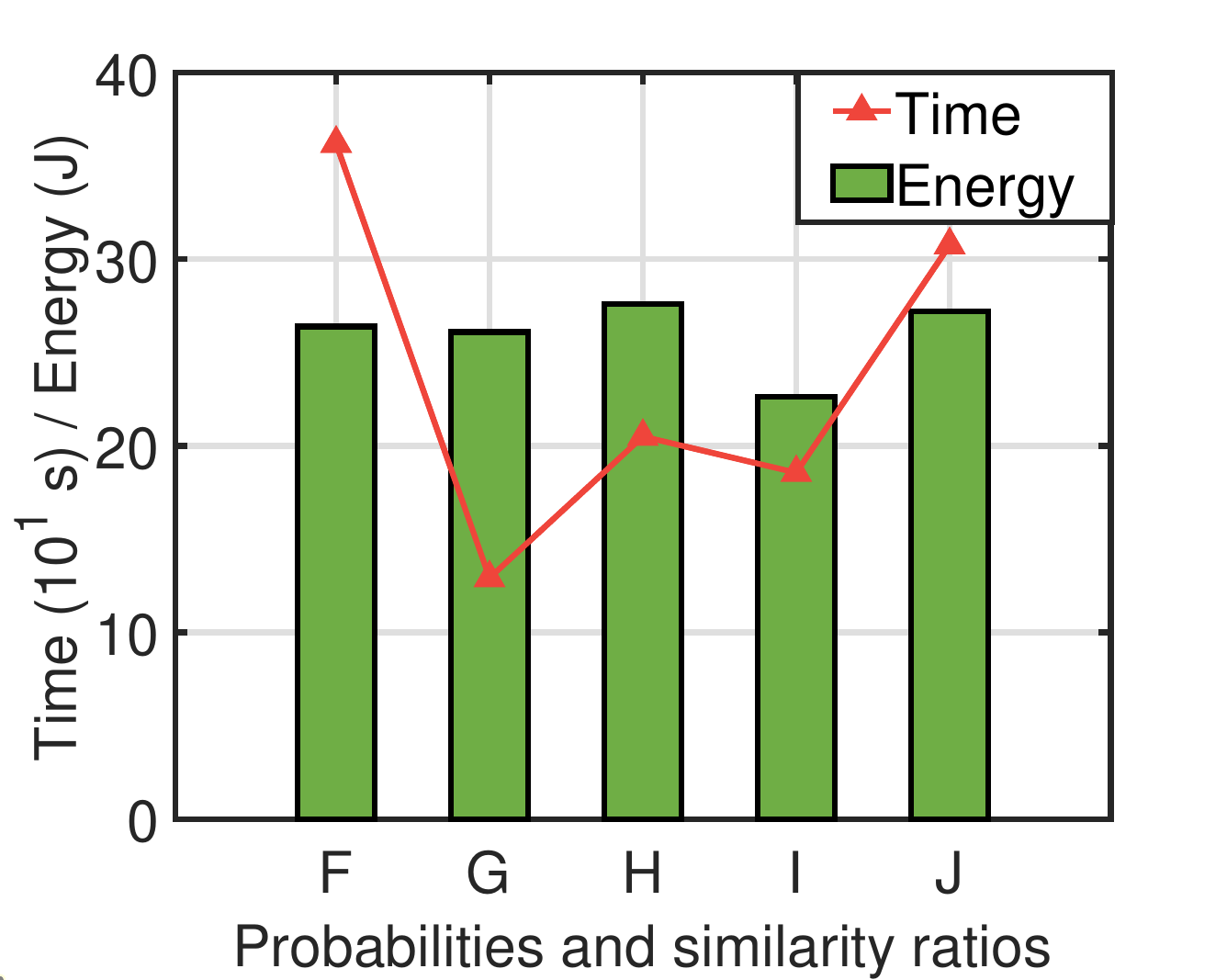}
	}
	\caption{(a), (b), (c), (d): Performance comparison in terms of the required number of global iterations, time, and energy consumption to reach various test accuracies under different data distributions. (e): The impact of different mobility patterns and the coefficient of data distribution in fitness score on model convergence. (f) The impact of different mobility patterns and the coefficient of data distribution in fitness score (the letters `F', `G', `H', `I', and `J' are specified in the legend of plot (e)) on energy and time costs.}
	\vspace{-3mm}
	\label{fig_8}
\end{figure}

\noindent \textbf{Resilience Against UAV Dropouts: Impact on Edge Iterations and Convergence:} In Fig.~8, we examine UAV dropouts by comparing CEHFed with DirectDrop, a benchmark assuming direct disconnections without mitigation. Scenarios simulate 2 or 3 UAVs dropping out among 5 under varying data distributions. As expected, higher test accuracy requires more global iterations, extending UAV flight time. In Fig.~8(a), for non-iid (A) (devices hold only two labels), CEHFed mitigates data loss, requiring fewer global iterations. The same holds for non-iid (B), where devices have more than two labels. Figs.~8(a) and 8(b) show that UAV disconnections impact convergence more as device label diversity increases: under non-iid (B), 6–12 additional global iterations are needed compared to non-iid (A). This highlights that higher data heterogeneity amplifies the disruptive effect of UAV dropouts on global model convergence.

\noindent \textbf{Reduction in Time and Energy Costs:} In Figs. 8(c) and 8(d), we analyze the time and energy costs needed to reach accuracy levels of 0.7, 0.8, and 0.9 for CEHFed vs. DirectDrop. For a 6k data volume, CEHFed reduces time and energy costs by 29.2$\%$ and 29$\%$ for non-iid (A), and 33.6$\%$ and 35.37$\%$ for non-iid (B), respectively. These results validate CEHFed’s ability to sustain efficient training even under UAV disconnections, demonstrating its superior resource utilization.

\noindent \textbf{Impact of Device Mobility and Data Distribution on Convergence:} Fig.~8(e) shows the effects of device mobility (\(\xi\)) and data distribution coefficient (\(\lambda_1\)) on convergence. Convergence for \(\xi = 0.3, \lambda_1 = 0.6\) matches \(\xi = 0.5, \lambda_1 = 0.2\), while \(\xi = 0.5, \lambda_1 = 0.6\) aligns with \(\xi = 0.5, \lambda_1 = 0.8\), indicating that higher \(\lambda_1\) slightly slows convergence at the same mobility. For the same \(\lambda_1\), higher device speed generally speeds up convergence via better data mixing (e.g., \(\xi = 0.1, \lambda_1 = 0.6\) vs. \(\xi = 0.3, \lambda_1 = 0.6\)), though excessive mobility (\(\xi = 0.5\)) disrupts convergence. Fig.~8(f) evaluates global cost under varying \(\xi\) and \(\lambda_1\). As mobility increases (`F' vs `G', `G' vs `I'), time cost first decreases then rises, while energy consumption steadily decreases. Proper \(\lambda_1\) selection can further optimize both (`H' vs `I' vs `J'), though the optimal balance requires further study.

\noindent \textbf{UAV Redeployment Performance After Disconnections:} We present the performance of the two-stage greedy algorithm, evaluating its effectiveness when one or two UAVs disconnect. See more details in \textbf{Appendix E.D and E.E}.

\section{Conclusion and Future Work}
This paper investigated a UAV-assisted HFL architecture over dynamic IoT devices, where training is constrained by UAV energy limitations. We proposed a framework that jointly optimizes learning configuration, bandwidth allocation, D2U association, global aggregator selection, and UAV redeployment to minimize global training costs while maintaining high model accuracy. Extensive simulations with real-world datasets show that \textit{CEHFed} effectively reduces global training time and energy consumption while alleviating the impact of UAV disconnections on model convergence. Future work may explore integrating satellite communication into UAV-assisted HFL to enable seamless global aggregation and enhance connectivity in remote or infrastructure-limited regions, with satellites serving as high-altitude aggregators to ensure continuous learning. Moreover, investigating dynamic asynchronous aggregation, more resilient UAV redeployment, and energy-harvesting UAVs—capable of utilizing solar, wind, or RF energy—represents a promising direction to further improve network robustness and learning resilience.


\newpage
\clearpage
\appendices

\section{}
\begin{table*}[h]
    \centering
    \scriptsize
    \vspace{-0.2 cm}
    \caption{Summary of existing related literature}
    \setlength{\tabcolsep}{2.5 pt} 
    \centering
    \vspace{-0.2 cm} 
    \begin{tabularx}{\linewidth}{|c|*{10}{>{\centering\arraybackslash}X|}}
        \hline
        \multirow{2}{*}{\textbf{Related studies}} & \multicolumn{2}{c|}{\makecell{\textbf{Device status} }} & \multicolumn{3}{c|}{\makecell{\textbf{UAV configuration}}} & \multicolumn{2}{c|}{\makecell{\textbf{Device--UAV association}}} & \multicolumn{2}{c|}{\makecell{\textbf{FL/HFL status}}} \\ \cline{2-10} 
        & I & II & III & IV & V & VI & VII & VIII & IX\\
        \hline
        [33] &  &  &  &  &  & & $\checkmark$ & & $\checkmark$\\
        \hline
        [3][29][35] & $\checkmark$ & $\checkmark$ &  & & & & & & \\
        \hline
        [2][6][13][30] & $\checkmark$ & $\checkmark$ &  & & & & & & $\checkmark$\\
        \hline
        [20][23][34][37] & & $\checkmark$ & $\checkmark$&  &  &  & & & \\
        \hline
        [22] & $\checkmark$ & $\checkmark$ & $\checkmark$& $\checkmark$ &  &  & & & \\
        \hline
        [24] & $\checkmark$ &  & &  &  &  & &$\checkmark$ & \\
        \hline
        [15] &  & $\checkmark$ & &  &  &  & $\checkmark$&$\checkmark$ & \\
        \hline
        [10][11] & $\checkmark$ & $\checkmark$ & &  &  & $\checkmark$ & & & \\
        \hline
        Our Paper& $\checkmark$ & $\checkmark$ & $\checkmark$  & $\checkmark$ & $\checkmark$ & $\checkmark$ & $\checkmark$ & $\checkmark$ & $\checkmark$\\
        \hline
    \end{tabularx}
    \label{relate work comparison}
\end{table*}
\subsection{Summary of Existing Literature - Table I}
In Table I, we have different numbers with various means, I--Dynamism in the network environment; II--Heterogeneity (e.g., the data distribution or computational resources of device); III--Dynamic deployment of UAVs; IV--Energy constraints, particularly referring to UAV disconnection due to insufficient battery power; V--Dynamic global aggregation mechanisms; VI--Multi-dimensional evaluation of client devices; VII--Adaptive association between UAVs and clients; VIII--Bandwidth optimization; IX--Optimization of local, edge, and global iterative parameters.

\subsection{Communication Types and Their Modeling}
In this paper, we define three different types of communication. Specifically, for D2U communication between device $n$ and active UAV $m$, the data rate during the $g^\text{th}$ global iteration is
$
	r^{\mathsf{D2U}}_{n \to m;[g]} = B^{\mathsf{D2U}}_{m,n;[g]} \log_2 \left(1 + S^{\mathsf{D2U}}_{n \to m;[g]} \right)
$,
where the signal-to-noise ratio (SNR) is 
$
	S^{\mathsf{D2U}}_{n \to m;[g]} = \frac{p^{\mathsf{D2U}}_n d_{m,n;[g]}^{-\alpha_{\mathsf{D2U}}}}{N_0B^{\mathsf{D2U}}_{m,n;[g]} }
$.
Here, $B^{\mathsf{D2U}}_{m,n;[g]}$ represents the amount of uplink bandwidth of UAV $m$ allocated to device $n$, $p^{\mathsf{D2U}}_n$ denotes the transmit power of device $n$, $\alpha_{\mathsf{D2U}}$ captures the D2U link path loss exponent, and $N_0$ is the noise power spectral density. Similarly, for U2D communication, the downlink data rate is given by
$
	r^{\mathsf{U2D}}_{m \to n;[g]} = B^{\mathsf{U2D}}_{m,n;[g]} \log_2 \left(1 + S^{\mathsf{U2D}}_{m \to n;[g]} \right)
$,
where the SNR is
$
	S^{\mathsf{U2D}}_{m \to n;[g]} = \frac{p^{\mathsf{U2D}}_m d_{m,n;[g]}^{-\alpha_{\mathsf{U2D}}}}{N_0  B^{\mathsf{U2D}}_{m,n;[g]}}.
$	
Here, $B^{\mathsf{U2D}}_{m,n;[g]}$ represents the downlink bandwidth allocated to device $n$, $p^{\mathsf{U2D}}_m$ denotes the transmit power of UAV $m$, and $\alpha_{\mathsf{U2D}}$ describes the U2D link path loss exponent. Finally, for U2U communication, the data rate is given by
$
	r^{\mathsf{U2U}}_{m \to m';[g]} = B^{\mathsf{U2U}}_{m,m';[g]} \log_2 \left(1 + S^{\mathsf{U2U}}_{m \to m';[g]} \right)
$,
where the SNR is:
$
	S^{\mathsf{U2U}}_{m \to m';[g]} = \frac{p^{\mathsf{U2U}}_m d_{m,m';[g]}^{-\alpha_{\mathsf{U2U}}}}{N_0B^{\mathsf{U2U}}_{m,m';[g]}  }
$.
Here, $B^{\mathsf{U2U}}_{m,m';[g]}$ represents the bandwidth allocated for communication between UAV $m$ and UAV $m'$, and $\alpha_{\mathsf{U2U}}$ represents the U2U link path loss exponent.

\subsection{Overall Procedure of The HFL of Our Interest}
We present a detailed description of the proposed HFL framework in Alg. 1, outlining its key components and procedural steps to ensure clarity and reproducibility.
\begin{algorithm}[]
	{\footnotesize  
		\caption{\small{The HFL procedure of our interest}}
		{\bf{Input :}} 
		$K_{[g]}$, $H$, $\eta$, $E^{\mathsf{Batt}}_{m;[0]}$;
		
		{\bf{Output :}} 
		$w_{[g]}$;
		
		{\bf{Initialization :}} 
		$ w_{n}=w_{[0]}$;
		\LinesNumbered 
		
		\If{the global model parameter changes does not satisfy \eqref{4}}{
			
			\textit{\textbf{Role 1. process:}} // running at each online UAV
			
			\For{
				each edge communication round $k=\{1,2,\cdots,K_{[g]}\}$
			}{
				\For{
					each UAV $m\in\mathcal{M}_{[g]}$ 
				}{
					Obtain $\mathcal{N}_{m;[g]}^\mathsf{Sel}$ through Alg. 3.	
					
					\textit{\textbf{Device process:}} // running at each device	
					
					\For{
						$n\in\mathcal{N}_{m;[g]}^\mathsf{Sel}$ in parallel
					}{
						Refresh the device's local model parameters: $w_{n;[g,k,0]}^\mathsf{Dev} \xleftarrow{} w_{m;[g,k-1]}^\mathsf{UAV}$.
						
						\For{for $h=\{1,2,\cdots,H\}$}{
							Update $w_{n;[g,k,h]}^\mathsf{Dev}$ based on \eqref{1}.
						}
						
						Upload $w_{n;[g,k,H]}^\mathsf{Dev}$ to the associated UAV.
						
					}
					
					UAV aggregation of model parameters from all devices according to \eqref{2}.
					
				}
            }
				
			\textit{\textbf{Role 2. process:}} // running at global aggregator UAV
					
			\For{$m \in \mathcal{M}_{[g]}$}{
						Move to the appropriate position based on Alg. 4.
					}

			According to Alg. 4, select global aggregator UAV and aggregate global model based on \eqref{3}.
					
			Global aggregator UAV broadcast global model $w_{[g]}$ to online UAVs and they continue to broadcast to devices.
					
			Update the battery capacity $E^{\mathsf{Batt}}_{m;[g]}$.
					
	}
}					
\end{algorithm}

\section{}
\subsection{Proof of Theorem 1}
\noindent \textit{Proof of Theorem 1:} Convex functions have the following properties:

\noindent $\bullet$ \textbf{Additivity property:} If $f_1,f_2, ..., f_I$ is a convex function, then their linear combination (i.e., weighted sum) is also a convex function.

\noindent $\bullet$ \textbf{Maximum property:} If $f_1,f_2, ..., f_I$ is a convex function, then their pointwise maximum function $\text{Max}(f_1,f_2,...,f_I)$ is also a convex function.

Combined with \eqref{30}, we only need to prove that the following function is a convex function
\setcounter{equation}{56}
\begin{align}
    f(H,B_{m,n;[g]}^\mathsf{D2U},B_{m,n;[g]}^\mathsf{U2D})=\underbrace{HC_n}_{\textcircled{\scriptsize \scriptsize 1}}+ \notag\\
    \underbrace{\frac{A_n^\mathsf{D2U}}{B_{m,n;[g]}^\mathsf{D2U}\text{log}_2(1+\frac{\mathcal{A}_{m,n}^\mathsf{D2U}}{B_{m,n;[g]}^\mathsf{D2U}})}}_{\textcircled{\scriptsize 2}} + \underbrace{\frac{A_m^\mathsf{U2D}}{B_{m,n;[g]}^\mathsf{U2D}\text{log}_2(1+\frac{\mathcal{A}_{m,n}^\mathsf{U2D}}{B_{m,n;[g]}^\mathsf{U2D}})}}_{\textcircled{\scriptsize 3}}. \label{77}
\end{align}

Since $H$ and $C_n$ are positive values, $\textcircled{\scriptsize 1}$ is a convex function, and the expressions of $\textcircled{\scriptsize 2}$ and $\textcircled{\scriptsize 3}$ are similar, so we show the convexity of $\textcircled{\scriptsize 2}$ using a similar approach to which the convexity of  $\textcircled{\scriptsize 3}$ can be proved. We choose $B_{m,n;[g]}^\mathsf{D2U}$ from \eqref{77} as the variable $x$, build a new function $f(x)$, and show its convexity in the following. Firstly, we set $c_1=A_n^\mathsf{D2U}$, and $c_2=\mathcal{A}_{m,n}^\mathsf{D2U}$ to simplify the calculation. Then, we calculate the first-order derivative of $\textcircled{\scriptsize 2}$ as follows:
\begin{align}
    f(x)&=\frac{c_1}{x\text{log}_2(1+\frac{c_2}{x})}= \frac{c_1\ln2}{x\text{ln}(1+\frac{c_2}{x})},\\
    f'(x) &= \frac{(c_1\ln2)'x\ln(1+\frac{c_2}{x})-(c_1\ln2)(x\ln(1+\frac{c_2}{x}))'}{(x\ln(1+\frac{c_2}{x}))^2}\notag\\
    &= - \frac{(c_1\ln2)(x\ln(1+\frac{c_2}{x}))'}{(x\ln(1+\frac{c_2}{x}))^2} \notag\\
    &= - \frac{(c_1\ln2)(\ln(1+\frac{c_2}{x})-\frac{c_2}{x+c_2})}{(x\ln(1+\frac{c_2}{x}))^2}. 
\end{align}

While $x > 0$, $f'(x)$ is always less than 0, so $f(x)$ is monotonically decreasing. Next, we set $\vmathbb{g}(x) = \ln(1+\frac{c_2}{x})$ and $\vmathbb{h}(x)=\frac{c_2}{x+c_2}$, and obtain the second-order derivative of $\textcircled{\scriptsize 2}$ as follows
\begin{align}
    f''(x) &= - \frac{(c_1\ln2)(\vmathbb{g}(x)-\vmathbb{h}(x))}{(x\vmathbb{g}(x))^2} \notag\\
    &= \frac{d}{d_x}\frac{-(c_1\ln2)(\vmathbb{g}(x)-\vmathbb{h}(x))}{(x\vmathbb{g}(x))^2} \notag\\
    &= \frac{u'v-uv'}{v^2}, 
\end{align}
where $u= (c_1\ln2)(\vmathbb{g}(x)-\vmathbb{h}(x)),v=(x\vmathbb{g}(x))^2$, and their corresponding derivatives are given by
\begin{align}
    \begin{cases}
    u' = \frac{d}{d_x}(u)=c_1\ln2\cdot(\vmathbb{g}'(x)-\vmathbb{h}'(x)),\\
    v' = \frac{d}{d_x}(v)=2(x\vmathbb{g}(x))(\vmathbb{g}(x)+x\vmathbb{g}'(x)),\\
    \vmathbb{g}'(x) = -\frac{c_2}{x^2+c_2x}, \vmathbb{h}'(x)=-\frac{c_2}{(x+c_2)^2}.  
    \end{cases}
\end{align}

Thus, we can get
\begin{align}
     f^{\prime\prime}(x) = \frac{ -c_1 \ln 2 \cdot \left( \vmathbb{g}^{\prime}(x) - \vmathbb{h}^{\prime}(x) \right)}{ \left( x \vmathbb{g}(x) \right)^2 } +\notag \\
    \frac{ 2 c_1 \ln 2 \cdot \left( \vmathbb{g}(x) - \vmathbb{h}(x) \right) \left( \vmathbb{g}(x) + x\vmathbb{g}^{\prime}(x) \right) }{ \left( x \vmathbb{g}(x) \right)^3 }.
\end{align}

Replacing $x$ back with $B^{\mathsf{D2U}}_{m,n;[g]}$, noting that $B^{\mathsf{D2U}}_{m,n;[g]}\geq 0$ (based on \eqref{27a}), $f^{\prime\prime}(B_{m,n;[g]}^\mathsf{D2U})$ is always greater than 0, so  $\textcircled{\scriptsize 2}$ is a convex function. Similarly, we can also prove that $\textcircled{\scriptsize 3}$ is a convex function, and combined with the properties of convex functions, we can prove that $\mathcal{P}_{2b}$ is a convex problem.

\subsection{Proof of Theorem 2}
\noindent \textit{Proof of Theorem 2:} To determine the optimal value of the variable $\mathcal{Y}^{\langle j\rangle}$ while keeping the variable $H^{\langle j\rangle}$ fixed, we  take the derivative of the function with respect to $\mathcal{Y}^{\langle j\rangle}$ and identify its critical points. Specifically, we have
\begin{align}
	\frac{df(\mathcal{Y}^{\langle j\rangle})}{d\mathcal{Y}^{\langle j\rangle}}=\upsilon^{\langle j\rangle}+\sigma^{\langle j\rangle}\left\{\mathcal{G}(H^{\langle j\rangle})-\mathcal{Y}^{\langle j\rangle}\right\}. 
\end{align}
Setting the above derivative to $0$, we get
\begin{align}
	\begin{cases}\upsilon^{\langle j\rangle}+\sigma^{\langle j\rangle}\left\{\mathcal{G}(H^{\langle j\rangle})-\mathcal{Y}^{\langle j\rangle}\right\}=0,\\
		\mathcal{Y}^*=-\frac{\upsilon^{\langle j\rangle}}{\sigma^{\langle j\rangle}}-\mathcal{G}(H^{\langle j\rangle}). 
	\end{cases}
\end{align}

From the above, the corresponding critical point $\mathcal{Y}^{*}=-\frac{\upsilon^{\langle j\rangle}}{\sigma^{\langle j\rangle}}-\mathcal{G}(H^{\langle j\rangle})$ can be obtained, which, however, may be unable to satisfy the non-negative constraint, namely, $\mathcal{Y}^{\langle j\rangle}\geq0$. Accordingly, we consider the following two cases:

\noindent $\bullet$ If $\mathcal{Y}^{*}\geq0$, meaning that $-\frac{\upsilon^{\langle j\rangle}}{\sigma^{\langle j\rangle}}-\mathcal{G}(H^{\langle j\rangle})\geq0$. Then, at the point $\mathcal{Y}^{\langle j\rangle}=\mathcal{Y}^{*}$, we get the optimal solution.

\noindent $\bullet$ If $\mathcal{Y}^{*}<0$, the optimal solution for $\mathcal{Y}^{\langle j\rangle}$ is 0 because $\mathcal{Y}^{\langle j\rangle}$ can not be negative.

The above discussions imply the proof of \textit{Theorem 2}.

\subsection{Derivation Related to $\mathcal{P}_{1}$} 
\noindent \textbf{(I) Constraint Violation:} When the precision constraint is met, we further compute the following function that characterizes the constraint violation degree of $H^{\langle j+1\rangle}$:
\begin{equation}
	\psi^{\langle j\rangle}(H^{\langle j+1\rangle})=\sqrt{\max\Big\{\mathcal{G}(H^{\langle j+1\rangle}),-\frac{\upsilon^{\langle j\rangle}}{\sigma^{\langle j\rangle}}\Big\}^2}.
\end{equation}
We then determine the relationship between the constraint violation $\varepsilon^{\langle j\rangle}$ and the initial constraint violation $\varepsilon^{\langle 0\rangle}$ as follows.

\noindent \textbf{(II) Constraint Violation is Acceptable:} If the solution $H^{\langle j+1 \rangle}$ meets $\psi^{\langle j\rangle}(H^{\langle j+1\rangle}) \leq \varepsilon^{\langle j\rangle}$ and $\psi^{\langle j\rangle}(H^{\langle j+1\rangle}) \leq \varepsilon^{\langle 0\rangle}$, then the algorithm terminates.

\noindent \textbf{(III) Constraint Violation is Unacceptable:} We consider two cases: \textbf{(Case 1)} If $H^{\langle j+1 \rangle}$ meets $\psi^{\langle j\rangle}(H^{\langle j+1\rangle}) \leq \varepsilon^{\langle j\rangle}$ but $\psi^{\langle j\rangle}(H^{\langle j+1\rangle}) > \varepsilon^{\langle 0\rangle}$, we update the multiplier $\upsilon^{\langle j \rangle}$ and keep the penalty factor $\sigma^{\langle j \rangle}$ unchanged to approach the optimal solution. To determine how \( \upsilon^{\langle j+1 \rangle} \) is updated, we first revisit $f(H)$ and $\mathcal{P}_{1e}$, and note that the optimal solution for $f(H)$, represented by $H^{*}$, ${\mathcal{Y}}^{*}$, and multiplier $\upsilon^{*}$, should fulfill Karush-Kuhn-Tucker (KKT) conditions. Similarly, for $\mathcal{P}_{1c}$, the solutions $H^{\langle j+1 \rangle}$ and ${\mathcal{Y}}^{\langle j+1 \rangle}$ are required to satisfy KKT conditions, as given by 
\begin{gather}
    \nabla f(H^*)+\upsilon^*\nabla \mathcal{G}(H^*)=0, \label{64}\\
    \max\left\{-\frac{\upsilon^{\langle j \rangle}}{\sigma^{\langle j \rangle}}-\mathcal{G}(H^{\langle j+1 \rangle}),0 \right\}=\mathcal{Y}^{\langle j+1 \rangle},  \\
    \hspace{-3mm}\nabla f(H^{\langle j+1 \rangle}){+}(\upsilon^{\langle j \rangle}+\sigma^{\langle j \rangle}(\mathcal{G}(H^{\langle j+1 \rangle}){+}\mathcal{Y}^{\langle j+1 \rangle}))\nabla \mathcal{G}(H^{\langle j+1 \rangle}){=}0.\label{66}
\end{gather}
Combining \eqref{64}-\eqref{66}, the update rules for \( \upsilon^{\langle j+1 \rangle} \) and \( \sigma^{\langle j+1 \rangle} \) are 
\begin{align}
    \upsilon^{\langle j+1 \rangle} &= \max\left\{\upsilon^{\langle j \rangle}+\sigma^{\langle j \rangle}\mathcal{G}(H^{\langle j+1 \rangle}),0\right\}, \label{67}\\
    \sigma^{\langle j+1 \rangle}&=\sigma^{\langle j \rangle}. \label{68}
\end{align}
Beside, we also adjust $\kappa^{\langle j+1 \rangle}$ and $\varepsilon^{\langle j+1 \rangle}$ to better approximate the optimal solution  according to Case i conditions below.
\begin{equation}
    \kappa^{\langle j+1 \rangle}=\begin{cases}\frac{\kappa^{\langle j \rangle}}{\sigma^{\langle j+1 \rangle}},&\text{Case i},\\
    \frac{1}{\sigma^{\langle j+1 \rangle}},&\text{Case ii},\end{cases},~~~~~\varepsilon^{\langle j+1 \rangle}=\begin{cases}\frac{\varepsilon^{\langle j \rangle}}{(\sigma^{\langle j+1 \rangle})^{-\zeta_2}},&\text{Case i},\\
    \frac{1}{(\sigma^{\langle j+1 \rangle})^{-\zeta_1}},&\text{Case ii},\end{cases}, \label{69}
\end{equation}
where $\zeta_1$, $\zeta_2$ represents two constants and $\zeta_1 \neq \zeta_2$. \textbf{(Case 2)} If $\psi^{\langle j \rangle}(H^{\langle j+1 \rangle}) > \varepsilon^{\langle j \rangle}$, we choose to keep the multiplier unchanged and update the penalty factor to better approximate the optimal solution, that is, 
\begin{align}
    \upsilon^{\langle j+1 \rangle} &= \upsilon^{\langle j \rangle}, \label{70}\\
    \sigma^{\langle j+1 \rangle}&=\rho\sigma^{\langle j \rangle},\label{71}
\end{align}
where $\rho \in [2,10]$ represents a fixed constant. At this time, $\kappa^{\langle j+1 \rangle}$ and $\varepsilon^{\langle j+1 \rangle}$ will also be adjusted according to Case ii conditions in \eqref{69}.

\subsection{Proof of Theorem 3}
\textit{Proof of Theorem 3:} Proving the convergence of the augmented Lagrangian function is based on the following foundations:

\noindent $\bullet$ Objective function $f(H)$ and the constraint function $\mathcal{G}(H)$ are continuously differentiable.

\noindent $\bullet$ The problem has a feasible solution and satisfies constraint normative conditions (e.g., linear independence constraint norm (LICQ)).

\noindent $\bullet$ The augmented Lagrangian function can find the global minimum in each iteration.

Revisiting \textbf{Appendix B.A}, we can see that $f(H)$ and $\mathcal{G}(H)$ are continuously differentiable, and $f(H)$ is a convex function, so the augmented Lagrangian function can find the global minimum in each iteration. Besides, $\nabla\mathcal{G}(H)$ is always less than 0, which means that at the feasible point $H^*$, $\nabla\mathcal{G}(H^*)\neq0$ always holds. Now that the above foundations are verified, we turn into proving the convergence of the augmented Lagrangian function, which requires verification of four conditions\cite{Rate-of1}:

\noindent \textbf{(I) Monotonicity:} We use the gradient descent method and update the Lagrange multiplier to approach the minimum value $H{\langle j\rangle}$ and $\mathcal{Y}^{\langle j\rangle}$. Also, through \textbf{Appendix B.B} and \eqref{40}, \eqref{41}, it can be obtained that the value of the augmented Lagrangian function decreases monotonically during the iteration process, that is, it satisfies:
\begin{align}
    \mathcal{L}_{\sigma^{\langle j\rangle}}(H^{\langle j+1\rangle},\mathcal{Y}^{\langle j+1\rangle},\upsilon^{\langle j+1\rangle})\leq\mathcal{L}_{\sigma^{\langle j\rangle}}(H^{\langle j\rangle},\mathcal{Y}^{\langle j\rangle},\upsilon^{\langle j\rangle}),
\end{align}

\noindent \textbf{(II) Boundedness:} From \textbf{(I)}, since the value of the augmented Lagrangian function is monotonically decreasing, and $f(H)$ and $\mathcal{G}(H)$ are continuous, we can naturally deduct the boundedness of $H^{\langle j\rangle}$, $\mathcal{Y}^{\langle j\rangle}$ and $\upsilon^{\langle j\rangle}$.

\noindent \textbf{(III) Limit point satisfies the KKT conditions:} To verify that any limit point $H^{\langle j\rangle},\mathcal{Y}^{\langle j\rangle},\upsilon^{\langle j\rangle}$ satisfies the KKT conditions, it is necessary to satisfy both the gradient condition and the feasibility condition, which are defined as
\begin{align}
    \begin{cases}\nabla f(H^*)+\upsilon^*\nabla \mathcal{G}(H^*)=0,\\
    \mathcal{G}(H^*) = \mathcal{Y}^*.
    \end{cases}
\end{align}

Since for $H^{\langle j+1\rangle},\mathcal{Y}^{\langle j+1\rangle}$ is the minimum value of $\mathcal{L}_{\sigma^{\langle j\rangle}}$, we can obtain that
\begin{align}
    \nabla f(H^{\langle j+1 \rangle})+(\upsilon^{\langle j \rangle}+\sigma^{\langle j \rangle}(\mathcal{G}(H^{\langle j+1 \rangle})+\mathcal{Y}^{\langle j+1 \rangle}))\nabla \mathcal{G}(H^{\langle j+1 \rangle})=0.
\end{align}

Besides, while $j \xrightarrow[]{} +\infty$, we get that $\mathcal{G}(H^{\langle j+1 \rangle})-\mathcal{Y}^{\langle j+1 \rangle} \xrightarrow[]{} 0$. Thus, the KKT conditions are satisfied.

\noindent \textbf{(IV) Convergence speed:} Under appropriate conditions (e.g., $\sigma^{\langle j+1 \rangle}$ is large enough), the augmented Lagrangian method has a linear convergence rate, which is because the Hessian of the augmented Lagrangian is positive definite near the solution and the increase of $\sigma^{\langle j+1 \rangle}$ accelerates the decay of the optimality gap/error.

The above discussions prove the convergence of the augmented Lagrangian algorithm with penalty term. Besides, by consulting \cite{complexityy}, We can derive that, for the general convex case, Alg. 2 has a time complexity of $\mathcal{O}(\frac{1}{\varepsilon^{3/2}})$ and a space complexity of $\mathcal{O}(M \times N)$.

\subsection{Local Iteration and Bandwidth Optimization Algorithm}
Alg. 2 shows the steps for optimizing local iterations and bandwidth in $\mathcal{P}_1$, where formulas \eqref{67}, \eqref{69} and \eqref{71} can be found in \textbf{Appendix B.C}.
\begin{algorithm}[]
	{\footnotesize 
		\caption{\small{Penalized Augmented Lagrangian Method for Local Iteration and Bandwidth Optimization (PALM-BLO)}}
		{\bf{Input :}} 
		$H$, $\upsilon$, $\kappa$, $\varepsilon$, $\sigma$, $\zeta_1, \zeta_2$, $\rho$, $B_{m,n;[g]}^\mathsf{D2U}$ and $B_{m,n;[g]}^\mathsf{U2D}$;
		
		{\bf{Output :}} 
		$H^{*}$, $B_{m,n;[g]}^\mathsf{D2U,*}$, $B_{m,n;[g]}^\mathsf{U2D,*}$;
		
		{\bf{Initialization :}} 
		$H^{\langle 0 \rangle}$, $0<\zeta_1\leq\zeta_2\leq1$, $\sigma^{\langle 0 \rangle}>0$, $\kappa^{\langle 0 \rangle}=\frac{1}{\sigma^{\langle 0 \rangle}}$, $\varepsilon^{\langle 0 \rangle}=\frac{1}{(\sigma^{{\langle 0 \rangle}})^{-\zeta_1}}$, $\rho>1$;
		\LinesNumbered 
		
		\For{
			$j=0,1,\cdots$
		}{
            Get $\mathcal{Y}^{\langle j \rangle}$ based \textit{Theorem 2} and $H^{\langle j \rangle}$.
        
			Obtain the solution $H^{\langle j+1 \rangle}$ (based on \eqref{40}, \eqref{41}) that satisfies the the accuracy condition $\|\nabla_{H^{\langle j+1 \rangle}}L_{\sigma^j}(H^{\langle j+1 \rangle},\mathcal{Y}^{\langle j \rangle})\|_2\leq\kappa^{\langle j \rangle}$ in $\mathcal{P}_{1e}$.
            
			\If{$\psi^{\langle j \rangle}(H^{\langle j+1 \rangle})\leq\varepsilon^{\langle j \rangle}$}{
				\If{$\psi^j(H^{\langle j+1 \rangle})\leq\varepsilon^{\langle 0 \rangle}$}{Obtain approximate solution $\mathcal{Y}^{\langle j \rangle}$ and $H^* \leftarrow H^{\langle j+1 \rangle}$, break.}
				
				Multiplier $\upsilon^{\langle j+1 \rangle}$ Update based on \eqref{67}.
				
				Penalty factor no change: $\sigma^{\langle j+1 \rangle} \xleftarrow{} \sigma^{\langle j \rangle}$.
				
				Adjust $\kappa^{\langle j+1 \rangle}$ and $\varepsilon^{\langle j+1 \rangle}$ based on \eqref{69}--Case i. 
			}
			Update penalty factor $\sigma^{\langle j+1 \rangle}$ based on \eqref{71}, but $\upsilon^{\langle j+1 \rangle} \xleftarrow{} \upsilon^{\langle j \rangle}$.
			
			Modify $\kappa^{\langle j+1 \rangle}$ and $\varepsilon^{\langle j+1 \rangle}$ based on \eqref{69}--Case ii.
		}

	    Replace $H$ with $B_{m,n;[g]}^\mathsf{D2U}$ or $B_{m,n;[g]}^\mathsf{U2D}$ and find their optimal values using the above procedure.
	}				
\end{algorithm}

\section{}
\subsection{Challenges of $\mathcal{P}_2$}
\begin{remark}
First, the device-to-UAV association process is influenced by time-varying channel conditions, introducing uncertainty. These variations make it difficult to design a static optimization strategy, as the optimal selection of devices can shift dynamically due to changes in network conditions. Second, the convergence of the UAV's model in the \( K_{[g]} \)-th round of aggregation, characterized by \( \varpi_{m;[g]}^{1} \) and \( \varpi_{m;[g]}^{2} \), depends not only on the current selection of participating devices --- which is determined by the device relevance score \( \alpha_{m,n;[g]} \) and the adaptive threshold \( \beta_{m;[g]} \) --- but also on the cumulative impact of all previous intermediate aggregations. Further, since the aggregation process builds upon prior rounds, device selection during each aggregation affects future model performance, making \( \mathcal{P}_2 \) an interdependent optimization process rather than an isolated decision at each aggregation round.
\end{remark}

\subsection{Proof of Theorem 4}
\noindent\textit{Proof of Theorem 4:} The optimal policy of $\mathcal{P}_{2a}$ denoted by $\pi^*$, is the one that maximizes the cumulative reward:
\begin{align}
    \pi^*=\arg\max_\pi\vmathbb{E}\left[\sum_{g=1}^{G}\gamma^{g-1} \vmathbb{r}_{m;[g]}(\vmathbb{s},\vmathbb{a})|\pi\right]. \label{88}
\end{align}

In the set of policies that satisfy constraint $\widetilde{G}_{m;[g]}(\vmathbb{s})\leq0$, we have the optimal cumulative reward $\overline{\vmathbb{r}}^{*}$ for $\mathcal{P}_{2a}$, defined as
\begin{align}
    \overline{\vmathbb{r}}^{*}=\arg\max_\pi\vmathbb{E}\left[\sum_{g=1}^G\gamma^{g-1} \vmathbb{r}_{m;[g]}(\vmathbb{s},\vmathbb{a})|\widetilde{G}_{m;[g]}(\vmathbb{s})\leq0,\forall{g}\right]. 
\end{align}

Upon having penalty terms, the optimal strategy $\widetilde{\pi}$ maximizes cumulative reward for $\mathcal{P}_{2b}$:
\begin{align}
    \widetilde{\pi}=\arg\max_\pi\vmathbb{E}\left[\sum_{g=1}^G \gamma^{g-1}\Bigl\{ \vmathbb{r}_{m;[g]}(\vmathbb{s},\vmathbb{a}) - \widetilde{\alpha}(g) \widetilde{\mathcal{Y}}_{m;[g]} \Bigr\}|\pi\right]. 
\end{align}

Next, we show our analysis from three aspects: \textbf{\textit{(i) Behavior analysis of the penalty term:}} For any state that violates the constraint (i.e., $\widetilde{G}_{m;[g]}(\vmathbb{s})>0$), the penalty term is:
\begin{align}
    \widetilde{\alpha}(g) \cdot\max( \widetilde{G}_{m;[g]}(\vmathbb{s}), 0)^2=\widetilde{\alpha}(g)\cdot \widetilde{G}_{m;[g]}(\vmathbb{s})^2.
\end{align}

As $\widetilde{\alpha}(g) \to \infty $, the penalty term tends towards infinity. Thus, when choosing the optimal strategy, it will strongly suppress the strategies that cause $\widetilde{G}_{m;[g]}(\vmathbb{s})>0$. \textbf{\textit{(ii) Proving asymptotic equivalence:}} Assume that existence of a constraint violating policy $\pi$, where its cumulative reward is given by 
\begin{align}
    \vmathbb{E}\left[\sum_{g=1}^G \gamma^{g-1}\Bigl\{ \vmathbb{r}_{m;[g]}(\vmathbb{s}, \vmathbb{a}) - \widetilde{\alpha}(g) \widetilde{\mathcal{Y}}_{m;[g]} \Bigr\}|\pi\right]. 
\end{align}

We compare it with the optimal strategy $\pi^{*}$ that satisfies the constraints: for any state $\vmathbb{s}_{m;[g]}$ and policy $\pi$ that violates constraints, when $\widetilde{G}_{m;[g]}(\vmathbb{s})>0$, the penalty $\widetilde{\alpha}(g) \cdot \widetilde{G}_{m;[g]}(\vmathbb{s})^2>0$ will increase with $\widetilde{\alpha}(g) \to +\infty$, making the accumulated reward decrease. Therefore, when having $\widetilde{\alpha}(g) \to \infty$, the expected cumulative reward of any policy $\pi$ that violates the constraint will be significantly reduced, that is:
\begin{align}
        \vmathbb{E}\left[\sum_{g=1}^G \gamma^{g-1} \Bigl\{\vmathbb{r}_{m;[g]}(\vmathbb{s}, \vmathbb{a}) - \widetilde{\alpha}(g) \widetilde{\mathcal{Y}}_{m;[g]} \Bigr\}|\pi\right]\to-\infty. 
\end{align}

As a result, only those strategies that satisfy constraint $\widetilde{G}_{m;[g]}(\vmathbb{s})\leq0 $ can avoid this infinite penalty and ensure that the cumulative reward does not tend towards negative infinity. \textbf{\textit{(iii)~Whether the optimality of the strategy is consistent with the original function:}} According to \textit{(ii)}, it can be concluded that in the policy set where the constraint $\widetilde{G}_{m;[g]}(\vmathbb{s}) \leq 0$ is strictly satisfied, we have 
\begin{align}
    \max( \widetilde{G}_{m;[g]}(\vmathbb{s}), 0)^2=0.
\end{align}

This indicates that the cumulative reward, which includes a penalty term, is equivalent to the original cumulative reward function. This is because the penalty term vanishes for states $\vmathbb{s}_{m;[g]}$ that satisfy the constraint, leaving the reward function unchanged. Therefore, the optimality of strategies in this constrained policy set is consistent with the optimality derived from the original cumulative reward function in \eqref{88}.

Based on the above three aspects, we have proven that the cumulative reward with a penalty term is equivalent to the original cumulative reward. Thus, in the set of policies that satisfy constraints, the optimal policy $\widetilde{\pi}$ should be equivalent to $\pi^*$ because they both maximize the original cumulative reward.
\begin{algorithm}[]
    \label{Alg.3}
	{\footnotesize  
		\caption{\small{Multi-Criteria Device-to-UAV Associations with Adaptive Threshold (MCCUA-AT)}}
		{\bf{Input :}} 
		$\gamma$, $\tau$, $\mathbb{d}$, $\lambda_1, \lambda_2, \lambda_3$, $\nu_m^\mathsf{Per}$, $\mathcal{N}_{m;[g]}^\mathsf{Cov}$;
		
		{\bf{Output:}} 
		$\mathcal{N}_{m;[g]}^\mathsf{Sel}$;
		
		{\bf{Initialization :}} 
		$Q_{\boldsymbol{\vartheta}_1}$, $Q_{\boldsymbol{\vartheta}_2}$, $Q_{{\boldsymbol{\vartheta}_1^{\prime}}}, Q_{\boldsymbol{\vartheta}_2^{\prime}}$, $\mu_\Omega$, $\mu_{\Omega^{\prime}}$,  $\beta_{m,[g]}$, $\mathcal{N}_{m;[g]}^\mathsf{Sel}=\emptyset$;
		\LinesNumbered 
		
		\If{the global model parameter changes does not satisfy \eqref{4}}{
			
			\For{
				each UAV $m\in \mathcal{M}_{[g]}$
			}{
				
				\For{
					time step $t=\{1,2,\cdots, t^\mathsf{Step}\}$
				}{
					
					\textbf{\textit{TD3 process:}} // running at online UAV
					
					\textit{1.} Execute actions to update rewards and status: $\vmathbb{r}^{\langle t+1 \rangle}_{m;[g]}$, $\vmathbb{s}^{\langle t+1 \rangle}_{m;[g]}$ based on \eqref{46}, \eqref{47}.
					
					\textit{2.} Store experience: $(\vmathbb{s}^{\langle t \rangle}_{m;[g]}, \vmathbb{a}^{\langle t \rangle}_{m;[g]}, \vmathbb{r}^{\langle t+1 \rangle}_{m;[g]}, \vmathbb{s}^{\langle t+1 \rangle}_{m;[g]})$ into $\mathcal{B}$.
					
					\textit{3.} Select a small batch of $B$ experience transfer samples:
					\textit{(i)} Calculate $\vmathbb{a}_{m;[g]}^{i+1,\prime}$ based on \eqref{48} and \textit{(ii)} update $\mathcal{Z}_{m;[g]}^i$ based on \eqref{49}.
					
					\textit{4.} Use MBGD update value network parameter $\boldsymbol{\vartheta}$ based on \eqref{50}.
					
					\If{$t$ mod $\vmathbb{d}$}{
						Use MBGA update policy network parameter $\Omega$ based on \eqref{51}.
						
						Update penalty coefficient based on \eqref{52}.
						
						Soft update $\boldsymbol{\vartheta}^{\prime}$ and $\Omega^{\prime}$ based on \eqref{53}.
					}
				}   
				\textit{5.} Select the action with the largest reward in the time step as the output $\vmathbb{a}_{m;[g]}$.
				
				\textbf{\textit{Device-UAV Association:}} //running at online UAV
				
				\For{$n\in\mathcal{N}_{m;[g]}^\mathsf{Cov}$}{
					
					\textit{6.} UAV broadcasts $\nu_m^\mathsf{Per}$, which devices use to calculate $R_{m,n;[g]}$, $d_{m,n;[g]}$, $f_{n}$.
					
					\textit{7.} All devices transmit $R_{m,n;[g]}$, $d_{m,n;[g]}$, $f_{n}$ to the corresponding UAV, which will be used to calculate $S^\mathsf{Sim}_{m,n;[g]}$, $S^\mathsf{Dis}_{m,n;[g]}$, $S^\mathsf{Fre}_{m,n;[g]}$.
					
					\textit{8.} Calculate $\alpha_{_{m,n;[g]}}$ according to \eqref{5}.
					
					\textit{9.} Get $\beta_{m;[g]}$: $\beta_{m;[g]} \leftarrow\vmathbb{a}_{m;[g]}$.
					
					\If{$\alpha_{m,n;[g]} \geq \beta_{m;[g]}$}{Choose device $n$ and set $\mathcal{N}_{m;[g]}^\mathsf{Sel} \xleftarrow{} \{n\}.$}
				}
			}
			\If{device $n$ in multiple UAVs coverage areas}{Device $n$ will be associated with a UAV with the highest $\alpha_{m,n;[g]}$.}
		}

	}				
\end{algorithm} 

\subsection{Key Improvements Brought by Our Designed TD3 over DDPG}
Recall our previous discussions, TD3 effectively
addresses several challenges through three key improvements
over DDPG, as detailed below.
\textit{(i)} Clipped Double Q-Learning: TD3 employs two Q-value networks and updates only the minimum Q-value estimate to avoid overestimation bias, ensuring that the learned policy does not exploit inaccurate Q-value estimations. \textit{(ii)} Delayed Policy Updates: The policy network (actor) is updated less frequently than the Q-value networks, preventing the policy from being trained on rapidly fluctuating or inaccurate value estimates.  \textit{(iii)} Target Policy Smoothing: To improve exploration robustness, TD3 adds noise to the target action before computing the target Q-value, making the policy more resistant to local perturbations and leading to better generalization in dynamic environments.

\subsection{Device-UAV Association Algorithm}
We have demonstrated the detailed process in Alg. 3.

\section{Algorithm and Analysis Related to $\mathcal{P}_3$}
\begin{algorithm}[]
    \label{Alg.4}
	{\footnotesize  
		\caption{\small{Two-Stage Greedy for UAV Redeployment and Central Aggregator Selection (TSG-URCAS)}}
		{\bf{Input :}} 
		$\mathtt{p}_{m;[g]}$, $\mathcal{M}_{[g]}$, $V_m$, $\overline{p_m}^\mathsf{Move}$, $\chi_1$, $\chi_2$;
		
		{\bf{Output :}} 
		$X_{m;[g]}=1$, $\mathtt{p}_{m;[g]}$;
		
		{\bf{Initialization :}} 
		$\mathcal{V}_{m;[g]}^{\langle 0,~0 \rangle}$, $\overline{\mathcal{V}}_{m;[g]}^{\langle 0 \rangle}$, $\overline{\mathcal{V}}_{m;[g]}^*=+\infty$, $\hat{b}=0$, $d_{m}^\mathsf{Set}=0$;
		\LinesNumbered 

        \For{each UAV server $m\in\mathcal{M}_{[g]}$}{
            
            \textbf{Stage 1:} // Rough Search

            \If{$q \leq \chi_1$}{
            
                \For{$\hat{a}=\{1,2,\cdots,10\}$}{
                    Calculate UAV coverage $|\mathcal{N}_{m;[g]}^\mathsf{Cov}|^{\langle \hat{a},~\hat{b} \rangle}/|\mathcal{N}_{m;[g]}^\mathsf{Cov}|^{\langle \hat{b}-1 \rangle}$.

                    Obtain $\mathcal{V}_{m;[g]}^{\langle \hat{a},~\hat{b} \rangle}$ based on \eqref{82}.

                    \If{$\mathcal{V}_{m;[g]}^{\langle \hat{a},~\hat{b} \rangle} > \mathcal{V}_{m;[g]}^{\langle \hat{b} \rangle}$}{
                    
                        $\mathcal{V}_{m;[g]}^{\langle \hat{b} \rangle} \leftarrow \mathcal{V}_{m;[g]}^{\langle \hat{a},~\hat{b} \rangle} $
                    }

                \If{$\mathcal{V}_{m;[g]}^{\langle \hat{b} \rangle} < \tilde{\xi}_1$}{

                        q += 1
                }

                Update the UAV $m$'s position.
            }

            $\hat{b}$ += 1

        }
        $\hat{b}$ = 0.

        \textbf{Stage 1:} // Precise Search

        Same as the rough search steps, but explore directions from 10 to 15, and $\chi_1 \rightarrow \chi_2$.

    }

    \textbf{Stage 2:} // global aggregator UAV choose

    \For{$m \in \mathcal{M}_{[g]}$}{
        Calculate distance between UAV $m$ and remaining online UAVs $d_{m^{\prime} \to m;[g]}^\mathsf{UAV}$ and sum of all online UAV-to-UAV distances and obtain $\overline{\mathcal{V}}_{m;[g]}$ based on \eqref{83}.

        \If{$\overline{\mathcal{V}}_{m;[g]}\leq \overline{\mathcal{V}}_{m;[g]}^{*}$}{
                
            $\overline{\mathcal{V}}_{m;[g]}^{*} \xleftarrow[]{} \overline{\mathcal{V}}_{m;[g]}$.
        }
    }        
    Select the UAV corresponding to $\overline{\mathcal{V}}_{m;[g]}^{*}$, and set $X_{m;[g]}=1$.

}
\end{algorithm}

\noindent \textbf{(I) First Stage:} We determine the UAV's moving direction through \textit{rough search} and \textit{precise search}. In the rough search stage, the UAV moves a fixed distance in 10 different directions and evaluates the change in device coverage after each move, using it as a reward metric. The UAV selects the direction that maximizes device coverage. However, since excessive movement may lead to high energy consumption, we define a benefit function to balance coverage expansion and movement cost. Specifically, the benefit of moving to the $\hat{a}$-th direction for the $\hat{b}$-th attempt during the rough search in the $g^\text{th}$ global round is defined as follows:
\setcounter{equation}{83}
\begin{align}
    \mathcal{V}_{m;[g]}^{\langle \hat{a},~\hat{b}\rangle} = \lambda_{8}\left\{\frac{|\mathcal{N}_{m;[g]}^\mathsf{Cov}|^{\langle \hat{a},~\hat{b} \rangle}}{|\mathcal{N}_{m;[g]}^\mathsf{Cov}|^{\langle \hat{b}-1 \rangle}}-1\right\}-\lambda_{9}\left\{\frac{\hat{b}d_{m}^\mathsf{Set}}{V_m}\overline{p_m}^\mathsf{Move}\right\}, \label{82} 
\end{align}
where $\frac{|\mathcal{N}_{m;[g]}^\mathsf{Cov}|^{\langle \hat{a},~\hat{b} \rangle}}{|\mathcal{N}_{m;[g-1]}^\mathsf{Cov}|^{\langle \hat{b}-1 \rangle}}-1$ represents the relative increase in device coverage, $\frac{\hat{b}d_{m}^\mathsf{Set}}{V_m}\overline{p_m}^\mathsf{Move}$ represents the cumulative energy consumption due to UAV movement and $d_{m}^\mathsf{Set}$ represents the distance moved of a single rough search, $\lambda_{8}, \lambda_{9} \in [0,1]$ represent the weight coefficients of the expansion of coverage and the impact of mobile costs on the benefit function. The UAV moves in the direction with the highest benefit value $\mathcal{V}_{m;[g]}^{\langle \hat{b} \rangle}$. If the maximum benefit does not exceed a predefined threshold \( \tilde{\xi}_1 \) for \( \chi_1 \) consecutive iterations, the UAV enters the precise search stage (at this time, $\hat{b}$ will be reset to 0 for the next calculation). In this stage, the UAV explores 15 finer movement directions within a smaller radius and re-evaluates the comprehensive benefit. If no movement direction achieves a benefit higher than \( \tilde{\xi}_2 \) for \( \chi_2 \) consecutive iterations, the UAV is considered to have reached its optimal position.

\noindent \textbf{(II) Second Stage:} Once UAV repositioning is finalized, we determine the global aggregation UAV based on a communication cost function, which accounts for the distance between the UAV selected as the central aggregator and the other UAVs. In particular, the benefit function for selecting UAV \( m \) as the central aggregator for the \( (g+1)^\text{th} \) global iteration is defined as follows:
\begin{align}
    \overline{\mathcal{V}}_{m;[g]} =\sum_{\forall m'\in\mathcal{M},m'\neq m}d_{m' \rightarrow m;[g]}^\mathsf{UAV}, \label{83}
\end{align}
where $d_{m' \rightarrow m;[g]}^\mathsf{UAV}$ represents the distance between UAV $m'$ to $m$. We naturally select the UAV with the smallest $\overline{\mathcal{V}}_{m;[g]}$ (i.e., $\overline{\mathcal{V}}_{m;[g]}^*$, that is, the sum of the distances from the selected central aggregator to the remaining UAVs) as global aggregator (i.e. $X_{m;[g]}=1$). The detailed UAV repositioning and aggregator selection process is outlined in Alg. 4, with time and space complexities denoted by $\mathcal{O}(\mathcal{M}^2_{[g]})$ and $\mathcal{O}(\mathcal{M}_{[g]})$, respectively.

\section{}
\subsection{Key Parameter Settings in Experiments}
\setlength{\tabcolsep}{0.5mm}{
	\begin{table}[h!t]
		\centering
		{\footnotesize
			\vspace{-0.2 cm}
			{\setlength{\extrarowheight}{2.3pt}
				{
					\footnotesize
					\caption{Key Settings}
					\renewcommand\arraystretch{1}
					\begin{tabular}{|ll|}
						\hline
                        \rowcolor{verylightgray}
						\multicolumn{1}{|c|}{{\textbf{Parameter}}} & \multicolumn{1}{c|}{{\textbf{Value}}} \\ 
						\hline
                        \hline
						\multicolumn{1}{|l|}{$\mathcal{H}$} & UAV height: 150 m \\
						\hline
                        \rowcolor{veryverylightblue}
						\multicolumn{1}{|l|}{$\vartheta_{n}$} & Chipset capacitance coefficient: $10^{-28}$\\
						\hline
						\multicolumn{1}{|l|}{$c_n$} & Number of CPU cycles: [30,100] cycle/bit \\
						\hline
                        \rowcolor{veryverylightblue}
						\multicolumn{1}{|l|}{$f_n$} & IoT devices' CPU frequencies: [1,10] GHz \\
						\hline
						\multicolumn{1}{|l|}{$N_0$} & Power spectral density of AWGN: -174 dBm/Hz \\
						\hline
                        \rowcolor{veryverylightblue}
						\multicolumn{1}{|l|}{$\overline{p_m}, \overline{p_m}^\mathsf{Move}$} & {UAV's hovering powers and moving powers: 100 W, 160 W} \\
						\hline
						\multicolumn{1}{|l|}{$p_{n}^\mathsf{D2U}$} & IoT devices' transmit powers: [200,800] mW \\
						\hline
                        \rowcolor{veryverylightblue}
						\multicolumn{1}{|l|}{$p_{m}^\mathsf{U2U}$} & UAVs' transmit powers: [500,1000] mW\\
						\hline
						\multicolumn{1}{|l|}{$p_{m}^\mathsf{U2D}$} & UAVs' broadcast powers: [300,1200] mW \\
						\hline
                        \rowcolor{veryverylightblue}
						\multicolumn{1}{|l|}{$B_m^\mathsf{D2U},B_m^\mathsf{U2D}$} & Total bandwidth resource of UAV $m$: [20,100] MHz \\
						\hline 
                            \multicolumn{1}{|l|}{$B^{\mathsf{U2U}}_{m,m';[g]}$} & Communication bandwidth between UAV $m$ and $m'$: 2 MHz \\
						\hline
                        \rowcolor{veryverylightblue}
						\multicolumn{1}{|l|}{$K^\mathsf{Max}$} & Maximum number of edge iterations: 10 \\
						\hline
						\multicolumn{1}{|l|}{$\eta,\hat{\eta}$} & Learing rate: 0.001, 0.002 \\
						\hline
                         \rowcolor{veryverylightblue}
						\multicolumn{1}{|l|}{$\chi_1$, $\chi_2$} & Maximum consecutive rough and precise searches allowed: 8, 6. \\
						\hline
					\end{tabular}
					\label{table3}
		}}}
		\vspace{-0.3 cm}
\end{table}}

\subsection{Simulation Results on UAV Dataset}
Fig. 9 illustrates representative categories from the ALI dataset, which closely emulate realistic UAV deployment scenarios. We assess algorithmic performance across three models. It is worth noting that, as ALI comprises color images whereas MNIST and FaMNIST are grayscale, the model architectures are accordingly adjusted to account for the different input modalities.

Fig. 10 presents the convergence rate, time cost, and energy cost comparisons among three models on ALI dataset. As shown in Figs. 10(a)-(c), CEHFed achieves a faster convergence rate than AHFed, HFedAT and CFed, while maintaining comparable performance with other algorithms. In terms of time cost (Figs. 10(d)-(f)), comparing to GDHFed, HFed, RHFed, GSHFed, AHFed, HFedAT and CFed of CNN on ALI, our method reduces execution time by 25.79\%, 16.26\%, 33.9\%, 19.3\%, 40.78\%, 49.73\% and 27.23\%. Similarly, in energy cost, it achieves a reduction of 36.11\%, 15.42\%, 28.54\%, 23.46\%, 51.94\%, 56.72\% and 46.11\% (Figs. 10(g)-(i)). Similar trends are observed for the other two models. Overall, the proposed algorithm exhibits strong robustness in practical UAV scenarios, achieving substantial reductions in time and energy consumption during global iterations without sacrificing convergence performance.

\begin{figure}[htbp]
	\centering
	\subfigure[Agriculture]{
		\includegraphics[trim=0cm 0cm 1cm 0cm, clip, width=0.295\columnwidth]{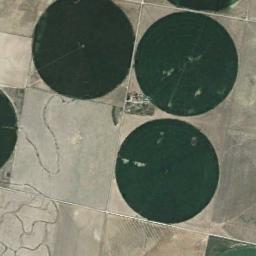}
	}
	\subfigure[Airport]{
		\includegraphics[trim=0cm 0cm 1cm 0cm, clip, width=0.295\columnwidth]{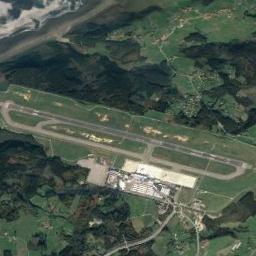}
	}
	\subfigure[Beach]{
		\includegraphics[trim=0cm 0cm 1cm 0cm, clip, width=0.295\columnwidth]{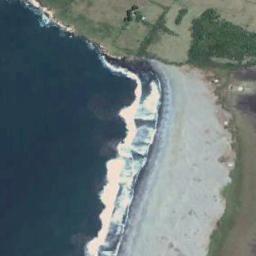}
	}

        \subfigure[City]{
		\includegraphics[trim=0cm 0cm 1cm 0cm, clip, width=0.295\columnwidth]{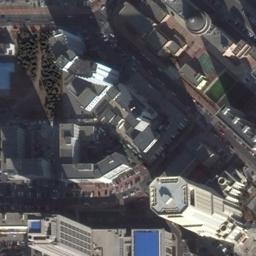}
	}
	\subfigure[Desert]{
		\includegraphics[trim=0cm 0cm 1cm 0cm, clip, width=0.295\columnwidth]{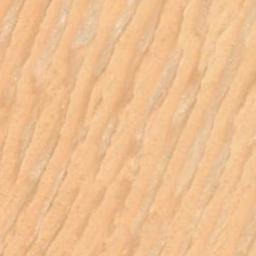}
	}
	\subfigure[Forest]{
		\includegraphics[trim=0cm 0cm 1cm 0cm, clip, width=0.295\columnwidth]{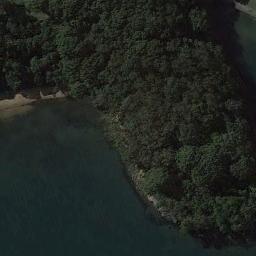}
	}
        \vspace{-3mm}
	
	\caption{ Different categories of images in ALI.}
   
	\label{fig_9}
\end{figure}

\begin{figure}[htbp]
	\centering
	\subfigure[CNN on ALI]{
		\includegraphics[trim=0.3cm 0.1cm 1.2cm 0cm, clip, width=0.295\columnwidth]{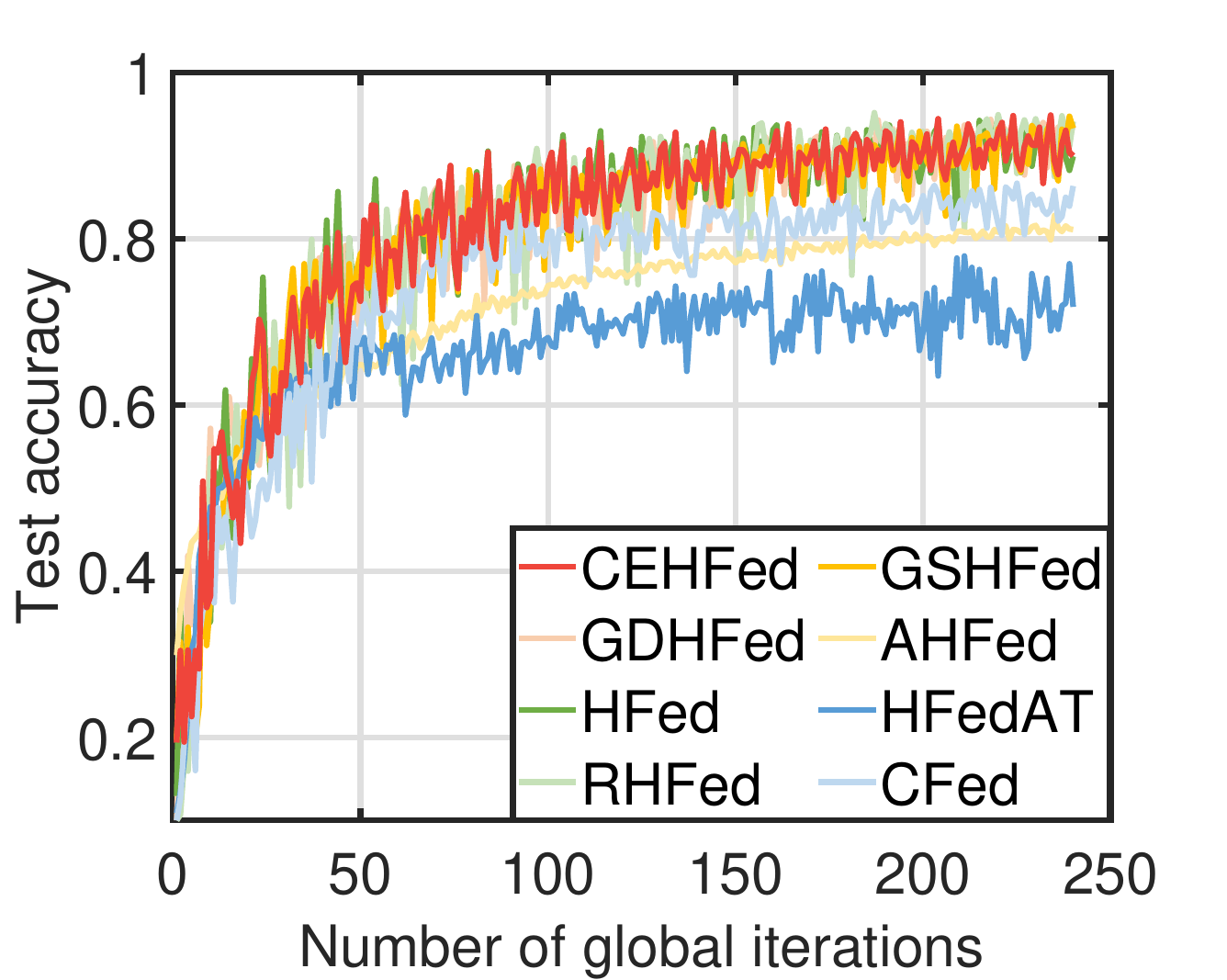}
	}
	\subfigure[LeNet5 on ALI]{
		\includegraphics[trim=0.3cm 0.1cm 1.2cm 0cm, clip, width=0.295\columnwidth]{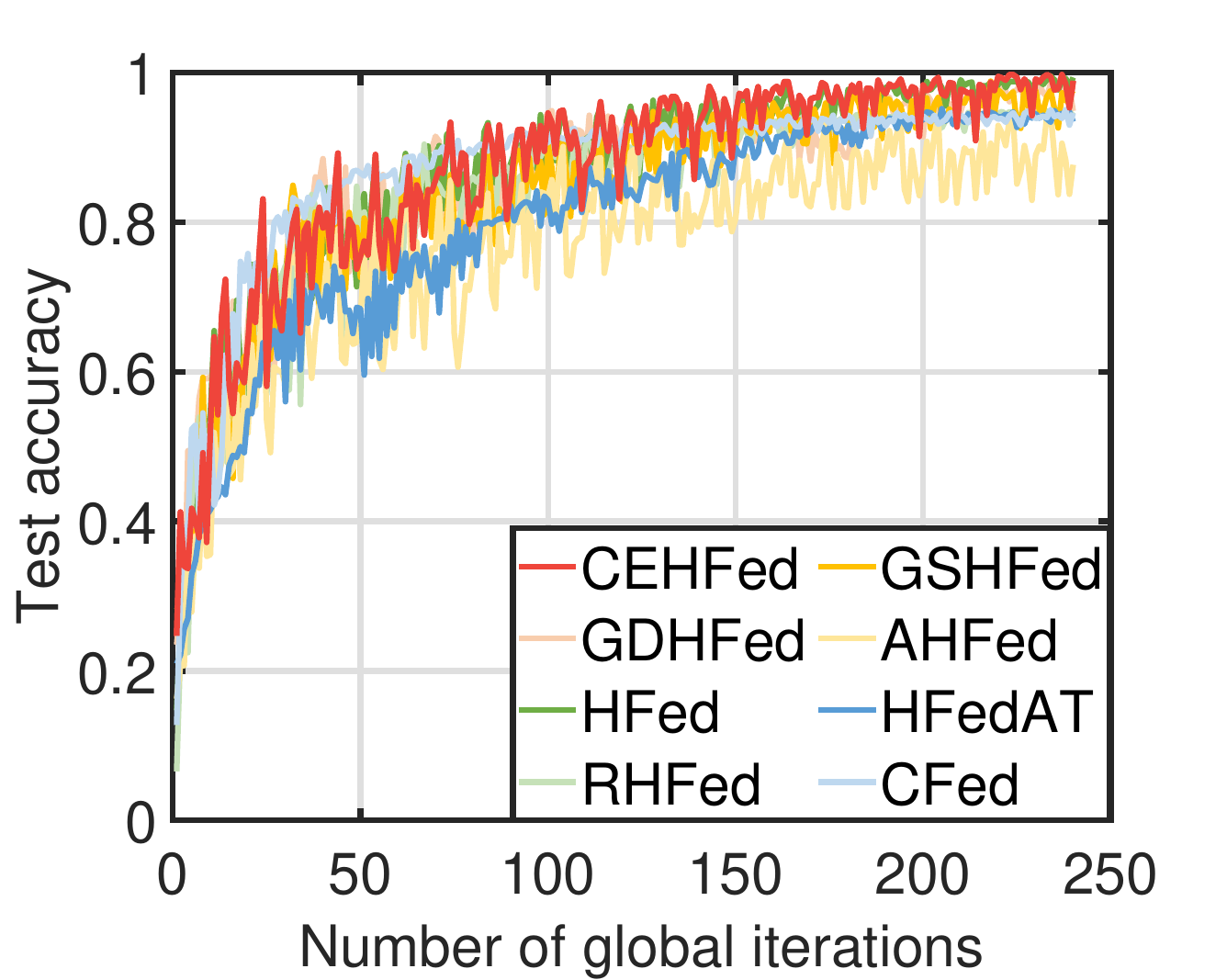}
	}
	\subfigure[VGG on ALI]{
		\includegraphics[trim=0.3cm 0.1cm 1.2cm 0cm, clip, width=0.295\columnwidth]{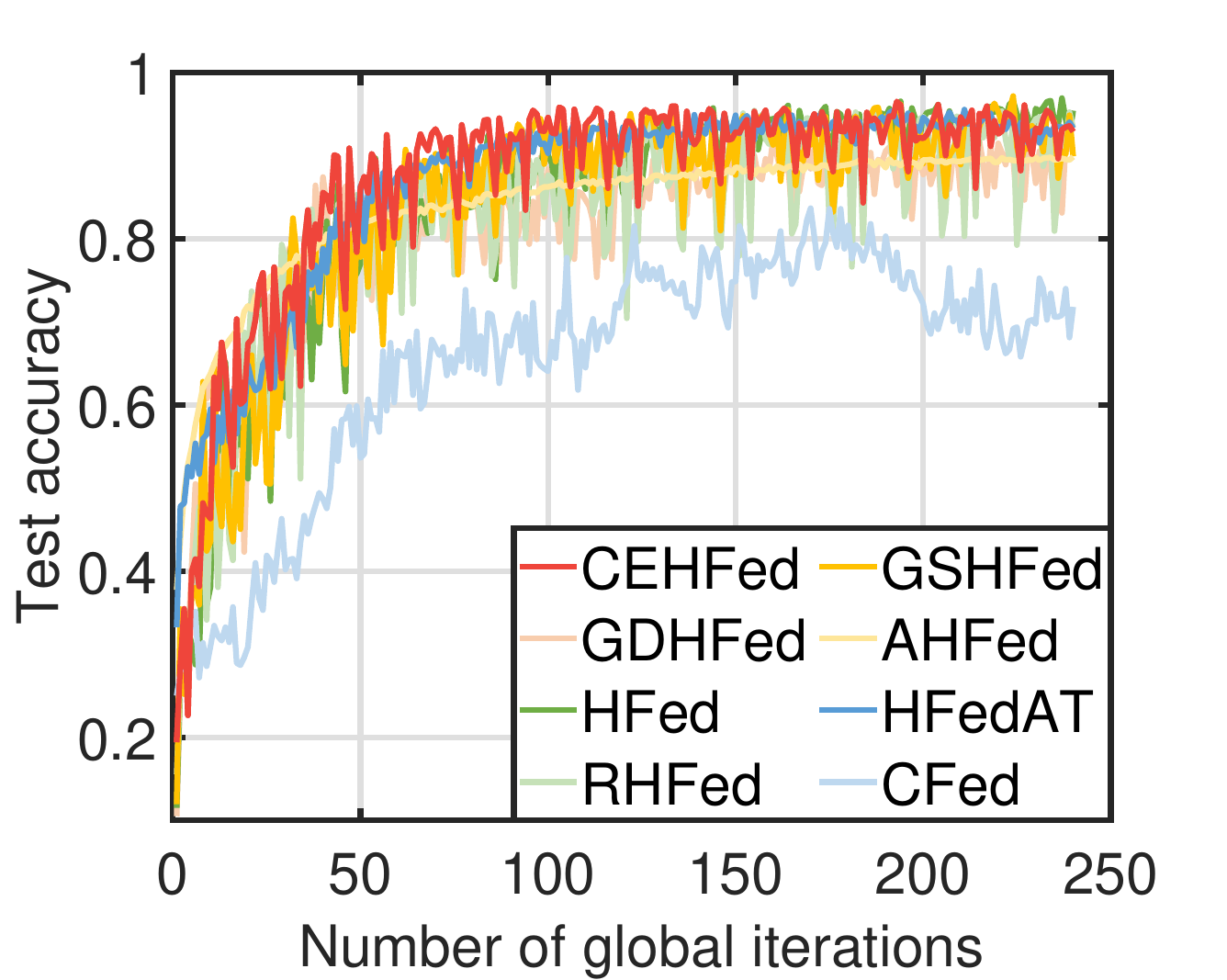}
	}

        \subfigure[CNN on ALI]{
		\includegraphics[trim=0.3cm 0.1cm 1.2cm 0cm, clip, width=0.295\columnwidth]{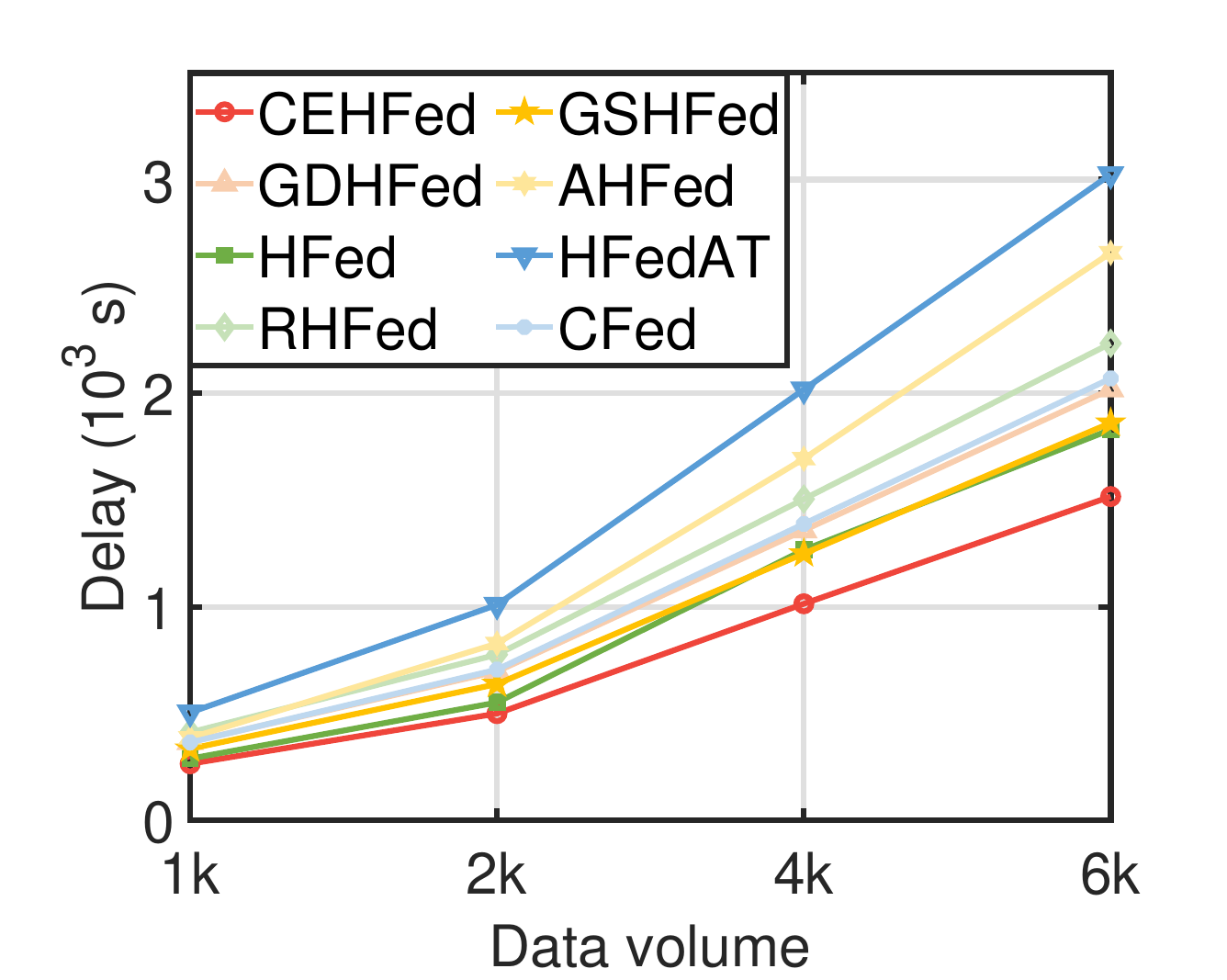}
	}
	\subfigure[LeNet5 on ALI]{
		\includegraphics[trim=0.3cm 0.1cm 1.2cm 0cm, clip, width=0.295\columnwidth]{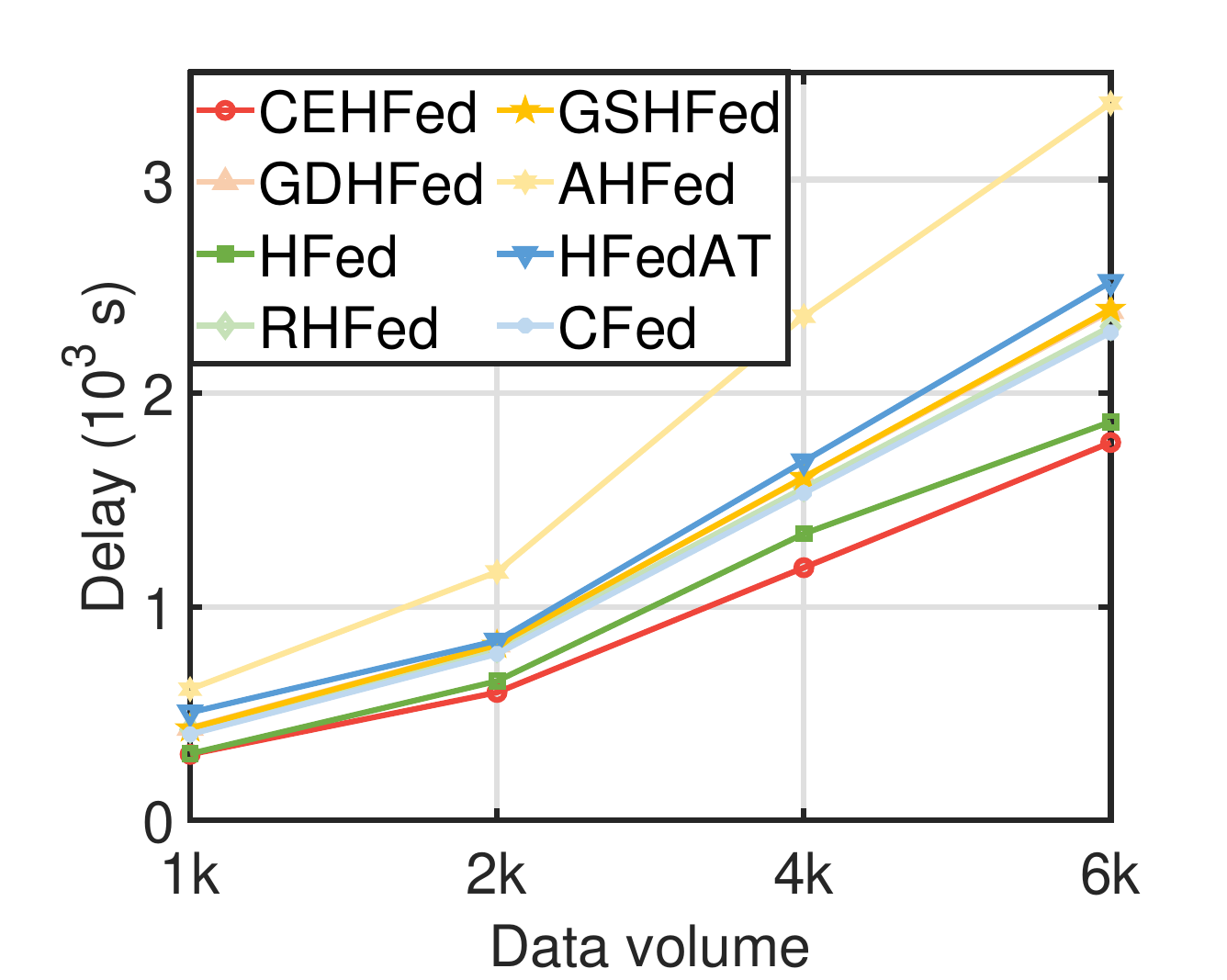}
	}
	\subfigure[VGG on ALI]{
		\includegraphics[trim=0.3cm 0.1cm 1.2cm 0cm, clip, width=0.295\columnwidth]{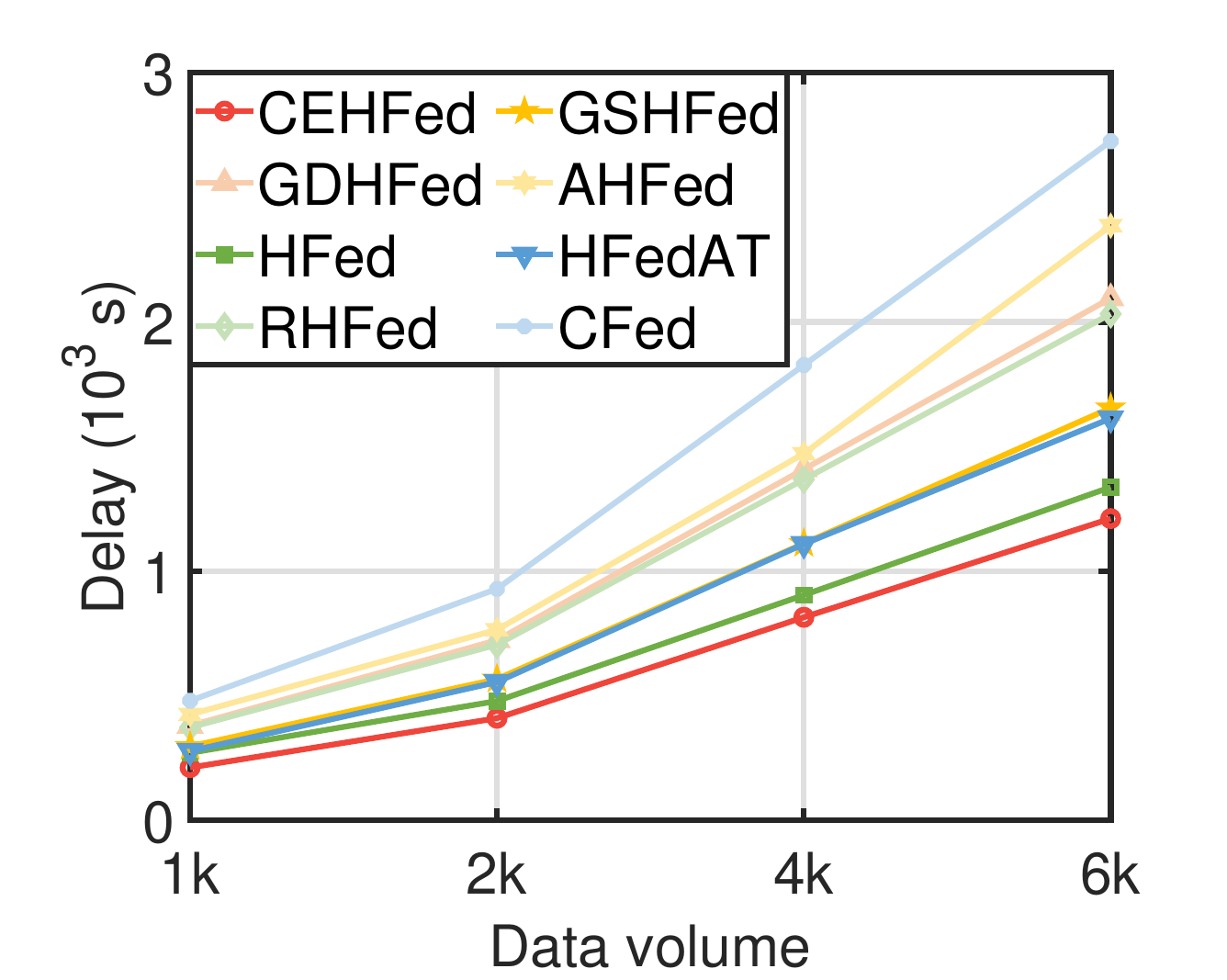}
	}

        \subfigure[CNN on ALI]{
		\includegraphics[trim=0.3cm 0.1cm 1.2cm 0cm, clip, width=0.295\columnwidth]{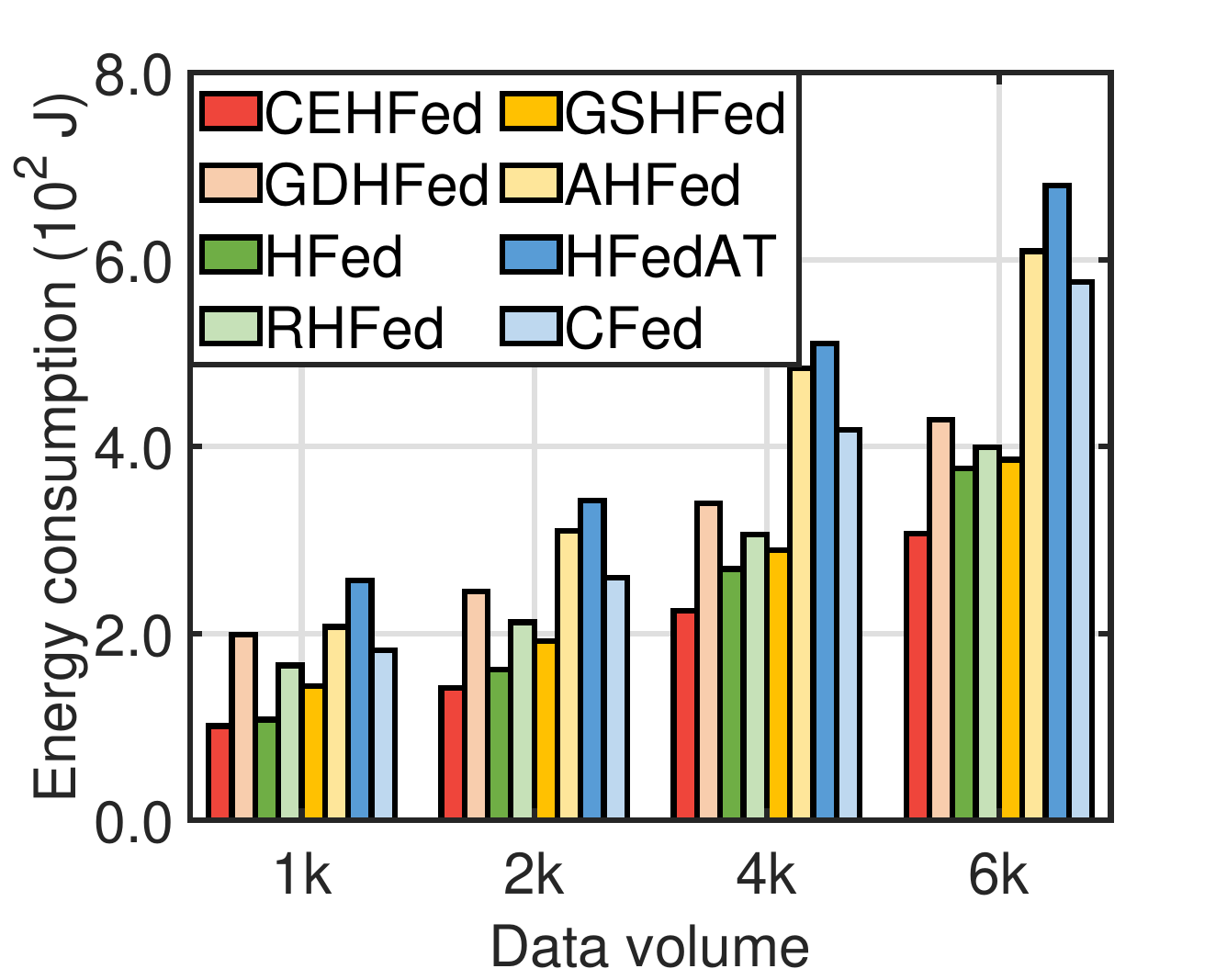}
	}
	\subfigure[LeNet5 on ALI]{
		\includegraphics[trim=0.3cm 0.1cm 1.2cm 0cm, clip, width=0.295\columnwidth]{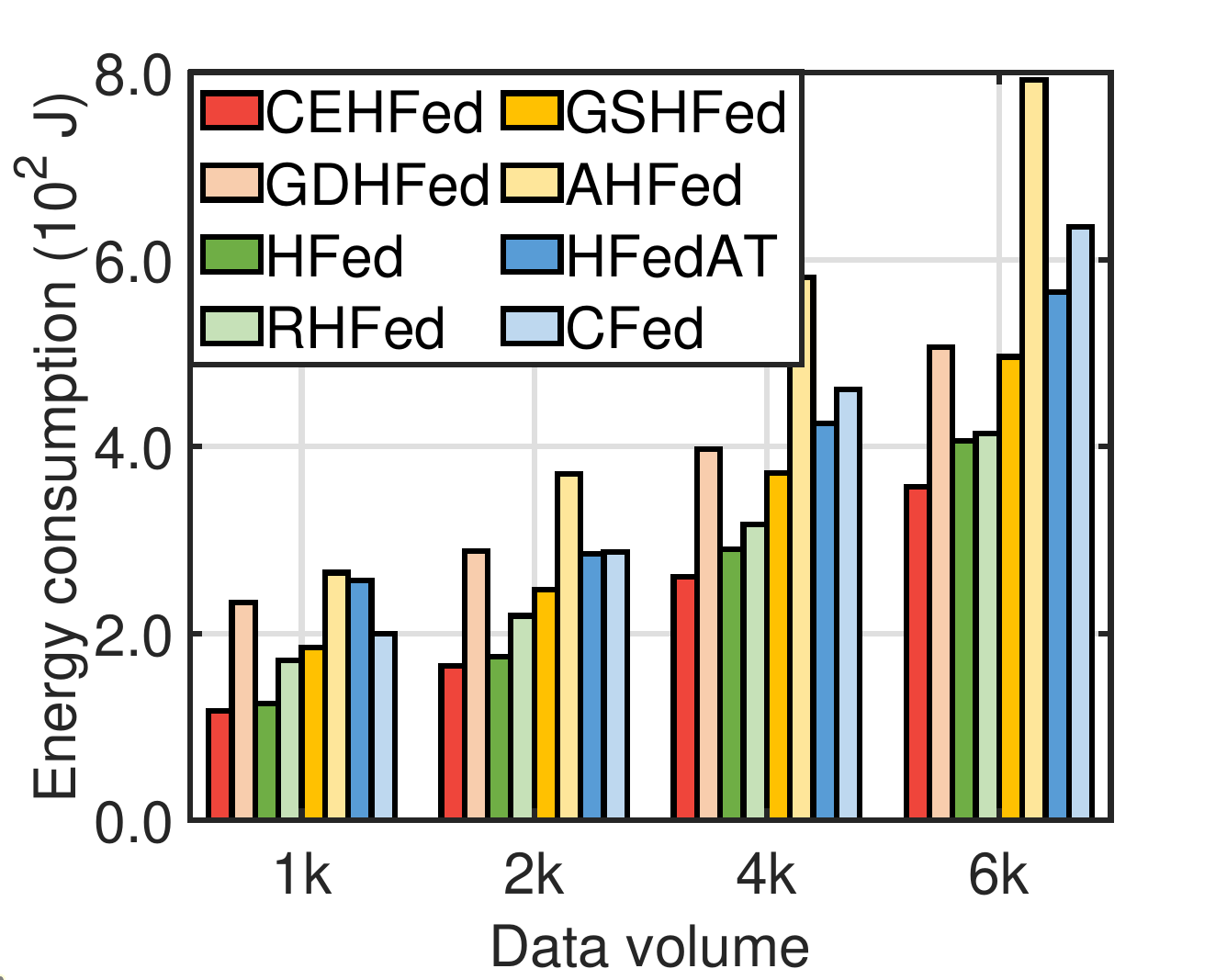}
	}
	\subfigure[VGG on ALI]{
		\includegraphics[trim=0.3cm 0.1cm 1.2cm 0cm, clip, width=0.295\columnwidth]{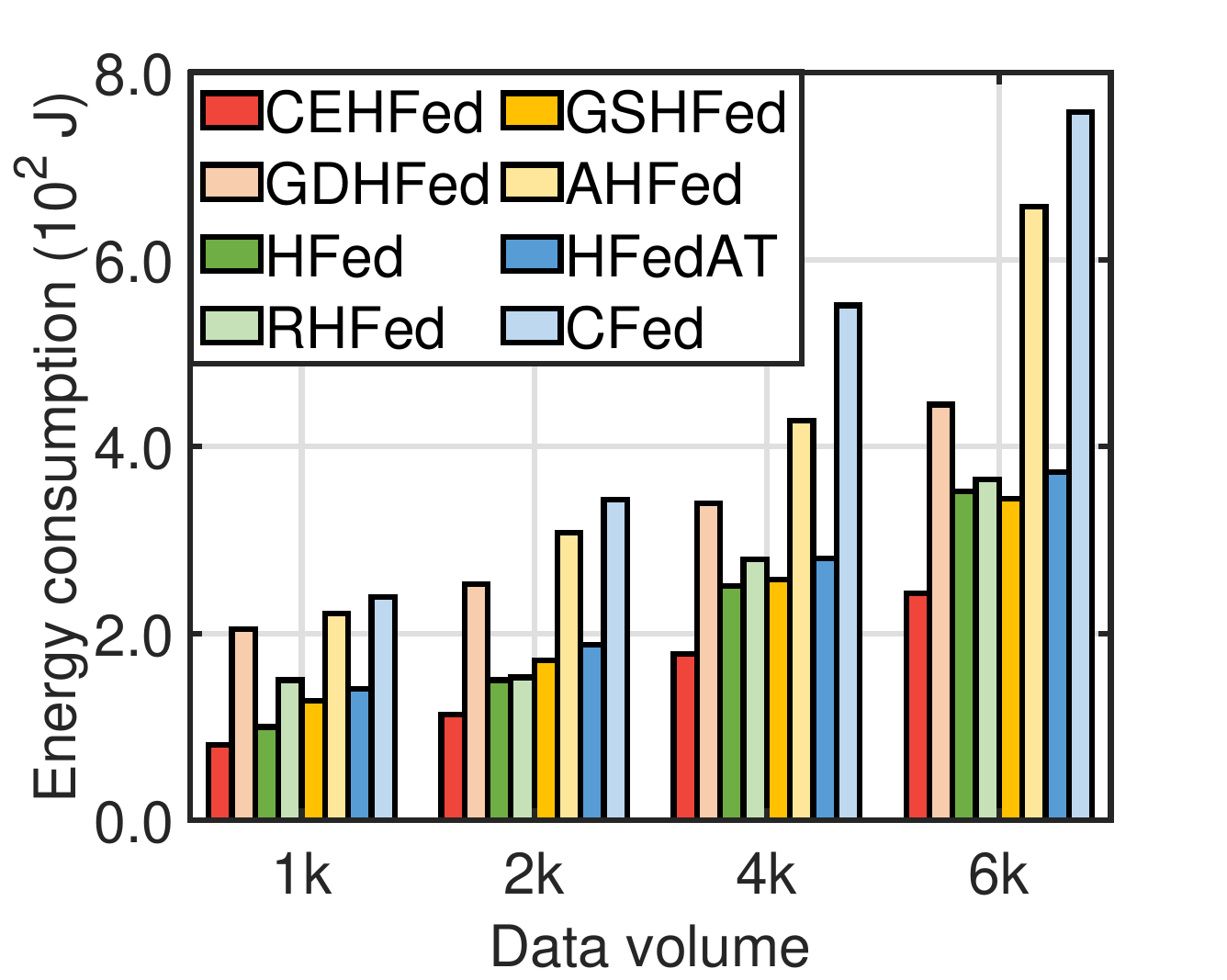}
	}
	\vspace{-3mm}
	\caption{(a), (b), (c): Performance comparisons in terms of test accuracy on ALI datasets using different models. (d), (e), (f): Time cost of model training on ALI datasets upon having different models. (g), (h), (i): Energy consumption of the model training operations on ALI datasets upon having different models.}
   \vspace{-3mm}
	\label{fig_10}
   \vspace{-3mm}
\end{figure}

\begin{figure}[htbp]  
	\centering
	\subfigure[Init. 1 UAV Drop]{
		\includegraphics[trim=0.2cm 0cm 0.2cm 0.2cm, clip, width=0.295\columnwidth]{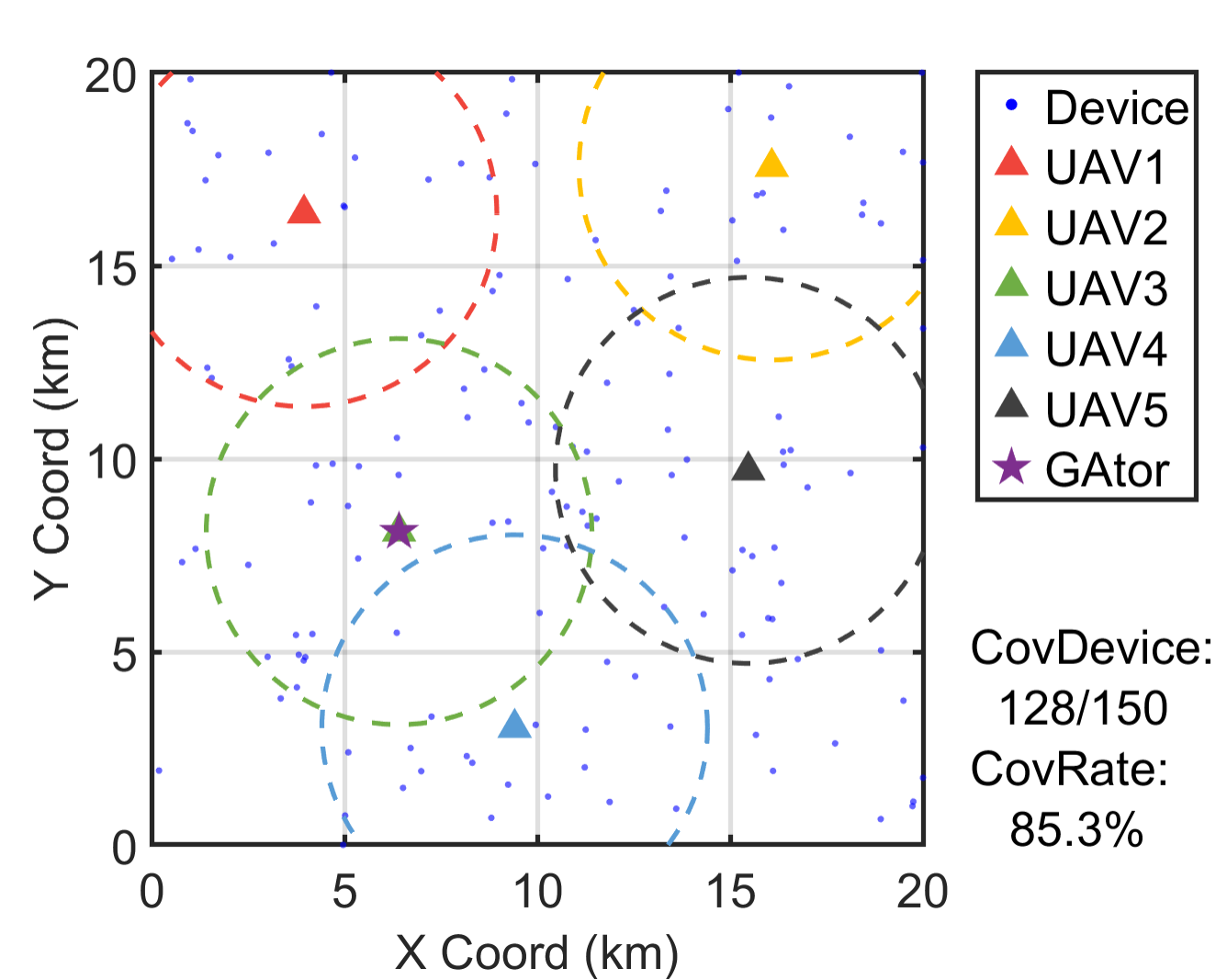}
	}
	\subfigure[Mid. 1 UAV Drop]{
		\includegraphics[trim=0.2cm 0cm 0.2cm 0.2cm, clip, width=0.295\columnwidth]{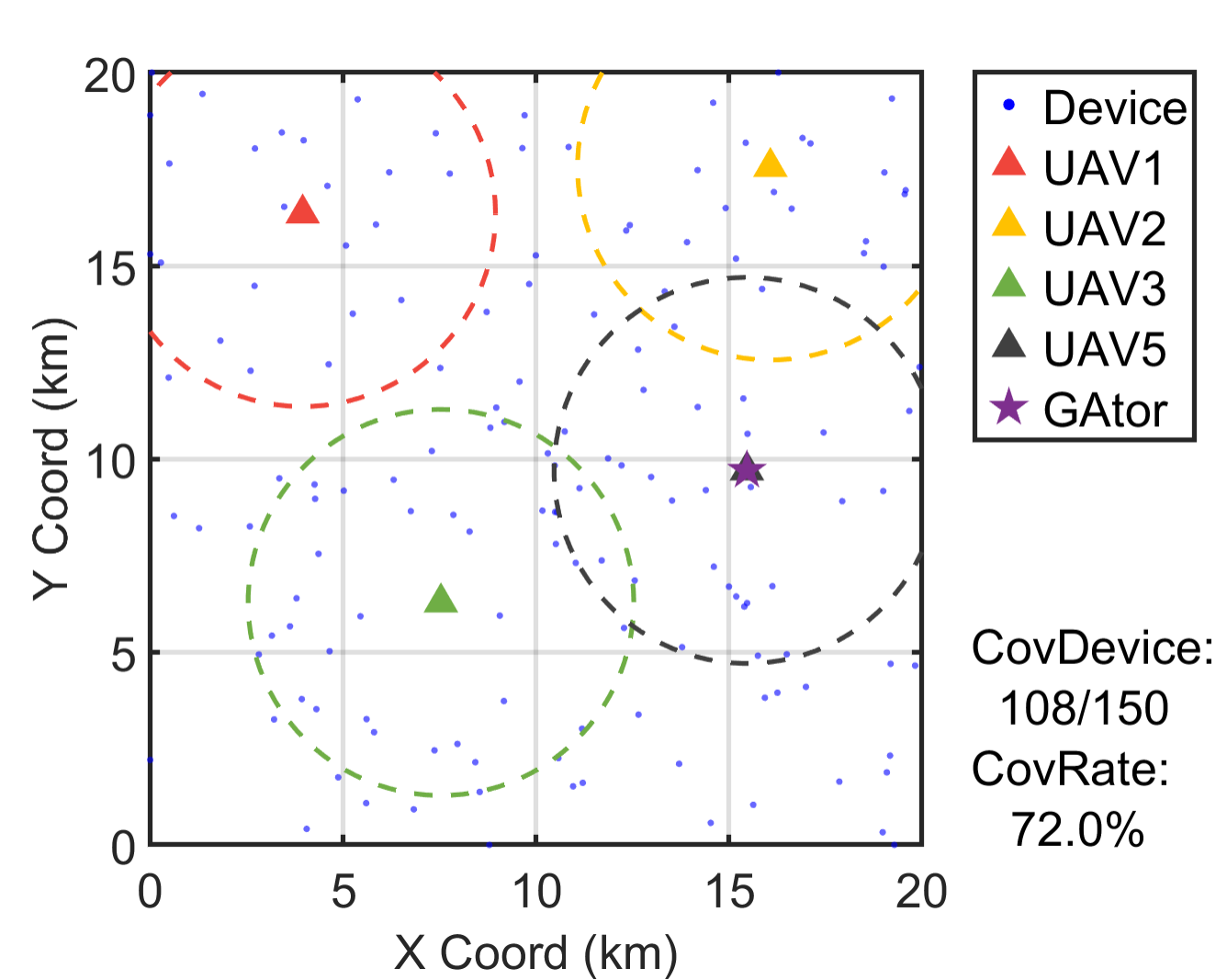}
	}
	\subfigure[Fin. 1 UAV Drop]{
		\includegraphics[trim=0.2cm 0cm 0.2cm 0.2cm, clip, width=0.295\columnwidth]{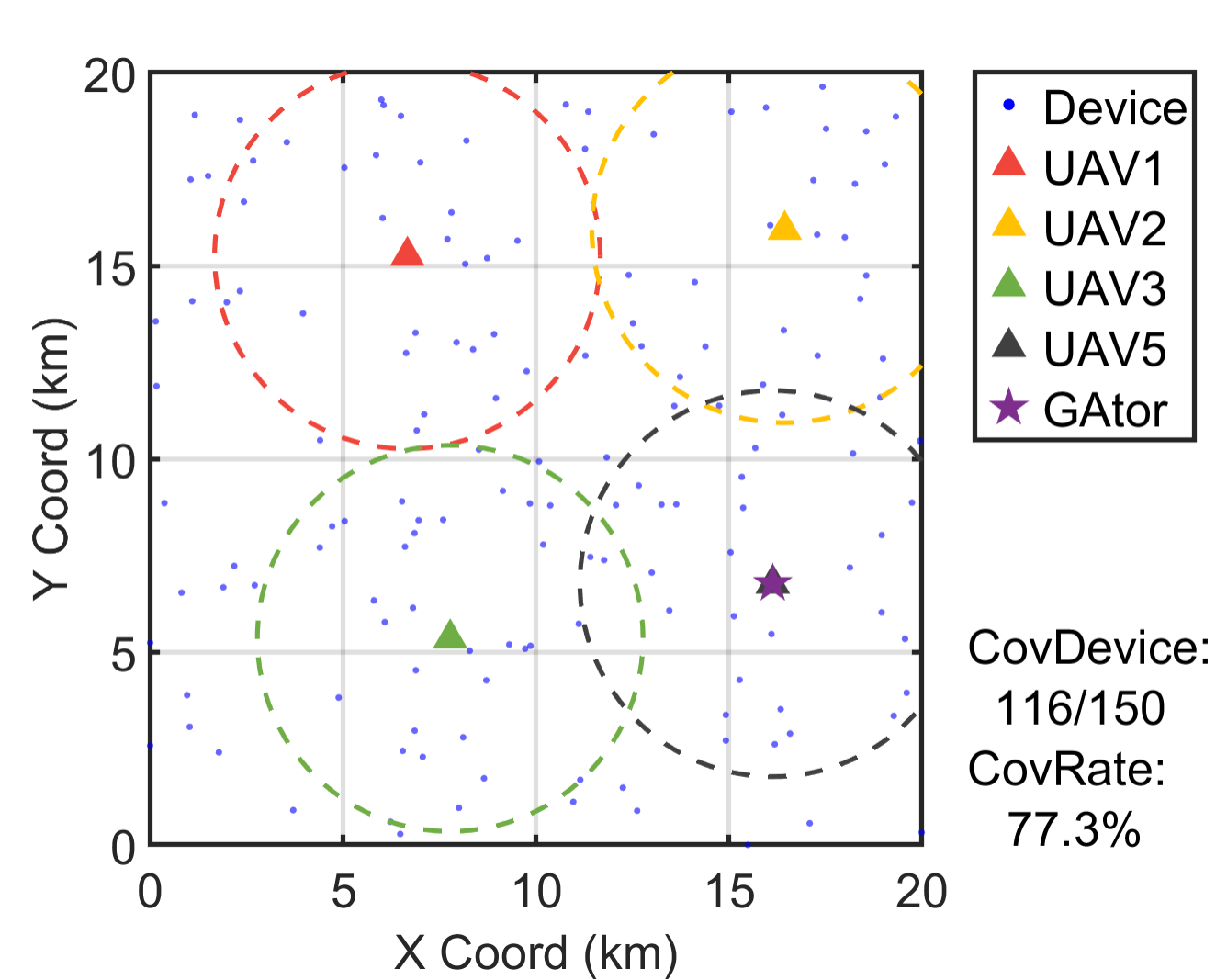}
	}

    \subfigure[Init. 2 UAVs drop]{
		\includegraphics[trim=0.2cm 0cm 0.2cm 0.2cm, clip, width=0.295\columnwidth]{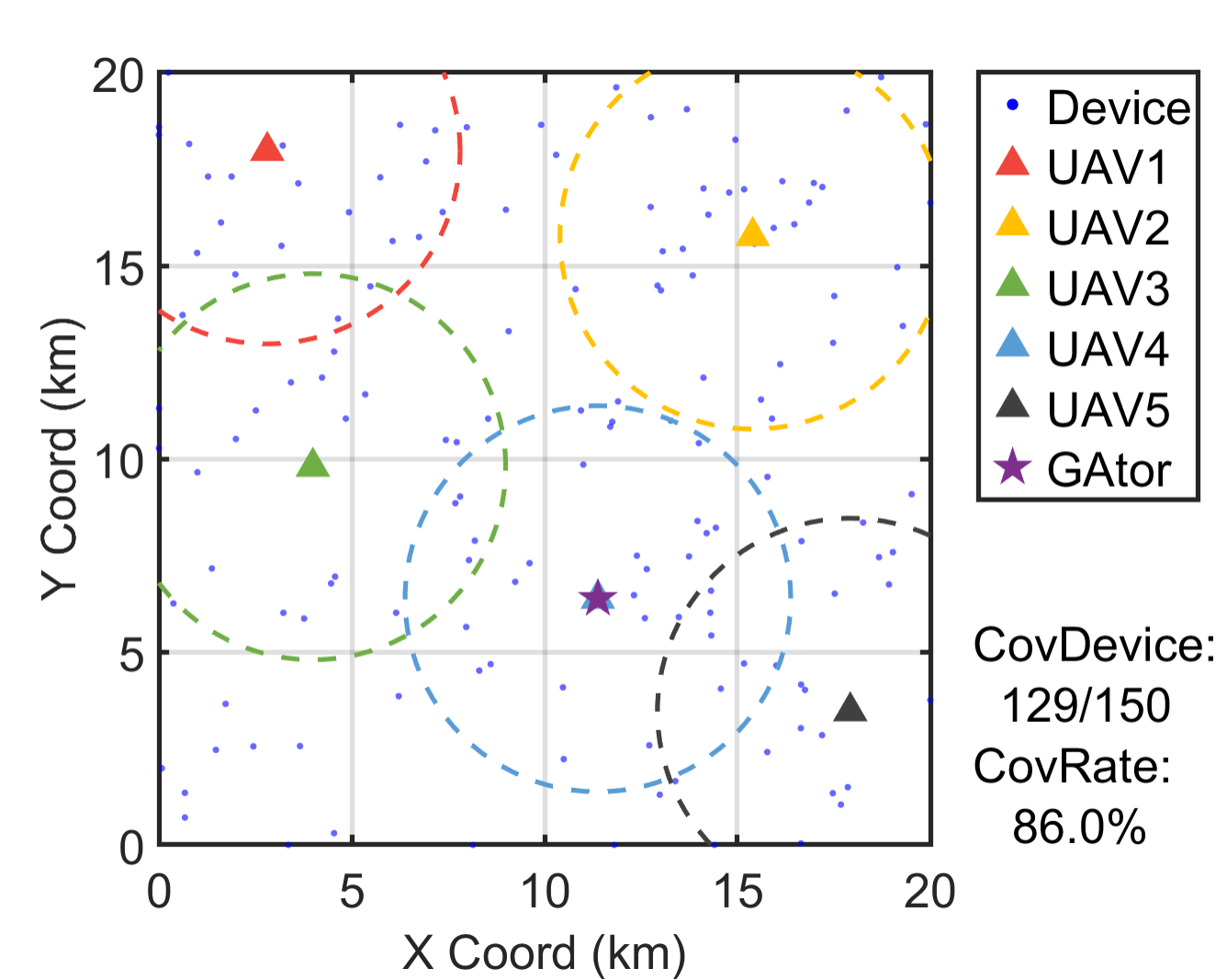}
	}
	\subfigure[Mid. 2 UAVs Drop]{
		\includegraphics[trim=0.2cm 0cm 0.2cm 0.2cm, clip, width=0.295\columnwidth]{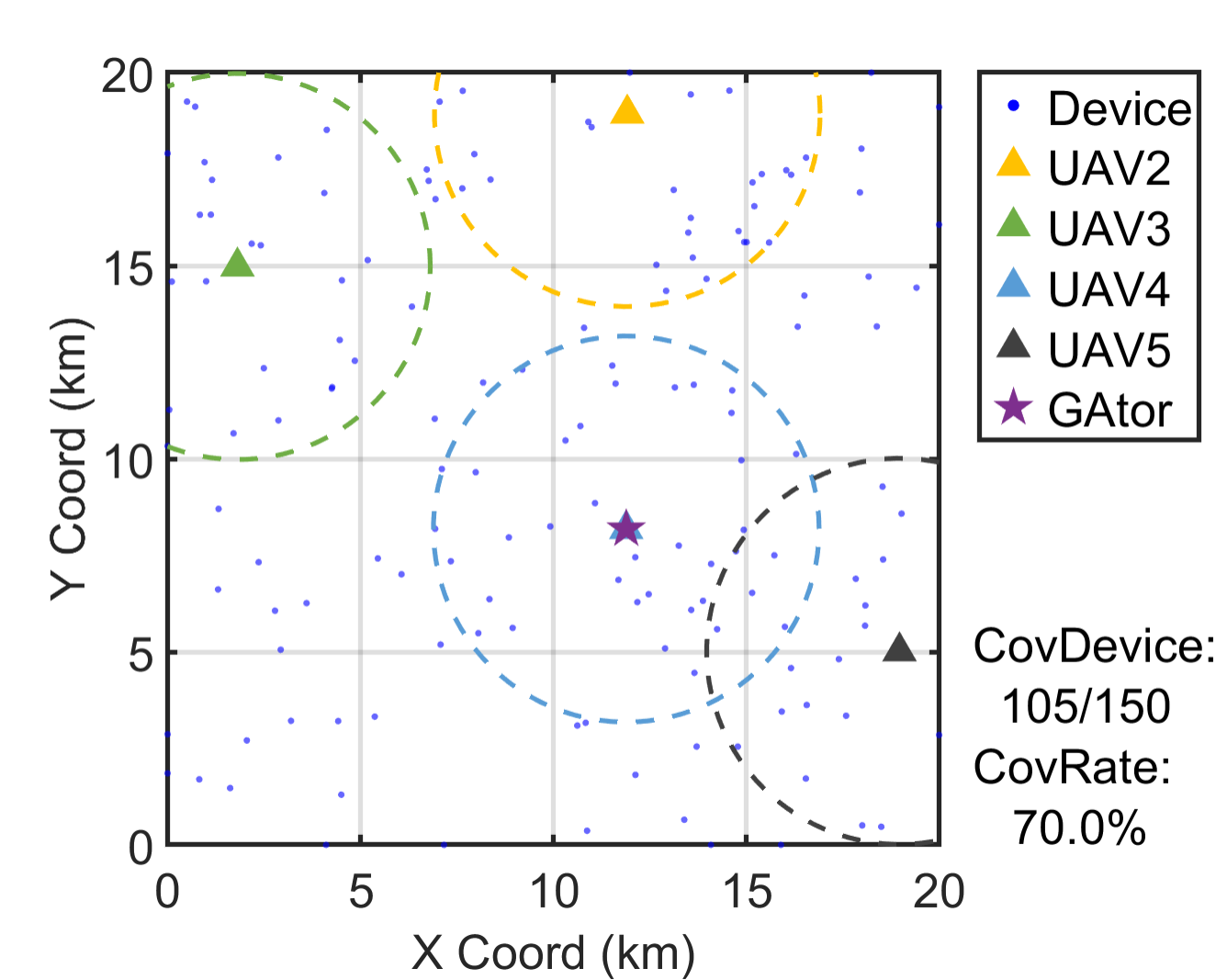}
	}
	\subfigure[Fin. 2 UAVs Drop]{
		\includegraphics[trim=0.2cm 0cm 0.2cm 0.2cm, clip, width=0.295\columnwidth]{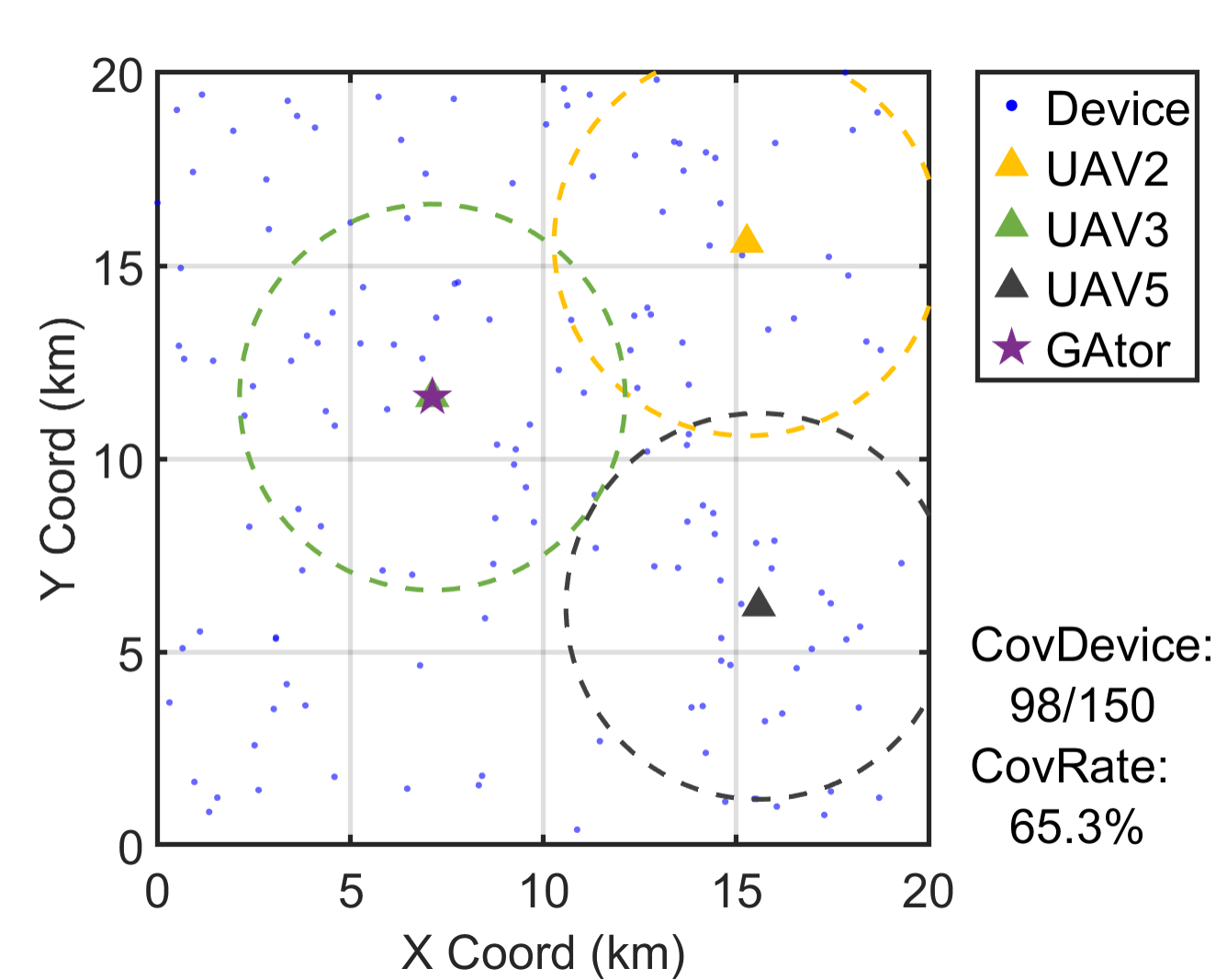}
	}
\vspace{-3mm}	
	\caption{Redeployment dynamics of UAVs following disconnection events.}
	\vspace{-3mm}
	\label{fig_11}
\end{figure}

\subsection{Convergence Analysis of Our HFL}
The HFL proposed in this paper consists of three modules. Firstly, we have demonstrated the convergence of the augmented Lagrangian algorithm with penalty term in \textbf{Appendix B.D}, while the convergence of the TD3 is detailed in \cite{Approximation}. Then, we prove the convergence of Alg. 4 as follows: First, the state space is finite due to the limited set of selectable positions for each UAV. Second, this algorithm employs a greedy non-decreasing strategy, accepting only solutions that improve upon the current one, thereby ensuring that the sequence of objective function values is monotonically non-decreasing. Third, since the objective function is bounded above, the sequence converges by the monotone convergence theorem. Also, Alg. 4 sets the maximum number of explorations $\chi_1 $ and $\chi_2 $ to ensure that it does not enter an infinite loop. In an overall view, Alg. 4 converges to a local optimal solution within a finite number of steps. Hence, the convergence of the proposed HFL can be ensured. More importantly, the convergence is further validated by simulation results in Fig. 4.

\subsection{UAV Redeployment Performance After Disconnections}
Fig. 11, we present the performance of the two-stage greedy algorithm, evaluating its effectiveness when one or two UAVs disconnect. Fig. 11(a) illustrates the initial scenario, where
UAV 4 disconnects at a specific round of global iterations. Following this, Figs. 11(b) and 11(c) depict the network state after UAV 4’s disconnection and the subsequent redeployment of the remaining UAVs. During this process, the UAV coverage rate for devices initially drops from 85$\%$ to 72$\%$ but recovers to 77.3$\%$ after redeployment, demonstrating the algorithm’s
ability to restore network coverage efficiently. Figs. 11(d)-(f) further examine the scenario where both UAV 1 and UAV 4 disconnect. When UAV 1 is lost, the coverage rate decreases by 16$\%$, followed by an additional 4.7$\%$ drop when UAV 4 disconnects. The above simulation results demonstrate the effectiveness of our proposed solution for $\mathcal{P}_3$. 

\subsection{More Simulations Regarding UAV Disconnections}
\begin{table}[!htbp]
  \centering
  \caption{Device coverage analysis for various UAV redeployment methods.}
  \begingroup
  \setlength{\tabcolsep}{2pt}     
  \renewcommand{\arraystretch}{1} 
  \setlength{\heavyrulewidth}{1.2pt}
  \small  
    \begin{tabular}{ccccccrrr}
    \toprule
          & \multicolumn{5}{c}{1 UAV drop (\%)}   & \multicolumn{3}{c}{2 UAVs drop (\%)} \\
          & 1     & 2     & 3     & 4     & 5     & \multicolumn{1}{c}{1 and 2} & \multicolumn{1}{c}{1 and 4} & \multicolumn{1}{c}{3 and 5} \\
    \midrule
    \multirow{2}[1]{*}{\makecell{\underline{L}}} & \multirow{2}[1]{*}{-8.21} & \multirow{2}[1]{*}{4.67} & \multirow{2}[1]{*}{-8.67} & \multirow{2}[1]{*}{{-8.00}} & \multirow{2}[1]{*}{-6.32} & -14.66 & {-16.00} & -11.33 \\
          &       &       &       &       &       & -7.34  & {-4.70}  & -8.67 \\
    \hline
    \multirow{2}[1]{*}{\makecell{\underline{M}}} & \multirow{2}[1]{*}{-17.34} & \multirow{2}[1]{*}{-21.33} & \multirow{2}[1]{*}{-24.05} & \multirow{2}[1]{*}{-23.34} & \multirow{2}[1]{*}{-21.06} & -23.34 & -20.13 & -16.22 \\
          &       &       &       &       &       & -7.66  & -24.07 & -19.33 \\
    \hline
    \multirow{2}[1]{*}{\makecell{\underline{N}}} & \multirow{2}[1]{*}{-9.67} & \multirow{2}[1]{*}{-9.34} & \multirow{2}[1]{*}{-1.33} & \multirow{2}[1]{*}{-6.67} & \multirow{2}[1]{*}{-8.67} & -11.81 & -12.67 & -8.67 \\
          &       &       &       &       &       & -12.00 & -13.50 & -16.20 \\
    \bottomrule
    \end{tabular}%
  \endgroup
  \label{tab:addlabel}
\end{table}

\begin{table}[!htbp]
  \centering
  \caption{Energy consumption of various UAV redeployment methods.}
  \begingroup
  \setlength{\tabcolsep}{2pt}    
  \renewcommand{\arraystretch}{1} 
  \setlength{\heavyrulewidth}{1.2pt}
  \small  
    \begin{tabular}{ccccccccc}
    \toprule
          & \multicolumn{5}{c}{1 UAV drop (J)}    & \multicolumn{3}{c}{2 UAVs drop (J)} \\
          & 1     & 2     & 3     & 4     & 5     & 1 and 2 & 1 and 4 & 3 and 5 \\
    \midrule
    \multirow{2}[1]{*}{\makecell{\underline{L}}} & \multirow{2}[1]{*}{20.62} & \multirow{2}[1]{*}{22.97} & \multirow{2}[1]{*}{21.56} & \multirow{2}[1]{*}{{18.00}} & \multirow{2}[1]{*}{20.10} & 10.78 & 9.84 & 10.75 \\
          &       &       &       &       &       & 31.87 & 25.63 & 15.62 \\
    \hline
    \multirow{2}[1]{*}{\makecell{\underline{N}}} & \multirow{2}[1]{*}{48.48} & \multirow{2}[1]{*}{53.36} & \multirow{2}[1]{*}{42.03} & \multirow{2}[1]{*}{{33.06}} & \multirow{2}[1]{*}{39.43} & 45.42 & 58.86 & 30.29 \\
          &       &       &       &       &       & 74.27 & 66.55 & 46.40 \\
    \bottomrule
    \end{tabular}%
  \endgroup
  \label{tab:addlabel}
\end{table}

For better comparison, we designed two methods: \textit{(M-i)} After the UAV is disconnected, the UAV does not move (i.e., the UAV is directly dropped). \textit{(M-ii)} Constructed by integrating UAV energy consumption, coverage rate, and inter-UAV communication energy, the benefit function is optimized using a greedy algorithm to identify the solution that maximizes its value during the movement process, and the letters `L' corresponds to our method, `M' corresponds to \textit{(M-i)}, and `N' corresponds to \textit{(M-ii)}. In Table III, we compare the changes in UAV coverage rate of the three methods when a UAV is dropped (e.g., in our method, while UAV 1 is disconnected, causing the overall UAV coverage rate to drop by 8.21$\%$), and when 2 UAVs are dropped (e.g., in our method, the coverage rate drops by 14.66$\%$ after UAV 1 is dropped, and then the coverage rate drops by an additional 7.34$\%$ after UAV 2 is dropped).  We can also find that, if no UAV repositioning occurs, the coverage reduction is more severe, dropping by 20.13$\%$ and 24.07$\%$ for UAV 1 and UAV 4, respectively (see Table III, M). These results confirm that our proposed redeployment strategy effectively mitigates device coverage loss caused by UAV disconnections. 
Additionally, for the disconnection event in the same scenario, we compare the energy consumed by the UAV in our method and `N' --- which is the best baseline according to Table IV --- in the process of finding the optimal position in Table IV. These results further demonstrate that our proposed algorithm achieves a lower UAV redeployment energy costs compared to the best baseline. Collectively, the results in Tables III and IV unveil that our method
strikes a desirable balance between maintaining UAV coverage and minimizing UAV mobility energy costs.
\end{document}